\documentclass[10pt,twocolumn,letterpaper]{article}

\usepackage{cvpr}
\usepackage{times}
\usepackage{epsfig}
\usepackage{graphicx}
\usepackage{amsmath}
\usepackage{amssymb}
\usepackage{array}
\usepackage{multirow}
\usepackage{bm}
\usepackage{algorithm, algpseudocode}
\usepackage{subfigure}
\usepackage{nicefrac}
\usepackage{caption}
\usepackage{makecell,rotating}
\usepackage[pagebackref=true,breaklinks=true,letterpaper=true,colorlinks,bookmarks=false]{hyperref}
\DeclareMathOperator*{\argmax}{arg\,max} 
\DeclareMathOperator*{\argmin}{arg\,min} 

\cvprfinalcopy 

\def\cvprPaperID{5414} 

\begin{document}

\title{Benchmarking Adversarial Robustness}

\author{Yinpeng Dong$^{1}$, Qi-An Fu$^{1}$, Xiao Yang$^{1}$, Tianyu Pang$^{1}$, Hang Su$^{1}$, Zihao Xiao$^{2}$, Jun Zhu$^{1}$\\
$^{1}$ Dept. of Comp. Sci. and Tech., BNRist Center, State Key Lab for Intell. Tech. \& Sys.,\\
$^{1}$ Institute for AI, THBI Lab, Tsinghua University, Beijing, 100084, China \hspace{2ex} $^{2}$ RealAI\\
\tt\small{\{dyp17, fqa19, yangxiao19, pty17\}@mails.tsinghua.edu.cn} \\ \tt\small{\{suhangss, dcszj\}@mail.tsinghua.edu.cn} \hspace{2ex} zihao.xiao@realai.ai}

\maketitle

\begin{abstract}
    Deep neural networks are vulnerable to adversarial examples, which becomes one of the most important research problems in the development of deep learning. While a lot of efforts have been made in recent years, it is of great significance to perform correct and complete evaluations of the adversarial attack and defense algorithms. In this paper, we establish a comprehensive, rigorous, and coherent benchmark to evaluate adversarial robustness on image classification tasks. After briefly reviewing plenty of representative attack and defense methods, we perform large-scale experiments with two robustness curves as the fair-minded evaluation criteria to fully understand the performance of these methods. Based on the evaluation results, we draw several important findings and provide insights for future research.
\end{abstract}

\section{Introduction}

Recent progress in deep learning (DL) has led to substantial improvements in a wide range of computer vision tasks~\cite{krizhevsky2012imagenet,He2015}.
However, the existing DL models are highly vulnerable to adversarial examples~\cite{szegedy2013intriguing,goodfellow2014explaining}, which are maliciously generated by an adversary to make a model produce erroneous predictions. 
As DL models have been integrated into various security-sensitive applications (\eg, autonomous driving, healthcare, and finance), the study of the adversarial robustness issue has attracted increasing attention with an enormous number of adversarial attack and defense methods proposed. Therefore, it is crucial to conduct correct and rigorous evaluations of these methods for understanding their pros and cons, comparing their performance, and providing insights for building new methods~\cite{carlini2019evaluating}.

The research on adversarial robustness is faced with an ``\emph{arms race}'' between attacks and defenses, \ie, a defense method proposed to prevent the existing attacks was soon evaded by new attacks, and vice versa~\cite{carlini2017adversarial,carlini2017towards,he2018decision,Athalye2018Obfuscated,uesato2018adversarial,zhang2019limitations}.
For instance, defensive distillation~\cite{PapernotDistillation2016} was proposed to improve adversarial robustness, but was later shown to be ineffective against a strong attack~\cite{carlini2017towards}. Many methods were introduced to build robust models by causing obfuscated gradients, which can be defeated by the adaptive ones~\cite{Athalye2018Obfuscated,uesato2018adversarial}.
As a result, it is particularly challenging to understand their effects, identify the real progress, and advance the field.

Moreover, the current attacks and defenses are often evaluated incompletely. 
First, most defenses are only tested against a small set of attacks under limited threat models, and many attacks are evaluated on a few models or defenses.
Second, the robustness evaluation metrics are too simple to show the performance of these methods. The accuracy of a defense against an attack for a given perturbation budget~\cite{kurakin2018competation} and the minimum distance of the adversarial perturbation~\cite{brendel2018adversarial} are used as the primary evaluation metrics, which are often insufficient to totally characterize the behaviour of the attacks and defenses.
Consequently, the incomplete evaluation cannot provide a comprehensive understanding of the strengths and limitations of these methods.

In this paper, we establish a comprehensive, rigorous, and coherent benchmark to evaluate adversarial robustness, which can provide a detailed understanding of the effects of the existing methods under different scenarios, with a hope to facilitate future research.
In particular, we focus on the robustness of image classifiers under the $\ell_p$ norm threat models, since the adversarial robustness issue has been extensively studied on image classification tasks with the $\ell_p$ additive noises.
We incorporate a lot of typical and state-of-the-art attack and defense methods for robustness evaluation, including $15$ attack methods and $16$ defense models---$8$ on CIFAR-10~\cite{krizhevsky2009learning} and $8$ on ImageNet~\cite{russakovsky2015imagenet}.
To fully demonstrate the performance of these methods, we adopt two complementary robustness curves as the major evaluation metrics to present the results. 
Then, we carry out large-scale experiments on the cross evaluation of the attack and defense methods under complete threat models, including 1) untargeted and targeted attacks; 2) $\ell_{\infty}$ and $\ell_2$ attacks; 3) white-box, transfer-based, score-based, and decision-based attacks.
We consider complete threat models for evaluation even if a defense does not claim to be robust under some threat models, since the purpose of this paper is \emph{not to contradict the previous results of some methods, but to provide a comprehensive evaluation of the existing methods}.

By analyzing the quantitative results, we have some important findings.
First, the relative robustness between defenses against an attack could be different under varying perturbation budgets or attack iterations. Thus it is hard to conclude that a defense is more robust than another against an attack by using a specific configuration. However, this is common in previous works.
Second, although various defense techniques have been proposed, the most robust defenses are still the adversarially trained models. Their robustness can also generalize to other threat models, under which they are not trained to be robust.
Third, defenses based on randomization are generally more robust to black-box attacks based on the query feedback.
More detailed discussions can be found in Sec.~\ref{sec:remarks}.

All evaluation experiments are conducted on a new adversarial robustness platform developed by us\footnote{The source code will be released soon.}, since the existing platforms (\eg, CleverHans~\cite{papernot2016technical}, Foolbox~\cite{rauber2017foolbox}, \etc) cannot fully support our comprehensive evaluations (details in Appendix~\ref{sec:app-a}). We hope that our platform could continuously incorporate and evaluate more methods, and be helpful for future works.

\section{Threat Models}
\label{sec:threat-model}

Precisely defining threat models is fundamental to perform adversarial robustness evaluations.
According to~\cite{carlini2019evaluating}, a threat model specifies the adversary's goals, capabilities, and knowledge under which an attack is performed and a defense is built to be robust.
We first define the notations and then illustrate the three aspects of a threat model. 

A classifier can be denoted as $C(\bm{x}): \mathcal{X} \rightarrow \mathcal{Y}$, where $\bm{x}\in\mathcal{X}\subset\mathbb{R}^d$ is the input, and $\mathcal{Y} = \{1,2,...,L\}$ with $L$ being the number of classes. 
Let $y$ denote the ground-truth label of $\bm{x}$, and $\bm{x}^{adv}$ denote an adversarial example for $\bm{x}$.

\subsection{Adversary's Goals}
An adversary can have different goals of generating adversarial examples. We study the \emph{untargeted} and \emph{targeted} adversarial examples in this paper. An untargeted adversarial example aims to cause misclassification of the classifier, as $C(\bm{x}^{adv}) \neq y$. A targeted one is crafted to be misclassified as the adversary-desired target class by the classifier, as $C(\bm{x}^{adv}) = y^*$, where $y^*$ is the target class.

\subsection{Adversary's Capabilities}
As adversarial examples are usually assumed to be indistinguishable from the corresponding original ones to human eyes~\cite{szegedy2013intriguing,goodfellow2014explaining}, the adversary can only make small changes to the inputs. In this paper, we study the well-defined and widely used $\ell_p$ norm threat models, although there also exist other threat models~\cite{xiao2018spatially,song2018constructing,engstrom2019exploring}. Under the $\ell_p$ norm threat models, the adversary is allowed to add a small perturbation measured by the $\ell_p$ norm to the original input. Specifically, we consider the $\ell_{\infty}$ and $\ell_2$ norms.

To achieve the adversary's goal, two strategies could be adopted to craft adversarial examples with small perturbations.
The first seeks to craft an adversarial example $\bm{x}^{adv}$ that satisfies $\|\bm{x}^{adv}-\bm{x}\|_p \leq \epsilon$, where $\epsilon$ is the perturbation budget, while misleads the model. This could be achieved by solving a constrained optimization problem. For instance, the adversary can get an untargeted adversarial example by maximizing a loss function $\mathcal{J}$ (\eg., the cross-entropy loss) in the restricted region as
\begin{equation}
\label{eq:constrain-pert}
 \bm{x}^{adv}= \argmax_{\bm{x}':\|\bm{x}'-\bm{x}\|_p\leq\epsilon}\mathcal{J}(\bm{x}',y).
\end{equation}
We call it the adversarial example with a \emph{constrained} perturbation.
The second strategy is generating an adversarial example by finding the minimum perturbation as 
\begin{equation}
\label{eq:optimized-pert}
    \bm{x}^{adv} =\argmin_{\bm{x}':\bm{x}' \text{is adversarial}}\|\bm{x}'-\bm{x}\|_p.
\end{equation}
We call it the adversarial example with an \emph{optimized} perturbation. However, it is usually intractable to solve Eq.~(\ref{eq:constrain-pert}) or Eq.~(\ref{eq:optimized-pert}) exactly, and thus various attack methods have been proposed to get an approximate solution.

\subsection{Adversary's Knowledge}

An adversary can have different levels of knowledge of the target model, from white-box access to the model architectures and parameters, to black-box access to the training data or model predictions.
Based on the different knowledge of the model, we consider four attack scenarios, including \emph{white-box attacks}, \emph{transfer-based}, \emph{score-based}, and \emph{decision-based black-box attacks}.

White-box attacks rely on detailed information of the target model, including architecture, parameters, and gradient of the loss w.r.t. the input. 
For the defense models, the adversary also has access to the specific defense mechanisms, and designs adaptive attacks to evade them. Transfer-based black-box attacks are based on the transferability of adversarial examples~\cite{Papernot2016}. These attacks do not rely on model information but assume the availability of the training data. It is used to train a substitute model from which the adversarial examples are generated.
Score-based black-box attacks can only acquire the output probabilities by querying the target model. 
And decision-based black-box attacks solely rely on the predicted classes of the queries.
Score-based and decision-based attacks are also restricted by a limited number of queries to the target model.

\section{Attacks and Defenses}
In this section, we summarize the typical adversarial attack and defense methods.

\subsection{Attack Methods}
\label{sec:attacks}


\textbf{White-box Attacks:} Most white-box attacks craft adversarial examples based on the input gradient.
To solve Eq.~(\ref{eq:constrain-pert}), the fast gradient sign method (\textbf{FGSM})~\cite{goodfellow2014explaining} linearizes the loss function in the input space and generates an adversarial example by an one-step update. 
The basic iterative method (\textbf{BIM})~\cite{Kurakin2016} extends FGSM by iteratively taking multiple small gradient steps.
Similar to BIM, the projected gradient descent method (\textbf{PGD})~\cite{madry2017towards} acts as a universal first-order adversary with random starts. 
To solve Eq.~(\ref{eq:optimized-pert}), \textbf{DeepFool}~\cite{Moosavidezfooli2016} has been proposed to generate an adversarial example with the minimum perturbation.
The Carlini \& Wagner's method (\textbf{C\&W})~\cite{carlini2017towards} takes a Lagrangian form and adopts Adam~\cite{Kingma2014} for optimization.
However, some defenses can be robust against these gradient-based attacks by causing obfuscated gradients~\cite{Athalye2018Obfuscated}. To circumvent them, the adversary can use \textbf{BPDA}~\cite{Athalye2018Obfuscated} to provide an approximate gradient when the true gradient is unavailable or useless, or \textbf{EOT}~\cite{Athalye2017Synthesizing} when the gradient is random.

\textbf{Transfer-based Black-box Attacks:} Transfer-based attacks generate adversarial examples against a substitute model, which have a probability to fool black-box models based on the transferability. Besides the above methods, some others have also been proposed to improve the transferability. The momentum iterative method (\textbf{MIM})~\cite{Dong2017} integrates a momentum term into BIM to stabilize the update direction during the attack iterations.
The diverse inputs method (\textbf{DIM})~\cite{xie2019improving} applies the gradient of the randomly resized and padded input for adversarial example generation. The translation-invariant method (\textbf{TI})~\cite{dong2019evading} further improves the transferability for defense models.

\textbf{Score-based Black-box Attacks:} In this setting, although the white-box access to the model gradient is unavailable, it can be estimated by the gradient-free methods through queries. \textbf{ZOO}~\cite{chen2017zoo} estimates the gradient at each coordinate by finite differences and adopts C\&W for attacks based on the estimated gradient. \textbf{NES}~\cite{ilyas2018black} and \textbf{SPSA}~\cite{uesato2018adversarial} can give the full gradient estimation based on drawing random samples and acquiring the corresponding loss values.
Prior-guided random gradient free method (\textbf{P-RGF})~\cite{cheng2019improving} estimates the gradient more accurately with a transfer-based prior.
$\mathcal{N}$\textbf{ATTACK}~\cite{li2019nattack} does not estimate the gradient but learns a Gaussian distribution centered around the input such that a sample drawn from it is likely adversarial.

\textbf{Decision-based Black-box Attacks:} This setting is more challenging since the model only provides discrete hard-label predictions. The \textbf{Boundary} attack~\cite{Brendel2018Decision} is the first method in this setting based on random walk on the decision boundary. An \textbf{optimization-based} method~\cite{cheng2019query} formulates this problem as a continuous optimization problem and estimates the gradient to solve it. The \textbf{evolutionary} attack method~\cite{dong2019efficient} is further proposed to improve the query efficiency based on the evolution strategy.

\subsection{Defenses}
\label{sec:defenses}

Due to the threat of adversarial examples, extensive research has been conducted on building robust models to defend against adversarial attacks.
In this paper, we roughly classify the defense techniques into five categories, including \emph{robust training}, \emph{input transformation}, \emph{randomization}, \emph{model ensemble}, and \emph{certified defenses}. Note that these defense categories are not exclusive, \ie, a defense can belong to many categories. Below we introduce each category.

\textbf{Robust Training:} The basic principle of robust training is to make the classifier robust against small perturbations internally. One line of work is based on adversarial training~\cite{goodfellow2014explaining,tramer2017ensemble,madry2017towards,kannan2018adversarial,zhang2019theoretically}, which augments the training data by adversarially generated examples.
Another line of work trains  robust models by regularizations, including those on the Lipschitz constant~\cite{cisse2017parseval}, input gradients~\cite{hein2017formal,ross2018improving}, or perturbation norm~\cite{yan2018deep}.

\textbf{Input Transformation:} Several defenses transform the inputs before feeding them to the classifier, including JPEG compression~\cite{dziugaite2016study}, bit-depth reduction~\cite{xu2018feature}, total variance minimization~\cite{guo2017countering}, autoencoder-based denoising~\cite{liao2018defense}, and projecting adversarial examples onto the data distribution through generative models~\cite{samangouei2018defense,song2017pixeldefend}.
However, these defenses can cause shattered gradients or vanishing/exploding gradients~\cite{Athalye2018Obfuscated}, which can be evaded by adaptive attacks.

\textbf{Randomization:} The classifiers can be made random to mitigate adversarial effects. The randomness can be added to either the input~\cite{xie2017mitigating} or the model~\cite{dhillon2018stochastic,liu2018towards}. The randomness can also be modeled by Bayesian neural networks~\cite{liu2019adv}. These methods partially rely on random gradients to prevent adversarial attacks, and can be defeated by attacks that take the expectation over the random gradients~\cite{he2018decision,Athalye2018Obfuscated}. 

\textbf{Model Ensemble:} An effective defense strategy in practice is to construct an ensemble of individual models~\cite{kurakin2018competation}. 
Besides aggregating the output of each model in the ensemble, some different ensemble strategies have been proposed. Random self-ensemble~\cite{liu2018towards} averages the predictions over random noises injected to the model, which is equivalent to ensemble an infinite number of noisy models.
Pang \etal~\cite{pang2019improving} propose to promote the diversity among the predictions of different models, and introduce an adaptive diversity promoting regularizer to achieve this.

\textbf{Certified Defenses:} There are a lot of works~\cite{raghunathan2018certified,sinha2018certifying,wong2018provable,wong2018scaling,raghunathan2018semidefinite,xiao2019training} on training certified defenses, which are provably guaranteed to be robust against adversarial perturbations under some threat models. Recently, certified defenses~\cite{zhang2019discretization,cohen2019certified} can apply to ImageNet~\cite{russakovsky2015imagenet}, showing the scalability of this type of defenses.

\begin{table*}[t]
\footnotesize
\setlength{\tabcolsep}{5pt}
  \begin{center}
  \begin{tabular}{c|c|c|c||c|c|c|c}
    \hline
    \multicolumn{4}{c||}{CIFAR-10~\cite{krizhevsky2009learning}} & \multicolumn{4}{c}{ImageNet~\cite{russakovsky2015imagenet}} \\
    \hline
    Defense Model  & Category & Intended Threat & Acc. &  Defense Model  & Category & Intended Threat & Acc. \\
    \hline\hline
    Res-56~\cite{He2015}   & natural training & - & 92.6 & Inc-v3~\cite{szegedy2016rethinking} & natural training & - & 78.0\\
    \hline
    PGD-AT~\cite{madry2017towards} & robust training & $\ell_\infty$ ($\epsilon=8/255$) & 87.3 &
    Ens-AT~\cite{tramer2017ensemble} & robust training & $\ell_\infty$ ($\epsilon=16/255$) & 73.5\\
    \hline
    DeepDefense~\cite{yan2018deep} &  robust training & $\ell_2$ & 79.7 & 
    ALP~\cite{kannan2018adversarial}  & robust training & $\ell_\infty$ ($\epsilon=16/255$) & 49.0 \\
    \hline
    TRADES~\cite{zhang2019theoretically} & robust training & $\ell_\infty$ ($\epsilon=8/255$) & 84.9 &
    FD~\cite{xie2019feature} & robust training & $\ell_\infty$ ($\epsilon=16/255$) & 64.3 \\
    \hline 
    Convex~\cite{wong2018scaling} & (certified) robust training & $\ell_\infty$ ($\epsilon=2/255$) & 66.3 & JPEG~\cite{dziugaite2016study} & input transformation & General & 77.3 \\
    \hline
    JPEG~\cite{dziugaite2016study} & input transformation & General & 80.9 & Bit-Red~\cite{xu2018feature} &  input transformation & General & 61.8 \\
    \hline
    RSE~\cite{liu2018towards} & rand. \& ensemble & $\ell_2$ & 86.1 & R\&P~\cite{xie2017mitigating} & (input) rand. & General & 77.0 \\
    \hline
     ADP~\cite{pang2019improving} & ensemble & General & 94.1 & RandMix~\cite{zhang2019discretization} & (certified input) rand. & General & 52.4 \\
    \hline
  \end{tabular}
  \end{center}
  \vspace{-4ex}
   \caption{We show the defense models that are incorporated into our benchmark for adversarial robustness evaluation. We also show the defense type, original intended threat model (\ie, the threat model under which the defense is trained to be robust or evaluated in the original paper; `General' means the defense can be used for any threat model), and accuracy (\%) on clean data of each method. The accuracy is re-calculated by ourselves. More details about their model architectures are shown in Appendix~\ref{sec:app-b}.}
   \label{tab:defense-summary}
   \vspace{-2ex}
\end{table*}

\section{Evaluation Methodology}
With the growing number of adversarial attacks and defenses being proposed, the correct and rigorous evaluation of these methods becomes increasingly important to help us better understand the strengths and limitations of these methods. However, there still lacks a comprehensive understanding of the effects of these methods due to the incorrect or incomplete evaluations. 
To address this issue and further advance the field, we establish a comprehensive, rigorous, and coherent benchmark to evaluate adversarial robustness empirically.
We incorporate $15$ attack methods and $16$ defense models on two image datasets in our benchmark for robustness evaluation. We also adopt two complementary robustness curves as the fair-minded evaluation metrics to better show the results.

\subsection{Evaluation Metrics}
\label{sec:metric}

Given an attack method $\mathcal{A}_{\epsilon, p}$ that generates an adversarial example $\bm{x}^{adv} = \mathcal{A}_{\epsilon, p}(\bm{x})$ for an input $\bm{x}$ with perturbation budget $\epsilon$ under the $\ell_p$ norm\footnote{For attacks that find minimum perturbations, \eg, DeepFool, C\&W, we let $\mathcal{A}_{\epsilon, p}(\bm{x})=\bm{x}$ if the $\ell_p$ norm of the perturbation is larger than $\epsilon$.}, and a (defense) classifier $C$ defined in Sec.~\ref{sec:threat-model}, the accuracy of the classifier against the attack is defined as
\begin{equation*}
    \mathrm{Acc}(C, \mathcal{A}_{\epsilon, p}) = \frac{1}{N}\sum_{i=1}^{N}\mathbf{1}\big(C(\mathcal{A}_{\epsilon, p}(\bm{x}_i))=y_i\big),
\end{equation*}
where $\{\bm{x}_i, y_i\}_{i=1}^{N}$ is the test set, $\mathbf{1}(\cdot)$ is the indicator function. The attack success rate of an untargeted attack on the classifier is defined as
\begin{equation*}
    \mathrm{Asr}(\mathcal{A}_{\epsilon, p}, C) = \frac{1}{M}\sum_{i=1}^{N}\mathbf{1}\big(C(\bm{x}_i) = y_i \land C(\mathcal{A}_{\epsilon, p}(\bm{x}_i))\neq y_i\big),
\end{equation*}
where $M=\sum_{i=1}^{N}\mathbf{1}\big(C(\bm{x}_i) = y_i\big)$, while the attack success rate of a targeted attack is defined as
\begin{equation*}
     \mathrm{Asr}(\mathcal{A}_{\epsilon, p}, C) = \frac{1}{N}\sum_{i=1}^{N} \mathbf{1}\big(C(\mathcal{A}_{\epsilon, p}(\bm{x}_i))= y_i^*\big).
\end{equation*}
where $y_i^*$ is the target class corresponding to $\bm{x}_i$.

The previous methods usually report the point-wise accuracy or attack success rate for some chosen perturbation budgets $\epsilon$, which may not reflect their behaviour totally.
In this paper, we adopt two complementary robustness curves to clearly and thoroughly show the robustness and resistance of the classifier against the attack, as well as the effectiveness and efficiency of the attack on the classifier. 

The first one is the \emph{accuracy (attack success rate) vs. perturbation budget} curve, which can give a global understanding of the robustness of the classifier and the effectiveness of the attack.
To generate such a curve, we need to calculate the accuracy or attack success rate for all values of $\epsilon$.
This can be efficiently done for attacks that find the minimum perturbations, by counting the number of the adversarial examples, the $\ell_p$ norm of whose perturbations is smaller than each $\epsilon$.
For attacks that craft adversarial examples with constrained perturbations, we perform a binary search on $\epsilon$ to find its minimum value that enables the generated adversarial example to fulfill the adversary's goal.

The second curve is the \emph{accuracy (attack success rate) vs. attack strength} curve, where the attack strength is defined as the number of iterations or model queries based on different attack methods. 
This curve can show the efficiency of the attack, as well as the resistance of the classifier to the attack, \eg, a defense whose accuracy drops to zero against an attack with $100$ iterations is considered to be more resistant to this attack than another defense that is totally broken by the same attack with $10$ iterations, although the worst-case accuracy of both models is zero.

\begin{table}
\footnotesize
\setlength{\tabcolsep}{3pt}
  \begin{center}
  \begin{tabular}{c|c|c|c|c}
    \hline
   Attack Method & Knowledge & Goals & Capability & Distance \\
    \hline\hline
    FGSM~\cite{goodfellow2014explaining} & white \& transfer & un. \& tar. & constrained & $\ell_{\infty}$, $\ell_2$ \\
    \hline
    BIM~\cite{Kurakin2016} & white \& transfer & un. \& tar. & constrained & $\ell_{\infty}$, $\ell_2$ \\
    \hline
    MIM~\cite{Dong2017} & white \& transfer & un. \& tar. & constrained & $\ell_{\infty}$, $\ell_2$ \\
    \hline
    DeepFool~\cite{Moosavidezfooli2016} & white & un. & optimized & $\ell_{\infty}$, $\ell_2$ \\
    \hline
    C\&W~\cite{carlini2017towards} & white & un. \& tar. & optimized & $\ell_2$ \\
    \hline
    DIM~\cite{xie2019improving} & transfer & un. \& tar. & constrained & $\ell_{\infty}$, $\ell_2$ \\
    \hline
    ZOO~\cite{chen2017zoo} & score & un. \& tar. & optimized & $\ell_2$ \\
    \hline
    NES~\cite{ilyas2018black} & score & un. \& tar. & constrained & $\ell_{\infty}$, $\ell_2$ \\
    \hline
    SPSA~\cite{uesato2018adversarial} & score & un. \& tar. & constrained & $\ell_{\infty}$, $\ell_2$ \\
    \hline
    $\mathcal{N}$ATTACK~\cite{li2019nattack} & score & un. \& tar. & constrained & $\ell_{\infty}$, $\ell_2$ \\
    \hline
    Boundary~\cite{Brendel2018Decision} & decision & un. \& tar. & optimized & $\ell_2$ \\
    \hline
    Evolutionary~\cite{dong2019efficient} & decision & un. \& tar. & optimized & $\ell_2$ \\
    \hline
  \end{tabular}
  \end{center}
  \vspace{-3ex}
  \caption{We show the attack methods that are implemented in our benchmark for adversarial robustness evaluation. We also show the adversary's knowledge (white-box, transfer-based, score-based, or decision-based), goals (`un.' stands for untargeted; `tar.' stands for targeted), capability (constrained or optimized perturbations), and distance metrics of each attack method.}
  \label{tab:attack-summary}
  \vspace{-2ex}
\end{table}

\begin{table*}[t]
\footnotesize
\setlength{\tabcolsep}{3pt}
  \begin{center}
  \begin{tabular}{c|c|c|c|c|c|c|c|c|c}
    \hline
    \multicolumn{2}{c|}{Attack} & Res-56 & PGD-AT & DeepDefense & TRADES & Convex & JPEG & RSE & ADP \\
    \hline\hline
    \multirow{4}{*}{White} & FGSM & $0.005 / 21.6\%$ & $0.039 / 56.0\%$ & $0.001 / 9.2\%$ & $0.047 / 60.9\%$ & $0.017 / 36.6\%$ & $0.012 / 31.2\%$ & $0.020 / 29.0\%$ & $0.037 / 56.0\%$\\
    & BIM & $0.002 / 0.0\%$ & $0.030 / 48.3\%$ & $0.001 / 0.0\%$ & $0.037 / 56.8\%$ & $0.016 / 34.3\%$ & $0.008 / 3.2\%$ & $0.018 / 23.5\%$ & $0.008 / 12.2\%$ \\
    & MIM & $0.003 / 0.0\%$ & $0.032 / 50.9\%$ & $0.001 / 0.0\%$ & $0.040 / 58.1\%$ & $0.016 / 34.9\%$ & $0.008 / 6.1\%$ & $0.019 / 25.1\%$ & $0.010 / 16.7\%$ \\
    & DeepFool & $0.003 / 0.0\%$ & $0.040 / 56.5\%$ & $0.001 / 0.0\%$ & $0.047 / 60.6\%$ & $0.015 / 32.9\%$ & $0.007 / 3.1\%$ & $0.021 / 35.9\%$ & $0.016 / 28.7\%$\\
    \hline
    \multirow{4}{*}{Transfer} & FGSM & $0.067 / 72.9\%$ & $0.067 / 71.3\%$ & $0.048 / 62.1\%$ & $0.087 / 73.6\%$ & $0.050 / 57.5\%$ & $0.051 / 62.8\%$ & $0.048 / 62.0\%$ & $0.066 / 73.4\%$ \\
    & BIM & $0.049 / 70.3\%$ & $0.055 / 70.2\%$ & $0.041 / 58.8\%$ & $0.069 / 72.2\%$ & $0.044 / 56.7\%$ & $0.039 / 58.9\%$ & $0.041 / 60.0\%$ & $0.048 / 71.4\%$ \\
    & MIM & $0.052 / 71.5\%$ & $0.056 / 70.4\%$ & $0.041 / 59.4\%$ & $0.067 / 72.2\%$ & $0.045 / 56.6\%$ & $0.041 / 59.9\%$ & $0.043 / 59.8\%$ & $0.050 / 70.4\%$ \\
    & DIM & $0.052 / 73.3\%$ & $0.056 / 70.0\%$ & $0.043 / 58.8\%$ & $0.063 / 70.5\%$ & $0.044 / 55.3\%$ & $0.043 / 61.1\%$ & $0.043 / 60.2\%$ & $0.051 / 73.4\%$ \\
    \hline
    \multirow{3}{*}{Score} & NES & $0.004 / 0.0\%$ & $0.048 / 65.5\%$ & $0.002 / 0.0\%$ & $0.055 / 66.7\%$ & $0.025 / 44.0\%$ & $0.001 / 2.1\%$ & $0.293 / 79.7\%$ & $0.007 / 12.1\%$\\
    & SPSA & $0.003 / 0.0\%$ & $0.042 / 61.1\%$ & $0.002 / 0.0\%$ & $0.049 / 64.9\%$ & $0.021 / 39.7\%$ & $0.001 / 2.1\%$ & $0.208 / 78.7\%$ & $0.007 / 9.7\%$ \\
    & $\mathcal{N}$ATTACK & $0.002 / 0.0\%$ & $0.030 / 48.6\%$ & $0.001 / 0.0\%$ & $0.037 / 55.8\%$ & $0.016 / 33.1\%$ & $0.000 / 0.0\%$ & $0.031 / 48.6\%$ & $0.005 / 2.4\%$ \\
    \hline
  \end{tabular}
  \end{center}
  \vspace{-4ex}
   \caption{The point-wise results of the $8$ models on CIFAR-10 against untargeted attacks under the $\ell_\infty$ norm given by the previous evaluation criteria. Each entry shows the median $\ell_\infty$ distance of the minimum adversarial perturbations across all samples (left) as well as the model’s accuracy for the fixed $\epsilon=8/255$ (right).}
   \label{tab:results}
   \vspace{-2ex}
\end{table*}

\subsection{Evaluated Datasets and Algorithms}
\label{sec:details}

\textbf{Datasets:} We use the CIFAR-10~\cite{krizhevsky2009learning} and ImageNet~\cite{russakovsky2015imagenet} datasets to perform adversarial robustness evaluation in this paper. We use the test set containing $10,000$ images of CIFAR-10, and randomly choose $1,000$ images from the ImageNet validation set for evaluation.
For each image, we select a target class uniformly over all other classes except its true class at random, which is used for targeted attacks.

\textbf{Defense Models:}
For fair evaluation, we test $16$ representative defense models whose original source codes and pre-trained models are publicly available. 
These models cover all defense categories and include the state-of-the-art models in each category.
On CIFAR-10, we choose 8 models---naturally trained ResNet-56 (Res-56)~\cite{He2015}, PGD-based adversarial training (PGD-AT)~\cite{madry2017towards}, DeepDefense~\cite{yan2018deep}, TRADES~\cite{zhang2019theoretically}, convex outer polytope (Convex)~\cite{wong2018scaling}, JPEG compression~\cite{dziugaite2016study}, random self-ensemble (RSE)~\cite{liu2018towards}, and adaptive diversity promoting (ADP)~\cite{pang2019improving}.
On ImageNet, we also choose $8$ models---naturally trained Inception v3 (Inc-v3)~\cite{szegedy2016rethinking}, ensemble adversarial training (Ens-AT)~\cite{tramer2017ensemble}, adversarial logit pairing (ALP)~\cite{kannan2018adversarial}, feature denoising (FD)~\cite{xie2019feature}, JPEG compression~\cite{dziugaite2016study}, bit-depth reduction (Bit-Red)~\cite{xu2018feature}, random resizing and padding (R\&P)~\cite{xie2017mitigating}, and RandMix~\cite{zhang2019discretization}.
We use the natural models as the backbone classifiers for defenses based on input transformation (\eg, JPEG). Table~\ref{tab:defense-summary} shows the defense details.
The reason why we choose many \emph{weak} defenses based on randomization or input transformation, which are already broken~\cite{Athalye2018Obfuscated}, is that we want to show their behaviour under various threat models comprehensively, and we indeed draw some findings for these defenses.

\textbf{Attacks:} We implement $15$ typical and widely used attack methods in our benchmark, including $5$ white-box attacks---FGSM, BIM, MIM, DeepFool, and C\&W, $4$ transfer-based attacks---FGSM, BIM, MIM, and DIM, $4$ score-based attacks---ZOO, NES, SPSA, and $\mathcal{N}$ATTACK, and $2$ decision-based attacks---Boundary and Evolutionary. More details of these attacks are outlined in Table~\ref{tab:attack-summary}.
Note that 1) we do not evaluate PGD since PGD and BIM are very similar and often result in similar performance; 2) for transfer-based attacks, we craft adversarial examples by those white-box methods on a substitute model; 3) for defenses that rely on obfuscated gradients, we implement the white-box attacks adaptively by replacing the true gradient with an approximate one when it is unavailable or an expected one when it is random, such that the white-box attacks can identify the worst-case robustness of the models.

\textbf{Platform:} All attacks and defenses are implemented on a new adversarial robustness platform developed by us.
We also conduct the experiments based on the platform.
The comparisons between the existing platforms and ours are detailed in Appendix~\ref{sec:app-a}. 
We acknowledge that many good works are not included in our current benchmark. We hope that our platform could continuously incorporate and evaluate more methods, and be helpful for future works.

\section{Evaluation Results}
\label{sec:exp}
We present the evaluation results on CIFAR-10 in Sec.~\ref{sec:exp-cifar}, and ImageNet in Sec.~\ref{sec:exp-imagenet}. 
Due to the space limitation, we mainly provide the accuracy vs. perturbation budget and attack strength curves of the defense models against untargeted attacks under the $\ell_{\infty}$ norm in this section, and leave the full experimental results (including targeted attacks under the $\ell_{\infty}$ norm, untargeted and targeted attacks under the $\ell_2$ norm, and attack success rate curves) in Appendix~\ref{sec:app-c}.
We also report some key findings in Sec.~\ref{sec:remarks}.

\begin{figure*}[t]
\begin{center}
\includegraphics[width=0.9\linewidth]{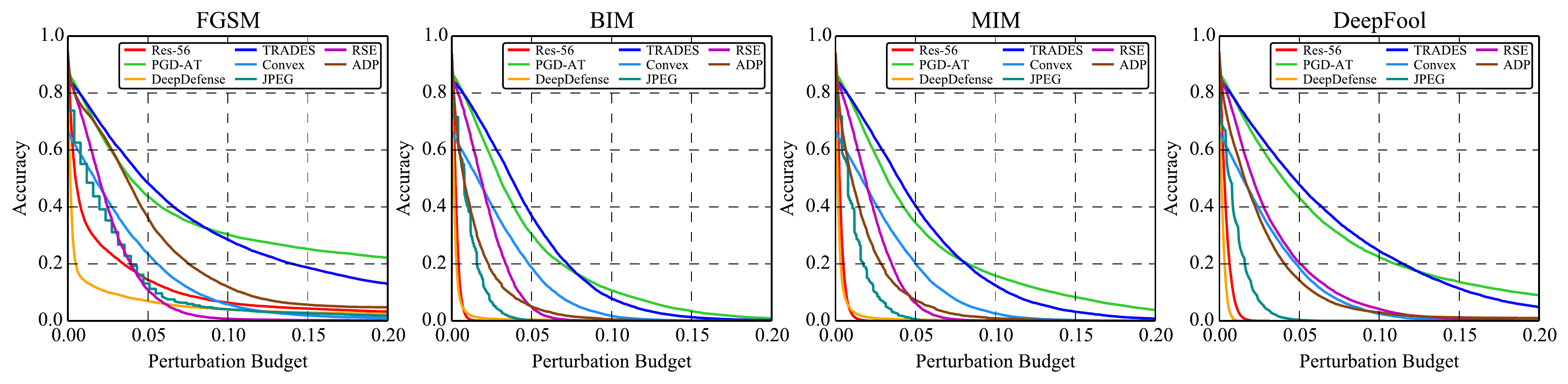}
\end{center}
\vspace{-4ex}
\caption{The \textit{accuracy vs. perturbation budget} curves of the $8$ models on CIFAR-10 against untargeted white-box attacks under the $\ell_{\infty}$ norm.}
\label{fig:white-ut-linf-cifar10-acc-pert}

\begin{center}
\includegraphics[width=0.9\linewidth]{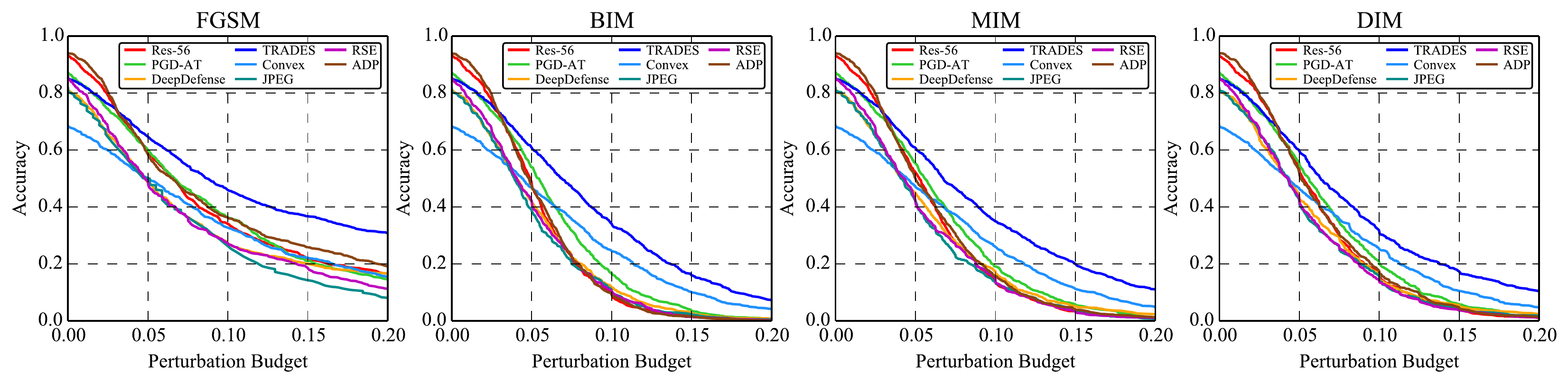}
\end{center}
\vspace{-4ex}
\caption{The \textit{accuracy vs. perturbation budget} curves of the $8$ models on CIFAR-10 against untargeted transfer-based attacks under the $\ell_{\infty}$ norm.}
\label{fig:trans-ut-linf-cifar10-acc-pert}
\vspace{-2ex}
\end{figure*}

\begin{figure*}[t]
\begin{minipage}{.57\linewidth}
\begin{center}
\includegraphics[width=1.0\linewidth]{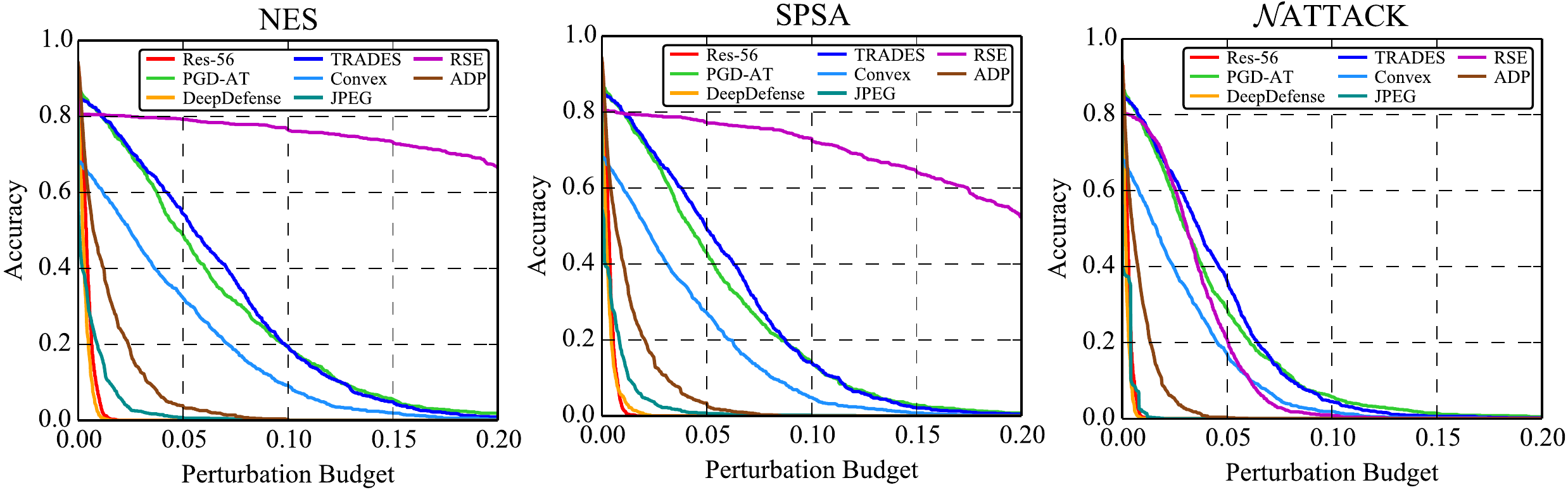}
\end{center}
\vspace{-4ex}
\caption{The \textit{accuracy vs. perturbation budget} curves of the $8$ models on CIFAR-10 against untargeted score-based attacks under the $\ell_{\infty}$ norm.}
\label{fig:score-ut-linf-cifar10-acc-pert}
\end{minipage}
\hspace{1ex}
\begin{minipage}{.41\linewidth}
\begin{center}
\includegraphics[width=0.9\linewidth]{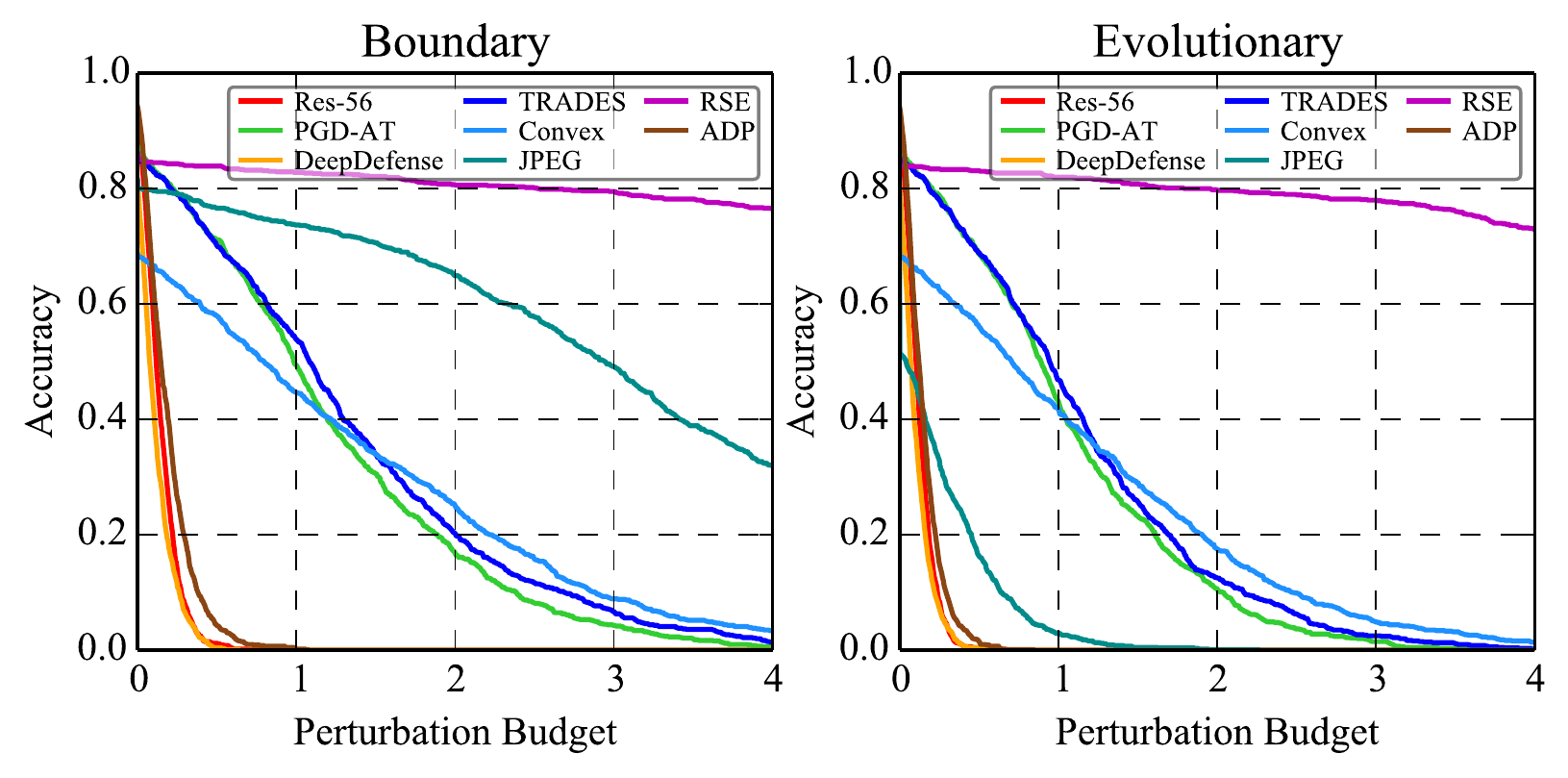}
\end{center}
\vspace{-4ex}
\caption{The \textit{accuracy vs. perturbation budget} curves of the $8$ models on CIFAR-10 against untargeted decision-based attacks under the $\ell_2$ norm.}
\label{fig:decision-ut-l2-cifar10-acc-pert}
\end{minipage}
\begin{minipage}{.57\linewidth}
\begin{center}
\includegraphics[width=1.0\linewidth]{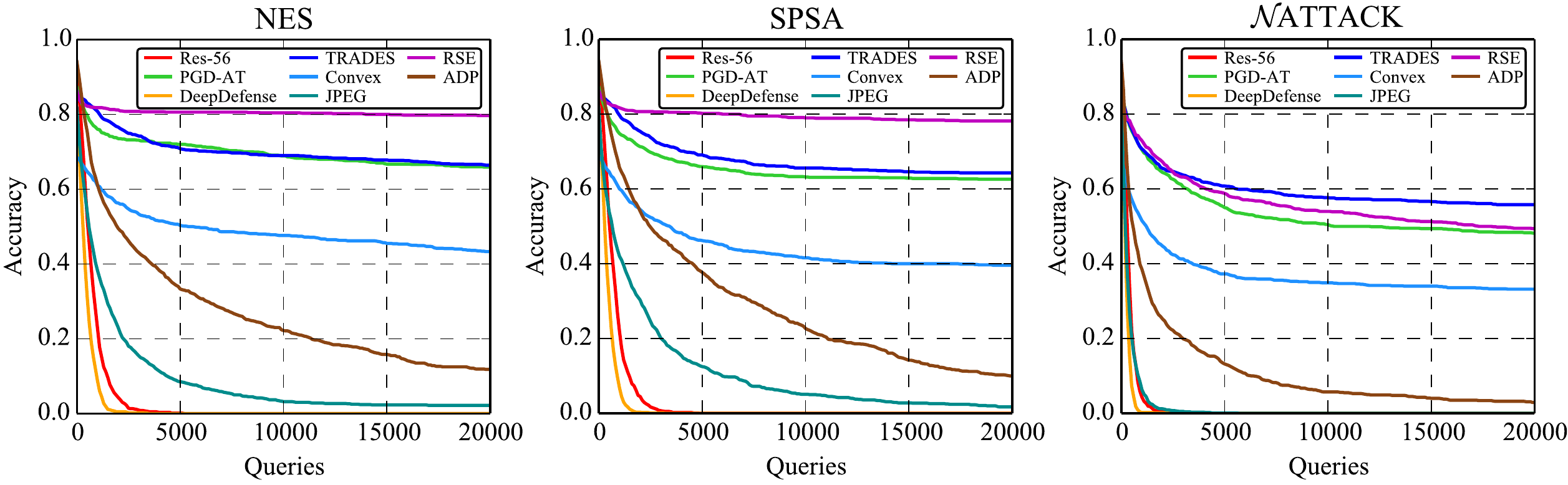}
\end{center}
\vspace{-4ex}
\caption{The \textit{accuracy vs. attack strength} curves of the $8$ models on CIFAR-10 against untargeted score-based attacks under the $\ell_{\infty}$ norm.}
\label{fig:score-ut-linf-cifar10-acc-iter}
\end{minipage}
\hspace{1ex}
\begin{minipage}{.41\linewidth}
\begin{center}
\includegraphics[width=0.9\linewidth]{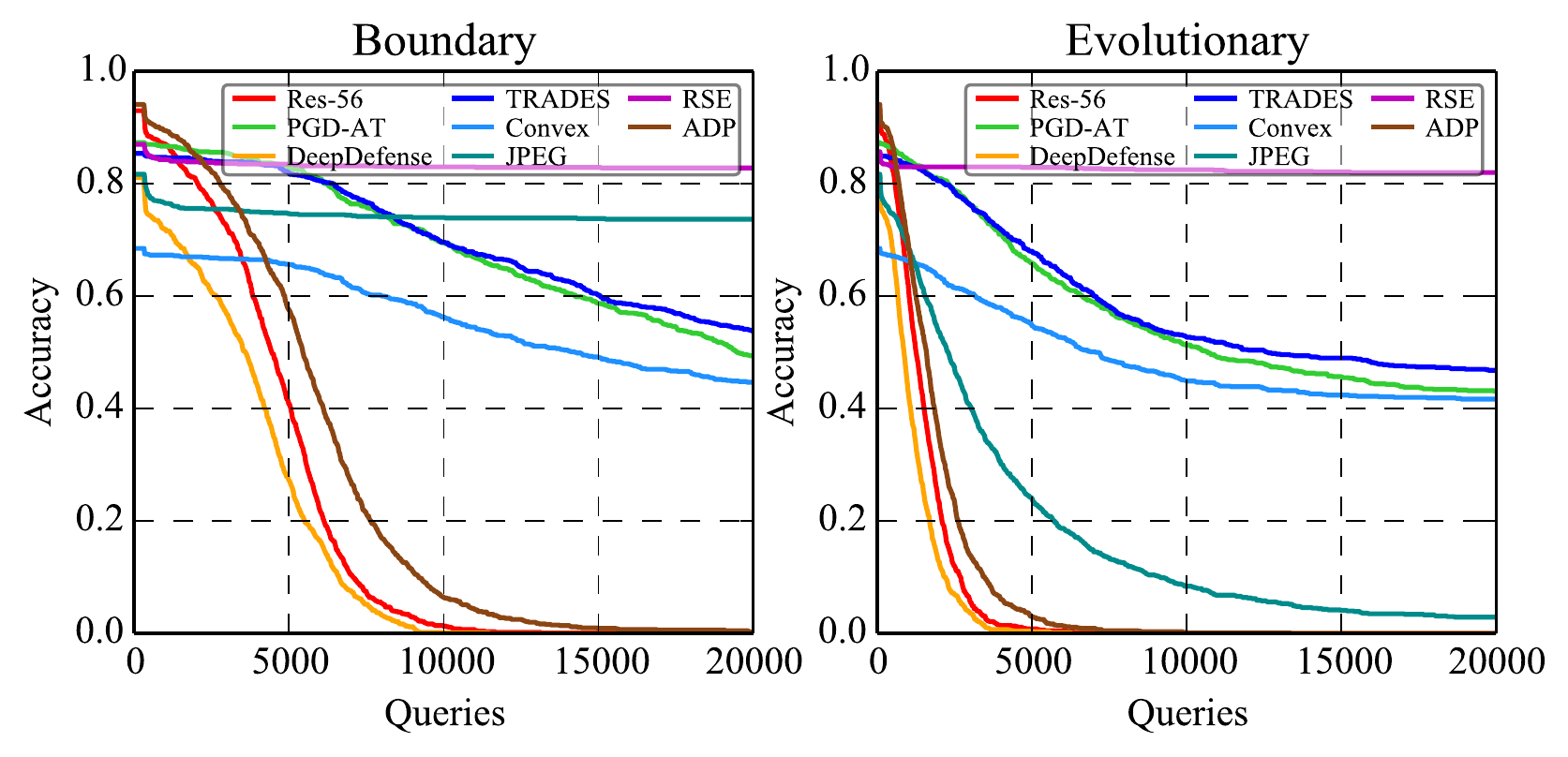}
\end{center}
\vspace{-4ex}
\caption{The \textit{accuracy vs. attack strength} curves of the $8$ models on CIFAR-10 against untargeted decision-based attacks under the $\ell_2$ norm.}
\label{fig:decision-ut-l2-cifar10-acc-iter}
\end{minipage}
\end{figure*}

\begin{figure*}[t]
\begin{center}
\includegraphics[width=0.7\linewidth]{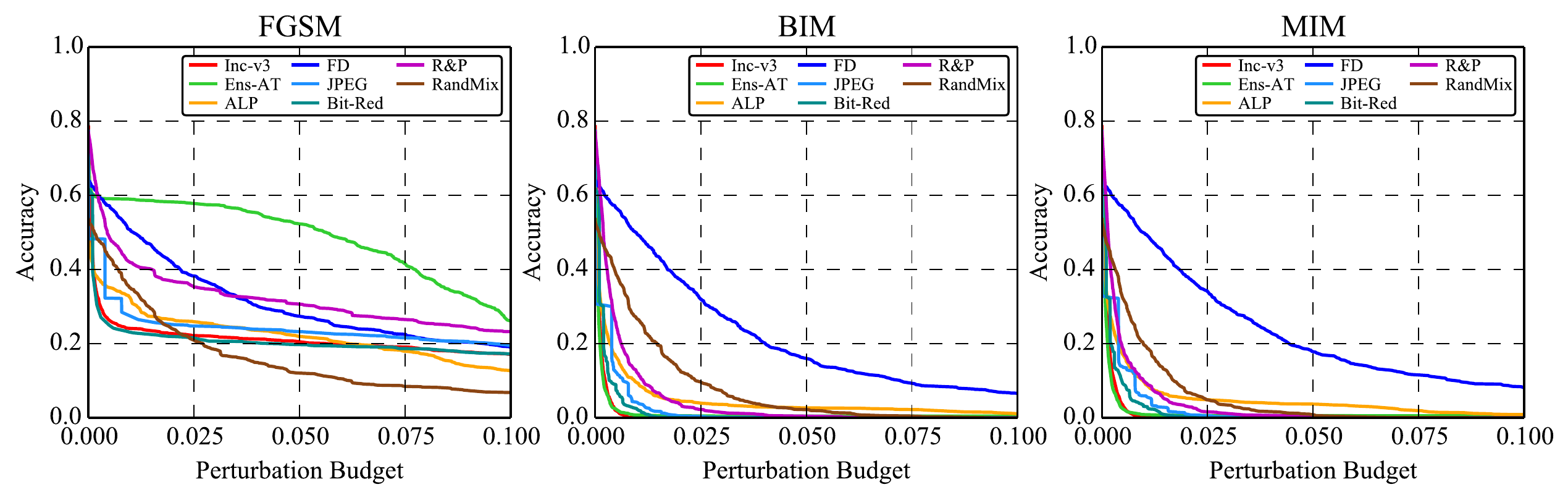}
\end{center}
\vspace{-4ex}
\caption{The \textit{accuracy vs. perturbation budget} curves of the $8$ models on ImageNet against untargeted white-box attacks under the $\ell_{\infty}$ norm.}
\label{fig:white-ut-linf-imagenet-acc-pert}
\begin{center}
\includegraphics[width=0.9\linewidth]{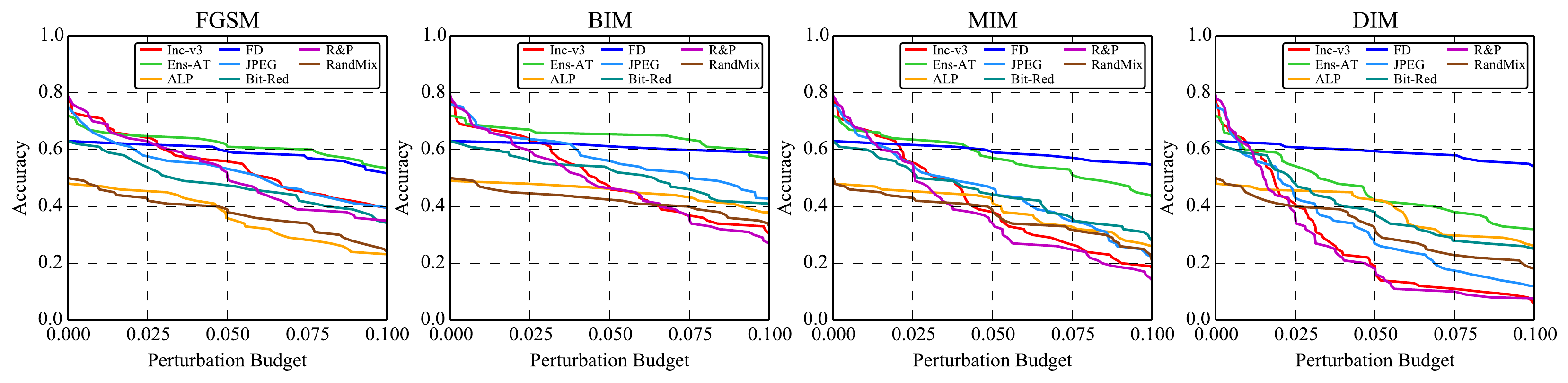}
\end{center}
\vspace{-4ex}
\caption{The \textit{accuracy vs. perturbation budget} curves of the $8$ models on ImageNet against untargeted transfer-based attacks under the $\ell_{\infty}$ norm.}
\label{fig:trans-ut-linf-imagenet-acc-pert}
\vspace{-2ex}
\end{figure*}

\subsection{Evaluation Results on CIFAR-10}
\label{sec:exp-cifar}
In this section, we show the accuracy of the $8$ models on CIFAR-10 against white-box, transfer-based, score-based, and decision-based attacks . To get the \textit{accuracy vs. perturbation budget} curves, we fix the attack strength (\ie, attack iterations or queries) for different budgets. To generate the \textit{accuracy vs. attack strength} curves, we use a fixed perturbation budget as $\epsilon=8/255$ for $\ell_{\infty}$ attacks and $\epsilon=1.0$ for $\ell_2$ attacks, with images in $[0,1]$. The detailed parameters of each attack are provided in Appendix~\ref{sec:app-b}. We let the attack parameters be the same for evaluating all defense models, and leave the study of attack parameters on robustness performance in future works.
To better show the superiority of the robustness curves adopted in this paper compared with the previous evaluation criteria (\ie, the median distance of the minimum adversarial perturbations~\cite{brendel2018adversarial} and the accuracy of a model against an attack for a given perturbation budget~\cite{kurakin2018competation}), we show the evaluation results based on the previous evaluation criteria in Table~\ref{tab:results}.

\textbf{White-box Attacks:} We show the \textit{accuracy vs. perturbation budget} curves of the $8$ models against untargeted FGSM, BIM, MIM, and DeepFool attacks under the $\ell_{\infty}$ norm in Fig.~\ref{fig:white-ut-linf-cifar10-acc-pert} and leave the \textit{accuracy vs. attack strength} curves in Appendix~\ref{sec:app-c}.
The accuracy of the models drops to zero against iterative attacks with the increasing perturbation budget.
Based on the results, we observe that under white-box attacks, the adversarially trained models (\ie, PGD-AT, TRADES) are more robust than other models, because they are trained on the worst-case adversarial examples. We also observe that the relative robustness between two models against an attack could be different under different perturbation budgets or attack iterations (shown in Appendix~\ref{sec:app-c}). For instance, the accuracy of TRADES is higher than that of PGD-AT against white-box attacks when the perturbation budget is small (\eg, $\epsilon=0.05$), but is lower when it is large (\eg, $\epsilon=0.15$).
This finding implies that the comparison between the defense models at a chosen perturbation budget or attack iteration, which is common in previous works, cannot fully demonstrate the performance of a model. But the robustness curves adopted in this paper can better show the global behaviour of these methods, compared with the point-wise evaluation results in Table~\ref{tab:results}.

\textbf{Transfer-based Black-box Attacks:}
We show the \textit{accuracy vs. perturbation budget} curves of the $8$ models against untargeted transfer-based FGSM, BIM, MIM, and DIM attacks under the $\ell_{\infty}$ norm in Fig.~\ref{fig:trans-ut-linf-cifar10-acc-pert}, and leave the \textit{accuracy vs. attack strength} curves in Appendix~\ref{sec:app-c}. In this experiment, we choose TRADES as the substitute model to attack the others, and use PGD-AT to attack TRADES, since these two models demonstrate superior white-box robustness compared with the other models, and thus the adversarial examples generated on the other models can rarely transfer to TRADES and PGD-AT.
From the results, the accuracy of the defenses also drops with the increasing perturbation budget. We also observe that the recent attacks (\eg, MIM, DIM) for improving the transferability do not actually perform better than the baseline BIM method.

\textbf{Score-based Black-box Attacks:} We show the curves of the \textit{accuracy vs. perturbation budget} and \textit{accuracy vs. attack strength (queries)} of the $8$ models against untargeted score-based NES, SPSA, and $\mathcal{N}$ATTACK under the $\ell_{\infty}$ norm in Fig.~\ref{fig:score-ut-linf-cifar10-acc-pert} and Fig.~\ref{fig:score-ut-linf-cifar10-acc-iter}.
We set the maximum number of queries as $20,000$ in these attack methods. The accuracy of the defenses also decreases along with the increasing perturbation budget or the number of queries.
$\mathcal{N}$ATTACK is more effective as can be seen from the figures.
From the results, we notice that RSE is quite resistant to score-based attacks, especially NES and SPSA. We think that the randomness of the predictions given by RSE makes the estimated gradients of NES and SPSA useless for attacks.

\textbf{Decision-based Black-box Attacks:} Since the decision-based Boundary and Evolutionary attack methods can be only used for $\ell_2$ attacks, we present the accuracy curves of the $8$ models against untargeted Boundary and Evolutionary attacks under the $\ell_2$ norm in Fig.~\ref{fig:decision-ut-l2-cifar10-acc-pert} and Fig.~\ref{fig:decision-ut-l2-cifar10-acc-iter}.
The behaviour of the defenses is similar to that of the score-based attacks.
It can be observed that RSE is also resistant to decision-based attacks compared with the other defenses due to the randomness of the predictions.


\begin{figure*}[t]
\begin{minipage}{.57\linewidth}
\begin{center}
\includegraphics[width=1.0\linewidth]{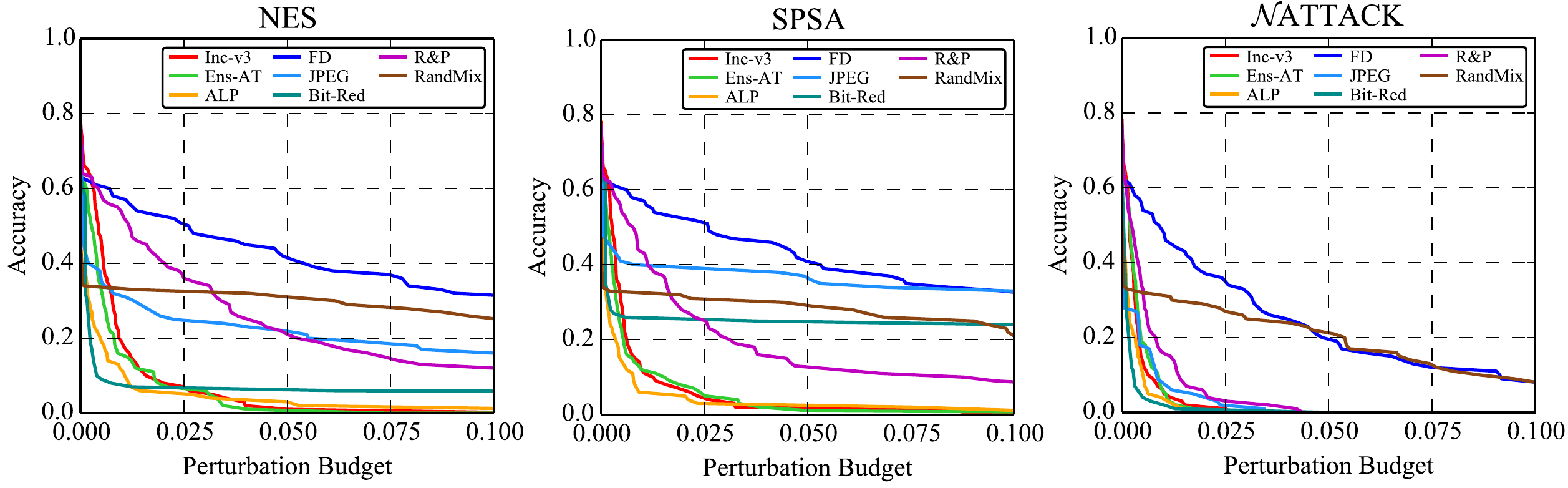}
\end{center}
\vspace{-4ex}
\caption{The \textit{accuracy vs. perturbation budget} curves of the $8$ models on ImageNet against untargeted score-based attacks under the $\ell_{\infty}$ norm.}
\label{fig:score-ut-linf-imagenet-acc-pert}
\end{minipage}
\hspace{1ex}
\begin{minipage}{.42\linewidth}
\begin{center}
\includegraphics[width=0.92\linewidth]{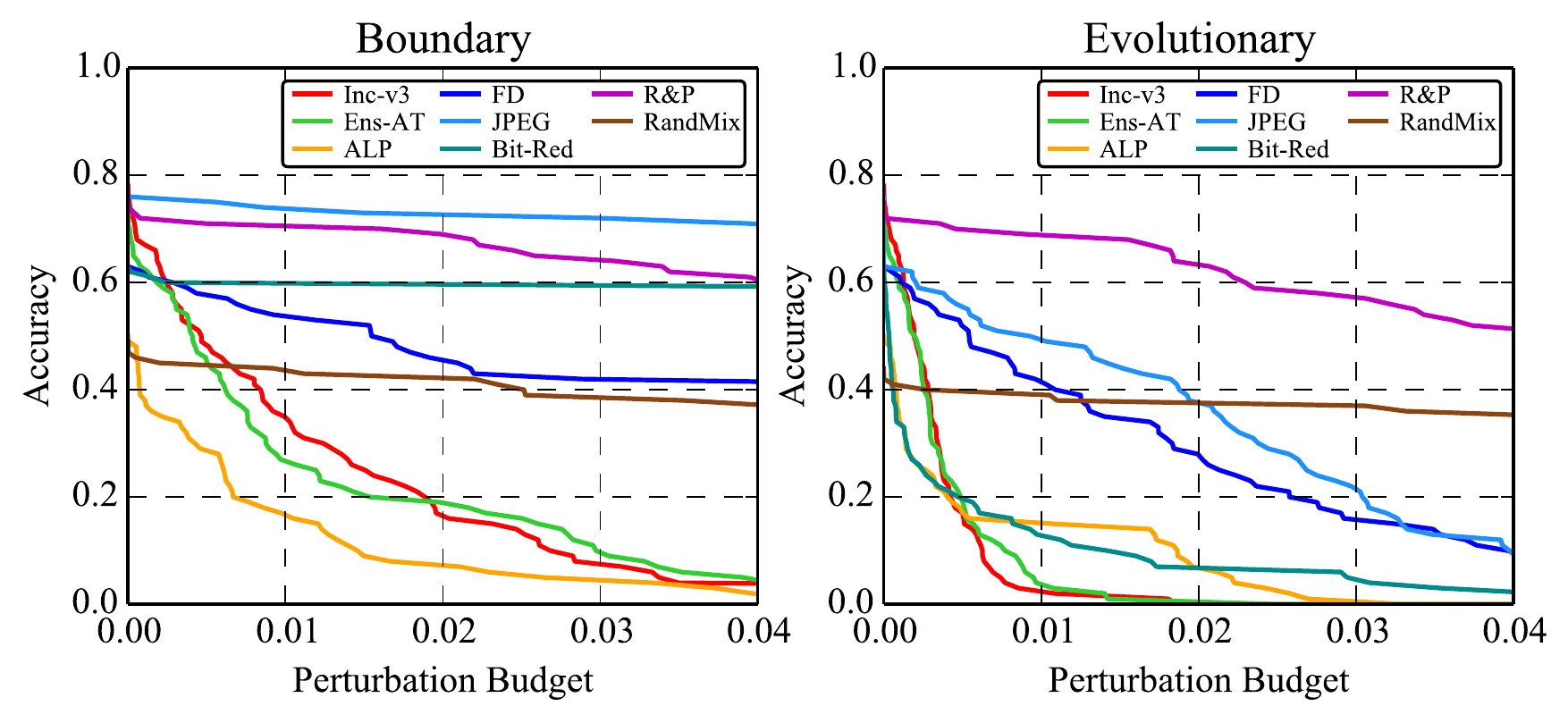}
\end{center}
\vspace{-4ex}
\caption{The \textit{accuracy vs. perturbation budget} curves of the $8$ models on ImageNet against untargeted decision-based attacks under the $\ell_2$ norm.}
\label{fig:decision-ut-l2-imagenet-acc-pert}
\end{minipage}
\begin{minipage}{.57\linewidth}
\begin{center}
\includegraphics[width=1.0\linewidth]{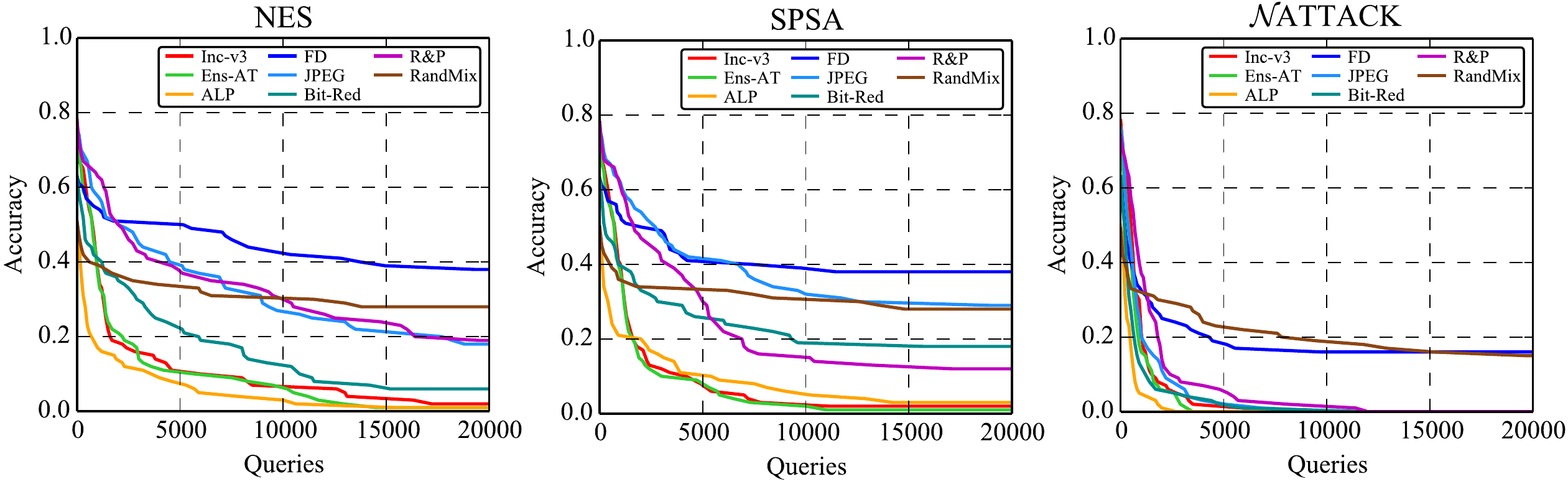}
\end{center}
\vspace{-4ex}
\caption{The \textit{accuracy vs. attack strength} curves of the $8$ models on ImageNet against untargeted score-based attacks under the $\ell_{\infty}$ norm.}
\label{fig:score-ut-linf-imagenet-acc-iter}
\end{minipage}
\hspace{1ex}
\begin{minipage}{.42\linewidth}
\begin{center}
\includegraphics[width=0.92\linewidth]{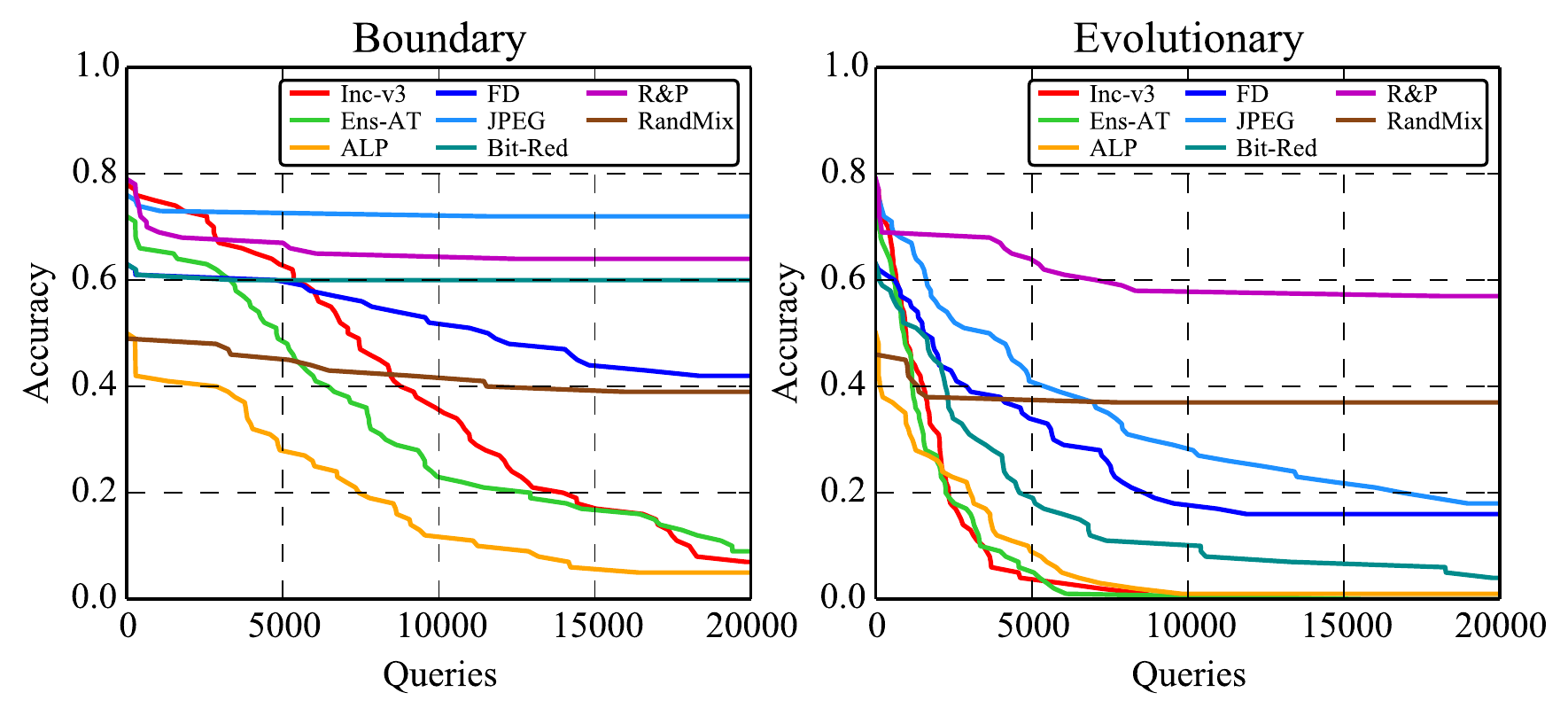}
\end{center}
\vspace{-4ex}
\caption{The \textit{accuracy vs. attack strength} curves of the $8$ models on ImageNet against untargeted decision-based attacks under the $\ell_2$ norm.}
\label{fig:decision-ut-l2-imagenet-acc-iter}
\end{minipage}
\vspace{-2ex}
\end{figure*}

\subsection{Evaluation Results on ImageNet}
\label{sec:exp-imagenet}

We present the experimental results on ImageNet in this section. We use the same settings with those on CIFAR-10 to get the evaluation curves. Since the input image size is different for the ImageNet defenses, we adopt the normalized $\ell_2$ distance defined as $\bar{\ell}_2(\bm{a}) = \nicefrac{\|\bm{a}\|_2}{\sqrt{d}}$ as the measurement for $\ell_2$ attacks, where $d$ is the dimension of a vector $\bm{a}$. To get the \textit{accuracy (attack success rate) vs. attack strength} curves, we fix the perturbation budget as $\epsilon=16/255$ for $\ell_{\infty}$ attacks and $\epsilon=\sqrt{0.001}$ for $\ell_2$ attacks.

\textbf{White-box Attacks:}
We show the \textit{accuracy vs. perturbation budget} curves of the $8$ models on ImageNet against untargeted FGSM, BIM, and MIM under the $\ell_{\infty}$ norm in Fig.~\ref{fig:white-ut-linf-imagenet-acc-pert}. We also leave the \textit{accuracy vs. attack strength} curves in Appendix~\ref{sec:app-c}. We find that FD exhibits superior performance over all other models. FD is also trained by the adversarial training method in~\cite{madry2017towards}, demonstrating the effectiveness of PGD-based adversarial training on ImageNet.

\textbf{Transfer-based Black-box Attacks:} We use a ResNet-152 model~\cite{He2015} as the substitute model. The \textit{accuracy vs. perturbation budget} curves of the defenses against untargeted transfer-based FGSM, BIM, MIM, and DIM under the $\ell_{\infty}$ norm are shown in Fig.~\ref{fig:trans-ut-linf-imagenet-acc-pert}.
Different from the results on CIFAR-10, MIM and DIM improve the transferability of adversarial examples over FGSM and BIM, resulting in lower accuracy of the black-box models. A potential reason is that the image size of ImageNet is much larger, and the adversarial examples generated by BIM can ``overfit'' the substitute model~\cite{Dong2017}, making them difficult to transfer to other black-box models.

\textbf{Score-based and Decision-based Attacks:} Fig.~\ref{fig:score-ut-linf-imagenet-acc-pert} and Fig.~\ref{fig:score-ut-linf-imagenet-acc-iter} show the \textit{accuracy vs. perturbation budget} and \textit{accuracy vs. attack strength (queries)} curves of the defense models on ImageNet against untargeted score-based attacks under the $\ell_{\infty}$ norm, while Fig.~\ref{fig:decision-ut-l2-imagenet-acc-pert} and Fig.~\ref{fig:decision-ut-l2-imagenet-acc-iter} show the two sets of curves for untargeted decision-based attacks under the $\ell_2$ norm. Similar to the results on CIFAR-10, we find that the two defenses based on randomization, \ie, R\&P and RandMix, have higher accuracy than the other methods in most cases. JPEG and Bit-Red that are based on input transformations also improve the robustness over the baseline model (\ie, Inc-v3).

\subsection{Discussions}
\label{sec:remarks}

Based on the above results and more results in Appendix~\ref{sec:app-c}, we highlight some key findings.

First, the relative robustness between defenses against the same attack could be different under varying attack parameters, such as the perturbation budget or the number of attack iterations. Not only the results of PGD-AT and TRADES in Fig.~\ref{fig:white-ut-linf-cifar10-acc-pert} can prove it, but also the results in many different scenarios show the similar phenomenon. Given this observation, the comparison between defenses at a specific attack configuration cannot fully demonstrate the superiority of a method upon another. We therefore strongly \textit{advise the researchers to adopt the robustness curves as the major evaluation metrics to present the robustness results.}

Second, among the defenses studied in this paper, we find that the most robust models are obtained by PGD-based adversarial training. Their robustness not only is good for the threat model under which they are trained (\ie, the $\ell_{\infty}$ threat model), but can also generalize to other threat models (\eg, the $\ell_2$ threat model).
However, adversarial training usually leads to a reduction of natural accuracy and high training cost. A research direction is to develop new methods that maintain the natural accuracy or reduce the training cost. And we have seen several works~\cite{shafahi2019adversarial} in this direction.
    
Third, we observe that the defenses based on randomization are quite resistant to score-based and decision-based attacks, which rely on the query feedback of the black-box models. We argue that the robustness of the randomization-based defenses against these attacks is due to the random predictions given by the models, making the estimated gradients or search directions unreliable for attacks. A potential research direction is to develop more powerful score-based and decision-based attacks that can efficiently evade the randomization-based defenses.

Fourth, the defenses based on input transformations (\eg, JPEG, Bit-Red) sightly improve the robustness over undefended ones, and sometimes get much higher accuracy against score-based and decision-based attacks. Since these methods are quite simple, they may be combined with other types of defenses to build more powerful defenses.

Fifth, we find that different transfer-based attack methods exhibit similar performance on CIFAR-10, while the recent methods (\eg, MIM, DIM) can improve the transferability of adversarial examples over BIM on ImageNet. One potential reason is that the input dimension of the models on ImageNet is much higher than that on CIFAR-10, and thus the adversarial examples generated by BIM can easily ``overfit'' the substitute model~\cite{Dong2017}, resulting in poor transferability. The recent methods proposed to solve this issue can generate more transferable adversarial examples.

Note that these findings are based on our current benchmark, which may be strengthened or falsified in the future if new results are given.

\section{Conclusion}
In this paper, we established a comprehensive, rigorous, and coherent benchmark to evaluate adversarial robustness of image classifiers. We performed large-scale experiments with two robustness curves as the fair-minded evaluation criteria to facilitate a better understanding of the representative and state-of-the-art adversarial attack and defense methods. We drew some key findings based on the evaluation results, which may be helpful for future research.

{\small
\bibliographystyle{ieee_fullname}
\bibliography{egbib}
}

\clearpage
\noindent \begin{center} {\large  \textbf{Appendix}} \end{center}
\appendix

\begin{table*}[t]
  \begin{center}
  \begin{tabular}{c|c||c|c}
    \hline
    \multicolumn{2}{c||}{CIFAR-10~\cite{krizhevsky2009learning}} & \multicolumn{2}{c}{ImageNet~\cite{russakovsky2015imagenet}} \\
    \hline
    Defense Model  & Architecture & Defense Model  & Architecture \\
    \hline\hline
    Res-56~\cite{He2015}  & ResNet-56 & Inc-v3~\cite{szegedy2016rethinking}  & Inception v3\\
    \hline
    PGD-AT~\cite{madry2017towards} & Wide ResNet-34-10 &
    Ens-AT~\cite{tramer2017ensemble} & Inception v3\\
    \hline
    DeepDefense~\cite{yan2018deep} &  5-layer CNN & 
    ALP~\cite{kannan2018adversarial}  & ResNet-50 \\
    \hline
    TRADES~\cite{zhang2019theoretically}  & Wide ResNet-34-10 &
    FD~\cite{xie2019feature} & ResNet-152 with denoising layers \\
    \hline 
    Convex~\cite{wong2018scaling} & ResNet & JPEG~\cite{dziugaite2016study} & Inception v3 \\
    \hline
    JPEG~\cite{dziugaite2016study} & ResNet-56 & Bit-Red~\cite{xu2018feature} &  Inception v3 \\
    \hline
    RSE~\cite{liu2018towards} & VGG & R\&P~\cite{xie2017mitigating} &  Inception v3 \\
    \hline
     ADP~\cite{pang2019improving}  & ResNet-110 $\times 3$ & RandMix~\cite{zhang2019discretization} & Inception v3 \\
    \hline
  \end{tabular}
  \end{center}
  \vspace{-2ex}
  \caption{We show the network architecture of each defense model. Defenses based on input transformations use the baseline natural models as the backbone classifiers. DeepDefense uses a very simple 5-layer CNN. FD modifies a ResNet-152 architecture with the proposed denoising layers. ADP ensembles the predictions of 3 ResNet-110 models. Convex uses a ResNet model with architecture provided in~\cite{wong2018scaling}.}
  \label{tab:defense-arch}
\end{table*}

\section{Adversarial Robustness Platforms}
\label{sec:app-a}
There are several public platforms for adversarial machine learning, including CleverHans~\cite{papernot2016technical}, Foolbox~\cite{rauber2017foolbox}, ART~\cite{nicolae2018adversarial}, \etc.
However, we observe that these platforms do not totally support our comprehensive evaluations in this paper. First, some attacks evaluated in this paper are not included in these platforms. There are less than $10$ out of the $15$ attacks adopted in this paper that are already implemented in each platform. And most of the available methods are white-box methods. Second, although these platforms incorporate a few defenses, they do not use the pre-trained models. But we use the original source codes and pre-trained models to perform unbiased evaluations.
Third, the evaluation metrics defined by the two robustness curves in this paper are not provided in the existing platforms.
Therefore, we develop a new adversarial robustness platform to satisfy our requirements.

Another similar work to ours is DeepSec~\cite{ling2019deepsec}, which also provides a uniform platform for adversarial robustness evaluation of DL models. However, as argued in~\cite{carlini2019critique}, DeepSec has several flaws, including 1) it evaluates the defenses by using the adversarial examples generated against undefended models; 2) it has some incorrect implementations; 3) it evaluates the robustness of the defenses as an average, \etc. We try our best to avoid these issues in this paper. Our work differs from DeepSec in three main aspects: 1) we consider complete threat models and use various attack methods in different settings; 2) we use the original source codes and pre-trained models provided by the authors to prevent implementation errors; 3) we adopt two complementary robustness curves as the fair-minded evaluation metrics to present the results.
We think that our evaluations can truly reflect the behavior of the attack and defense methods, and provide us with a detailed understanding of these methods.

\section{Evaluation Details}
\label{sec:app-b}

In this section, we provide additional evaluation details. Table~\ref{tab:defense-arch} shows the network architecture of each defense model. Below we show the details of the attack methods
as well as their parameters in our experiments. For clarity, we only introduce the untargeted attacks.

\textbf{FGSM}~\cite{goodfellow2014explaining} generates an untargeted adversarial example under the $\ell_{\infty}$ norm as 
\begin{equation}
    \bm{x}^{adv} = \bm{x} + \epsilon\cdot\mathrm{sign}(\nabla_{\bm{x}}\mathcal{J}(\bm{x},y)),
\end{equation}
where $\mathcal{J}$ is the cross-entropy loss. It can be extended to an $\ell_2$ attack as  
\begin{equation}
    \bm{x}^{adv} = \bm{x} + \epsilon\cdot\frac{\nabla_{\bm{x}}\mathcal{J}(\bm{x},y)}{\|\nabla_{\bm{x}}\mathcal{J}(\bm{x},y)\|_2}.
\end{equation}
To get the accuracy (attack success rate) vs. perturbation budget curves, we perform a line search followed by a binary search on $\epsilon$ to find its minimum value.

\textbf{BIM}~\cite{Kurakin2016} extends FGSM by iteratively taking multiple small gradient updates as
\begin{equation}
\label{eq:bim}
    \bm{x}_{t+1}^{adv} = \mathrm{clip}_{\bm{x},\epsilon} \big(\bm{x}_t^{adv} + \alpha\cdot\mathrm{sign}(\nabla_{\bm{x}}\mathcal{J}(\bm{x}_t^{adv},y))\big),
\end{equation}
where $\mathrm{clip}_{\bm{x},\epsilon}$ projects the adversarial example to satisfy the $\ell_{\infty}$ constrain and $\alpha$ is the step size. It can also be extended to an $\ell_2$ attack similar to FGSM. For most experiments, we set $\alpha = 0.15 \cdot \epsilon$. 
To get the accuracy (attack success rate) vs. perturbation budget curves, we also perform a binary search on $\epsilon$. For each $\epsilon$ during the binary search, we set the number of iterations as $20$ in white-box attacks and $10$ in transfer-based attacks.

\textbf{MIM}~\cite{Dong2017} integrates a momentum term into BIM as 
\begin{equation}
    \bm{g}_{t+1} = \mu \cdot \bm{g}_t + \frac{\nabla_{\bm{x}}\mathcal{J}(\bm{x}_t^{adv},y)}{\|\nabla_{\bm{x}}\mathcal{J}(\bm{x}_t^{adv},y)\|_1};
\end{equation}
\begin{equation}
    \bm{x}^{adv}_{t+1}=\mathrm{clip}_{\bm{x},\epsilon}(\bm{x}^{adv}_t+\alpha\cdot\mathrm{sign}(\bm{g}_{t+1})).
\end{equation}
MIM can similarly be extended to the $\ell_2$ case. We set the step size $\alpha$ and the number of iterations identical to those in BIM. We set the decay factor as $\mu=1.0$.

\textbf{DeepFool}~\cite{Moosavidezfooli2016} is also an iterative attack method, which generates an adversarial example on the decision boundary of a classifier with the minimum perturbation. We set the maximum number of iterations as $100$ in DeepFool, and it will early stop when the solution at an intermediate iteration is already adversarial.

\textbf{C\&W}~\cite{carlini2017towards} is an optimization-based attack method, which generates an $\ell_2$ adversarial example by solving
\begin{equation}
\begin{aligned}
\label{eq:cw}
    \bm{x}^{adv} = & \argmin_{\bm{x}'} \big\{\|\bm{x}'-\bm{x}\|_2^2 \\+ & c\cdot\max(Z(\bm{x}')_y - \max_{i\neq y}Z(\bm{x}')_i, 0)\big\},
\end{aligned}
\end{equation}
where $Z(\bm{x}')$ is the logit output of the classifier and $c$ is a constant. This optimization problem is solved by an Adam~\cite{Kingma2014} optimizer. $c$ is found by binary search. To get the accuracy (attack success rate) vs. perturbation budget curves, we optimize Eq.~(\ref{eq:cw}) for $100$ iterations. To get the accuracy (attack success rate) vs. attack strength curves, we optimize Eq.~(\ref{eq:cw}) for $10$, $20$, $30$, $40$ iterations on CIFAR-10, and $10$, $20$ iterations on ImageNet to show the results.

\textbf{DIM}~\cite{xie2019improving} randomly resizes and pads the input, and uses the transformed input for gradient calculation. It also adopts the momentum technique. In our experiments, we set the common parameters the same as those of MIM. For its own parameters, we set the input $\bm{x}\in \mathbb{R}^{s\times s \times 3}$ is first resized to a $rnd\times rnd \times 3$ image, with $rnd \in [0.9 * s, s]$, and then padded to the original size.

\textbf{ZOO}~\cite{chen2017zoo} has been proposed to optimize Eq.~(\ref{eq:cw}) in the black-box manner through queries. It estimates the gradient at each coordinate as 
\begin{equation}
    \hat{g}_i = \frac{\mathcal{L}(\bm{x}+\sigma \bm{e}_i,y)-\mathcal{L}(\bm{x}-\sigma \bm{e}_i,y)}{2\sigma} \approx \frac{\partial \mathcal{L}(\bm{x},y)}{\partial x_i},
\end{equation}
where $\mathcal{L}$ is the objective in Eq.~(\ref{eq:cw}), $\sigma$ is a small constant, and $\bm{e}_i$ is the $i$-th unit basis vector. In our experiments, we perform one update with $\hat{g}_i$ at one randomly sampled coordinate. We set $\sigma=10^{-4}$.

\textbf{NES}~\cite{ilyas2018black} and \textbf{SPSA}~\cite{uesato2018adversarial} adopt the update rule in Eq.~(\ref{eq:bim}) for adversarial example generation. Although the true gradient is unavailable, NES and SPSA give the full gradient estimation as
\begin{equation}
    \bm{\hat{g}} = \frac{1}{q}\sum_{i=1}^{q}\frac{\mathcal{J}(\bm{x}+\sigma \bm{u}_i,y)-\mathcal{J}(\bm{x}-\sigma\bm{u}_i,y)}{2\sigma}\cdot \bm{u}_i,
\end{equation}
where we use $\mathcal{J}(\bm{x}, y) = Z(\bm{x})_y - \max_{i\neq y}Z(\bm{x})_i$ instead of the cross-entropy loss, $\{\bm{u}_i\}_{i=1}^q$ are the random vectors sampled from a Gaussian distribution in NES, and a Rademacher distribution in SPSA. We set $\sigma=0.001$ and $q=100$ in experiments.

$\mathcal{N}$\textbf{ATTACK}~\cite{li2019nattack} does not estimate the gradient but learns a Gaussian distribution centered around the input such that a sample drawn from it is likely an adversarial example. We set the sampling variance as $0.1$, the learning rate as $0.02$, the number of samples per iteration as $100$ in $\mathcal{N}$ATTACK.

The decision-based black-box attacks---\textbf{Boundary}~\cite{Brendel2018Decision} and \textbf{Evolutionary}~\cite{dong2019efficient} rely on heuristic search on the decision boundary. They need a starting point, which is already adversarial, to initialize an attack. For untargeted attacks, we sample each pixel of the initial image from a uniform distribution. For targeted attacks, we specify the starting point as a sample that is classified by the model as the target class.
We use the default hyperparameters of these two attacks given by their authors.

\section{Full Evaluation Results}
\label{sec:app-c}
We provide the full evaluation results in this section.
\subsection{Full Evaluation Results on CIFAR-10}

\begin{figure*}[t]
\begin{center}
\includegraphics[width=0.85\linewidth]{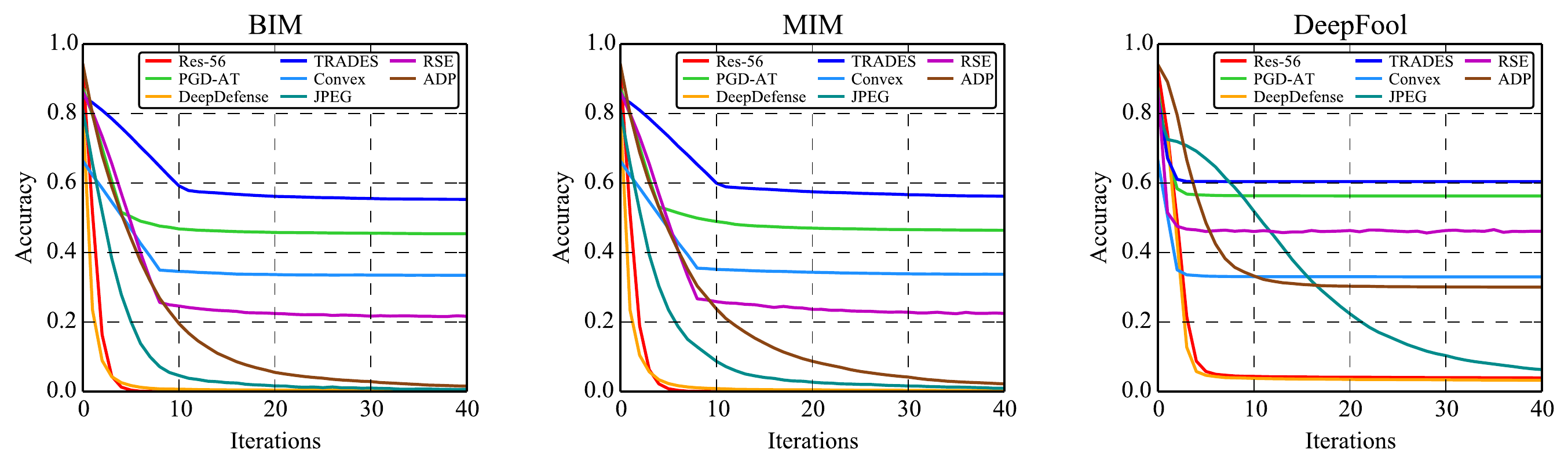}
\end{center}
\vspace{-2ex}
\caption{The \textit{accuracy vs. attack strength} curves of the $8$ models on CIFAR-10 against untargeted white-box attacks under the $\ell_{\infty}$ norm.}
\label{fig:white-ut-linf-cifar10-acc-iter}
\end{figure*}

\begin{figure*}[t]
\begin{center}
\includegraphics[width=0.85\linewidth]{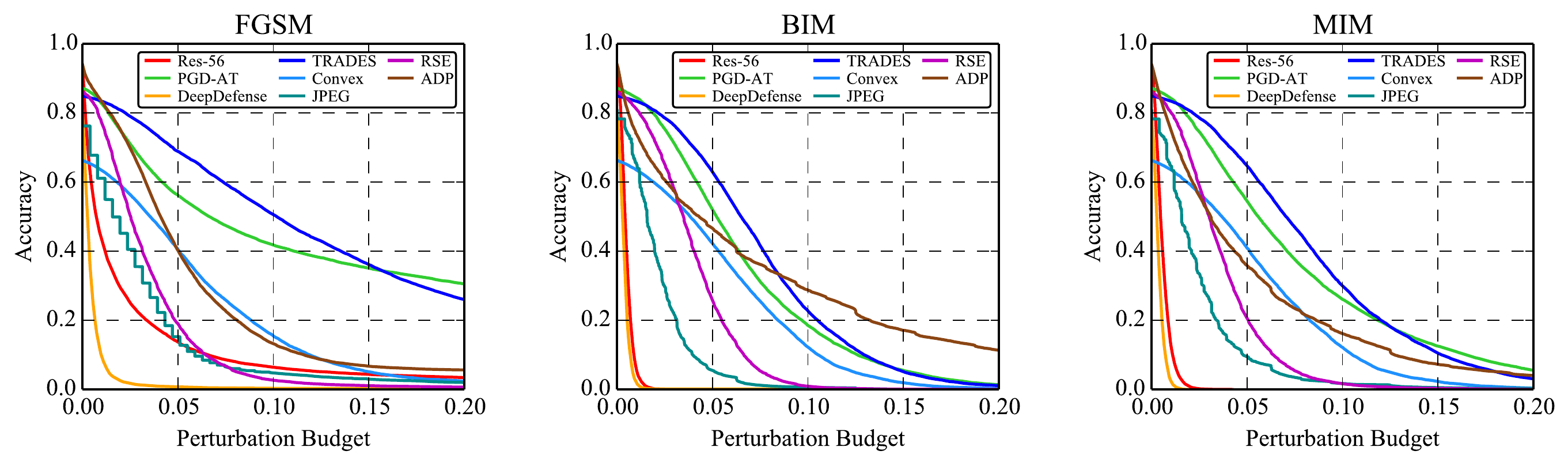}
\end{center}
\vspace{-2ex}
\caption{The \textit{accuracy vs. perturbation budget} curves of the $8$ models on CIFAR-10 against targeted white-box attacks under the $\ell_{\infty}$ norm.}
\label{fig:white-t-linf-cifar10-acc-pert}
\begin{center}
\includegraphics[width=0.55\linewidth]{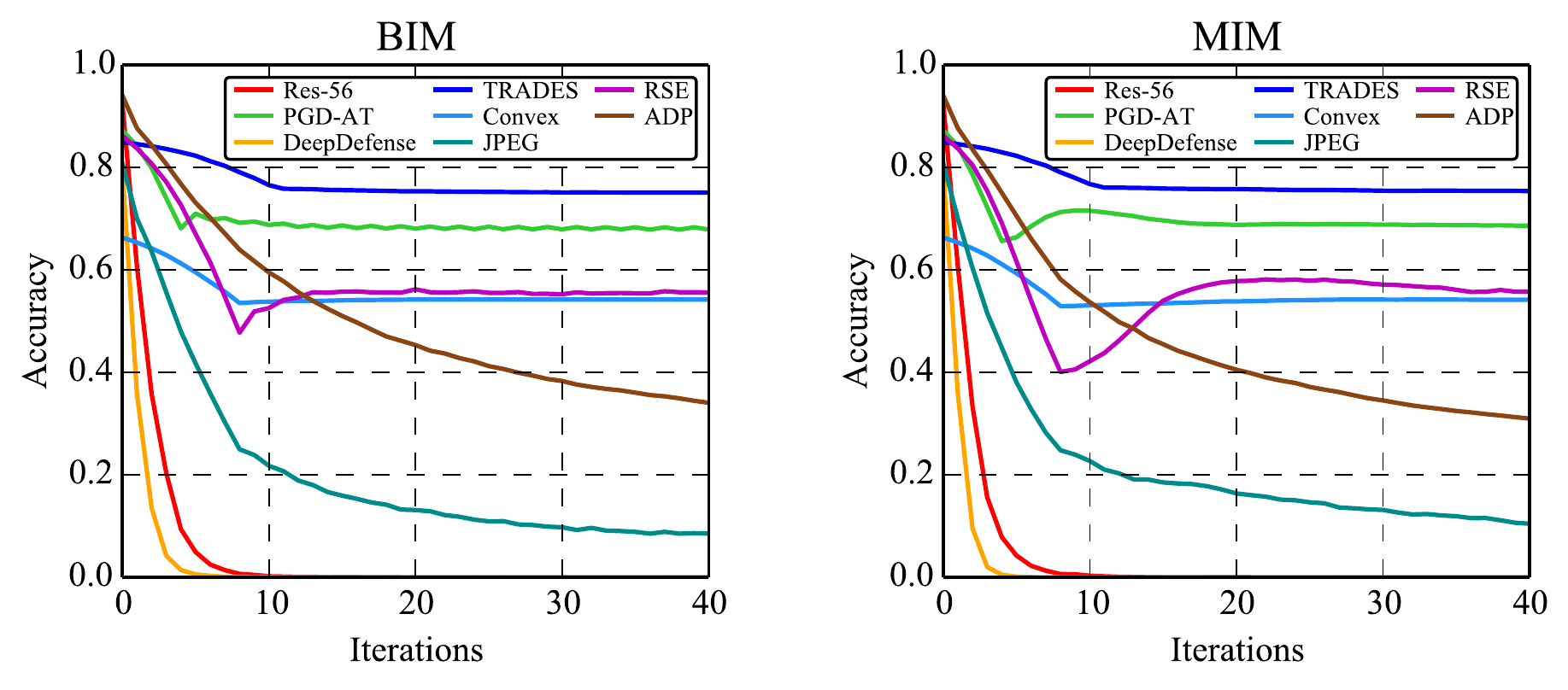}
\end{center}
\vspace{-2ex}
\caption{The \textit{accuracy vs. attack strength} curves of the $8$ models on CIFAR-10 against targeted white-box attacks under the $\ell_{\infty}$ norm.}
\label{fig:white-t-linf-cifar10-acc-iter}
\end{figure*}

\begin{figure*}[t]
\begin{center}
\includegraphics[width=0.85\linewidth]{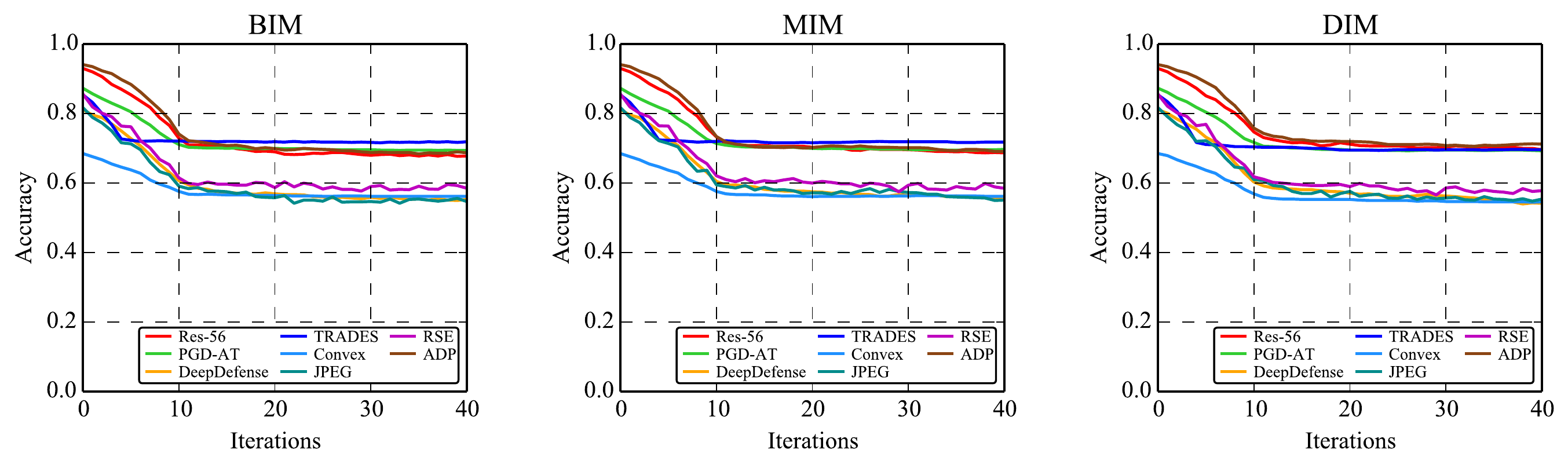}
\end{center}
\vspace{-2ex}
\caption{The \textit{accuracy vs. attack strength} curves of the $8$ models on CIFAR-10 against untargeted transfer-based attacks under the $\ell_{\infty}$ norm.}
\label{fig:trans-ut-linf-cifar10-acc-iter}
\end{figure*}

\begin{figure*}[t]
\begin{center}
\includegraphics[width=1.0\linewidth]{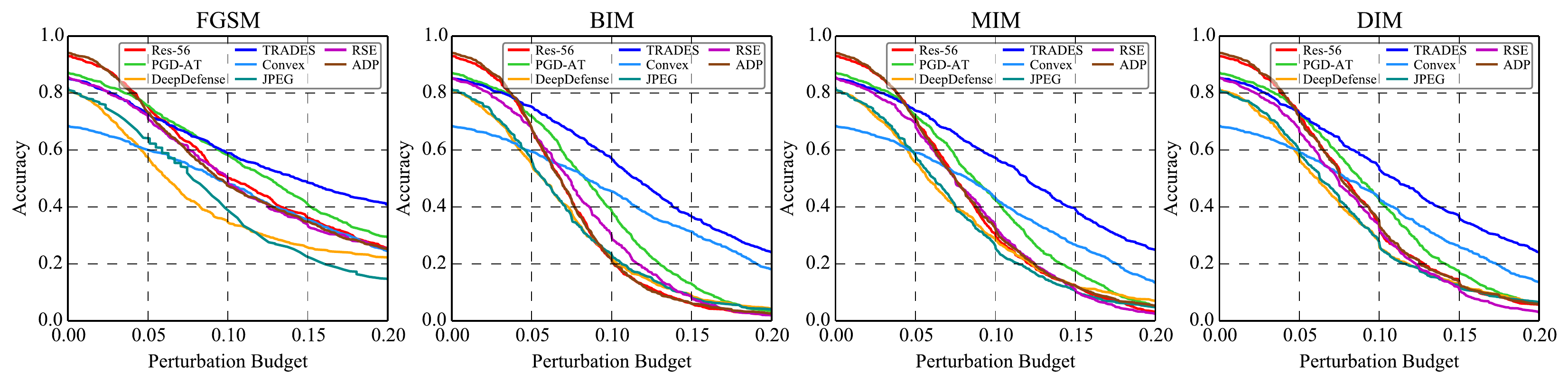}
\end{center}
\vspace{-2ex}
\caption{The \textit{accuracy vs. perturbation budget} curves of the $8$ models on CIFAR-10 against targeted transfer-based attacks under the $\ell_{\infty}$ norm.}
\label{fig:trans-t-linf-cifar10-acc-pert}
\begin{center}
\includegraphics[width=0.85\linewidth]{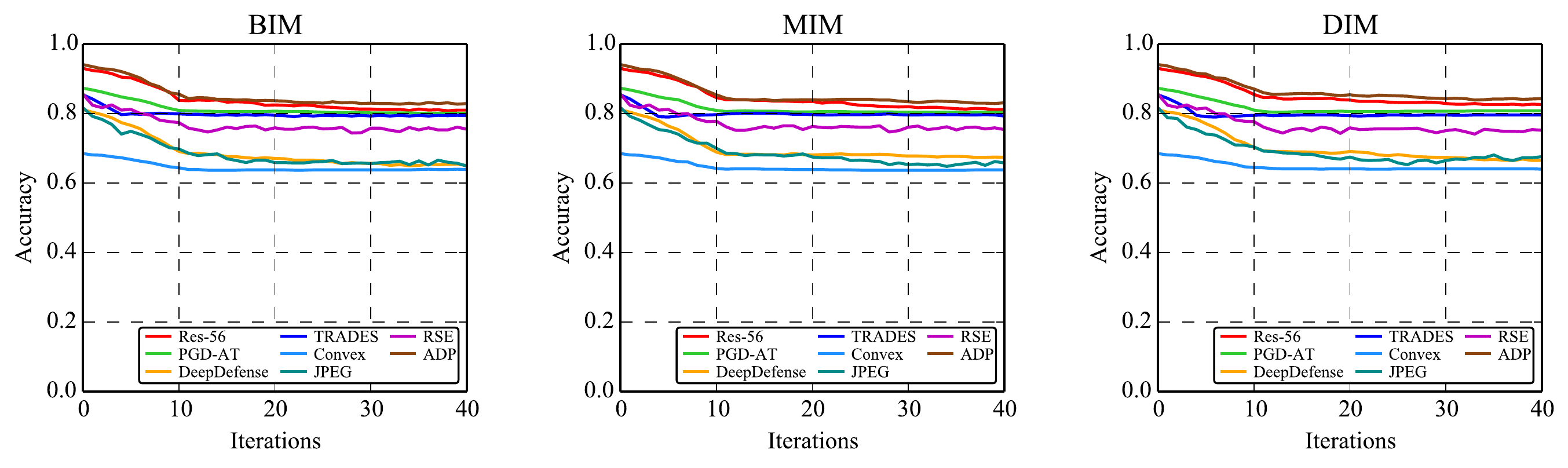}
\end{center}
\vspace{-2ex}
\caption{The \textit{accuracy vs. attack strength} curves of the $8$ models on CIFAR-10 against targeted transfer-based attacks under the $\ell_{\infty}$ norm.}
\label{fig:trans-t-linf-cifar10-acc-iter}
\end{figure*}

\begin{figure*}[t]
\begin{center}
\includegraphics[width=0.85\linewidth]{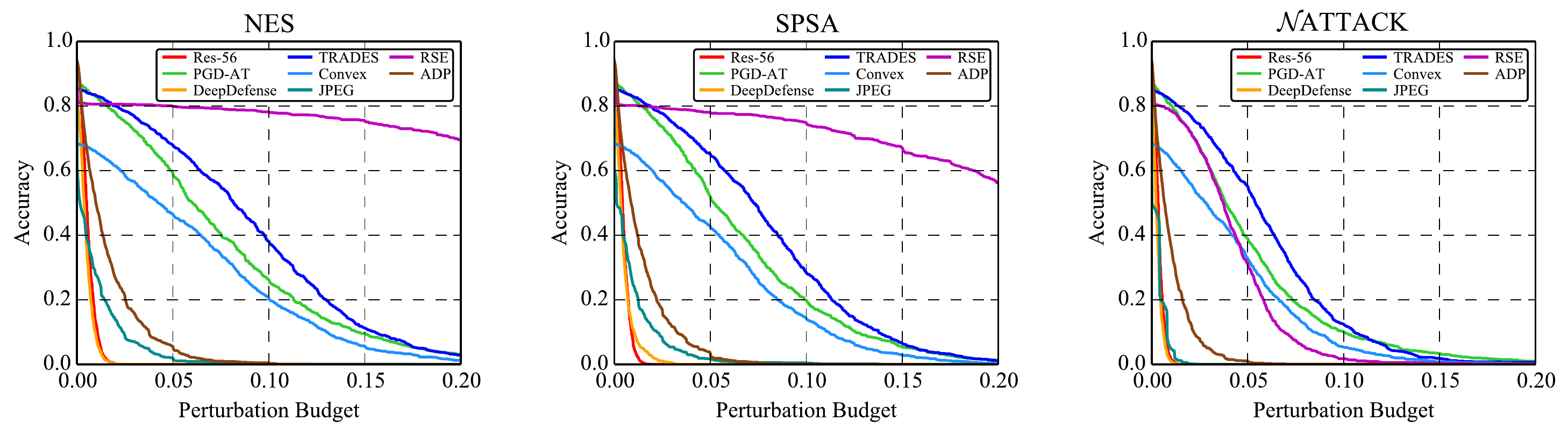}
\end{center}
\vspace{-2ex}
\caption{The \textit{accuracy vs. perturbation budget} curves of the $8$ models on CIFAR-10 against targeted score-based attacks under the $\ell_{\infty}$ norm.}
\label{fig:score-t-linf-cifar10-acc-pert}
\begin{center}
\includegraphics[width=0.85\linewidth]{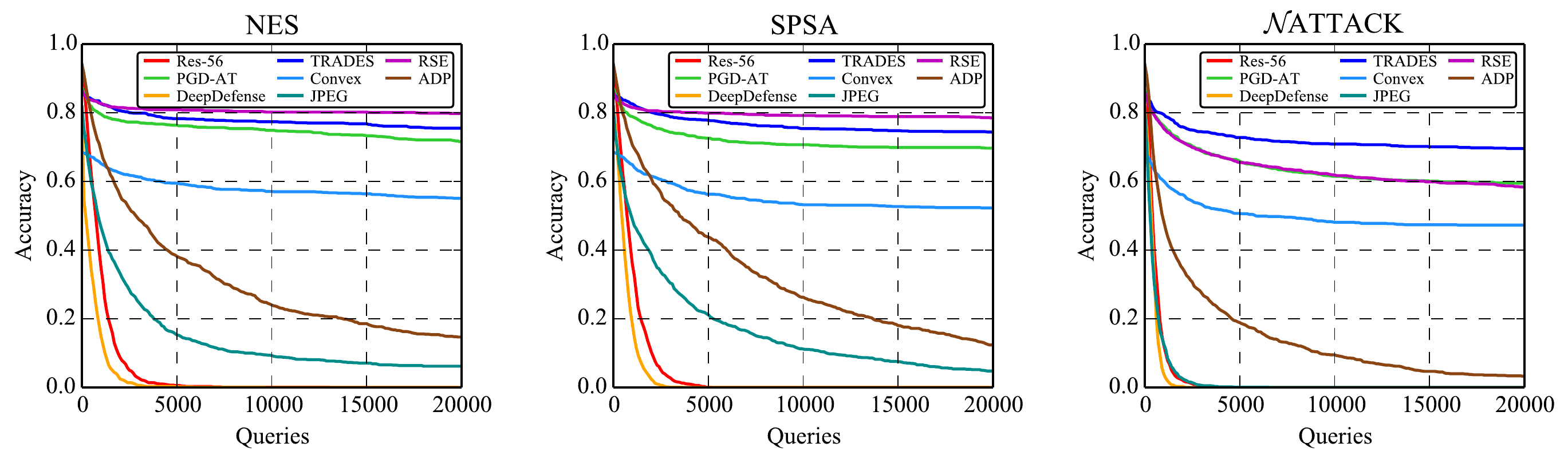}
\end{center}
\vspace{-2ex}
\caption{The \textit{accuracy vs. attack strength} curves of the $8$ models on CIFAR-10 against targeted score-based attacks under the $\ell_{\infty}$ norm.}
\label{fig:score-t-linf-cifar10-acc-iter}
\end{figure*}

\begin{figure*}[t]
\begin{center}
\includegraphics[width=1.0\linewidth]{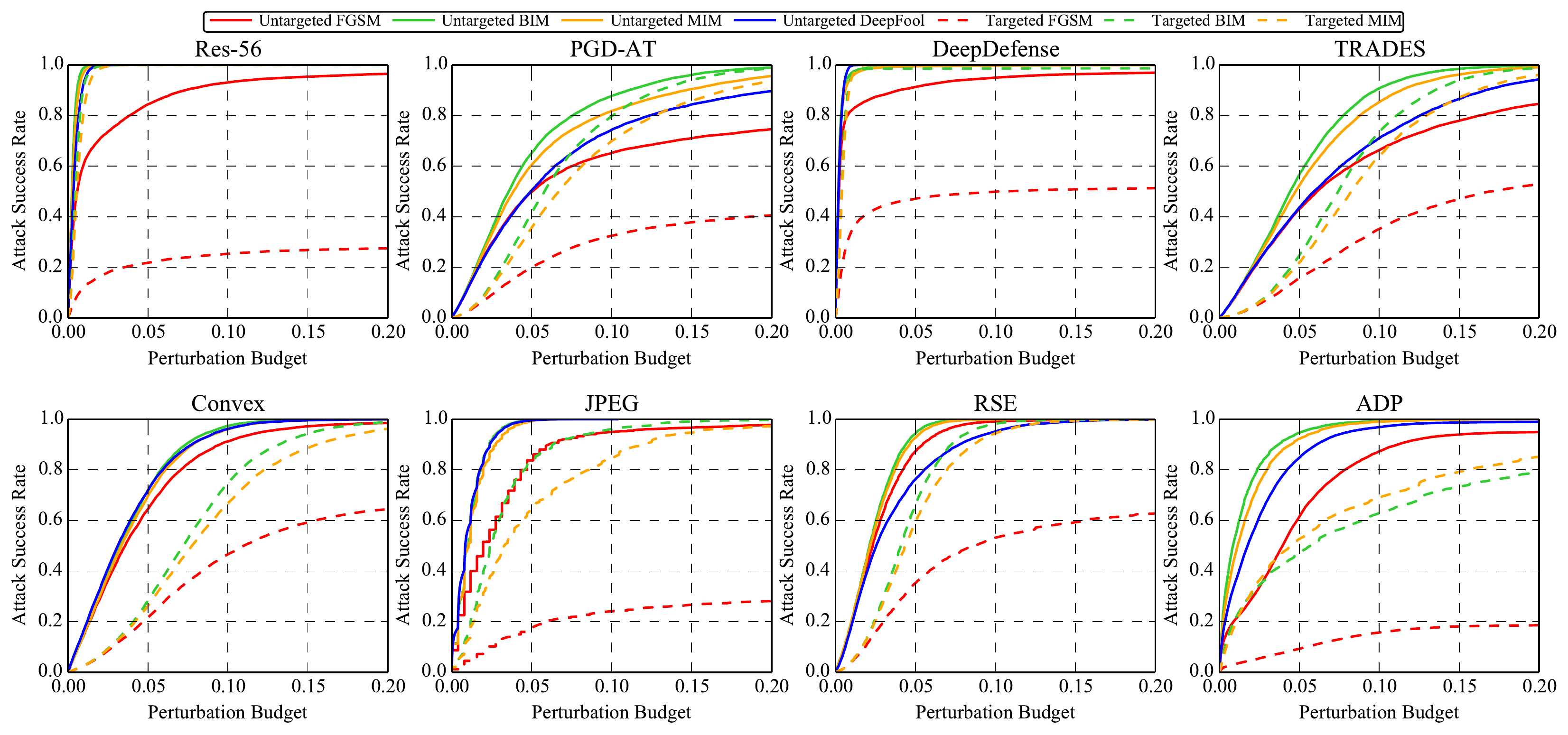}
\end{center}
\vspace{-2ex}
\caption{The \textit{attack success rate vs. perturbation budget} curves of white-box attacks under the $\ell_{\infty}$ norm on the $8$ models on CIFAR-10.}
\label{fig:white-linf-cifar10-asr-pert}
\end{figure*}

\begin{figure*}[t]
\begin{center}
\includegraphics[width=1.0\linewidth]{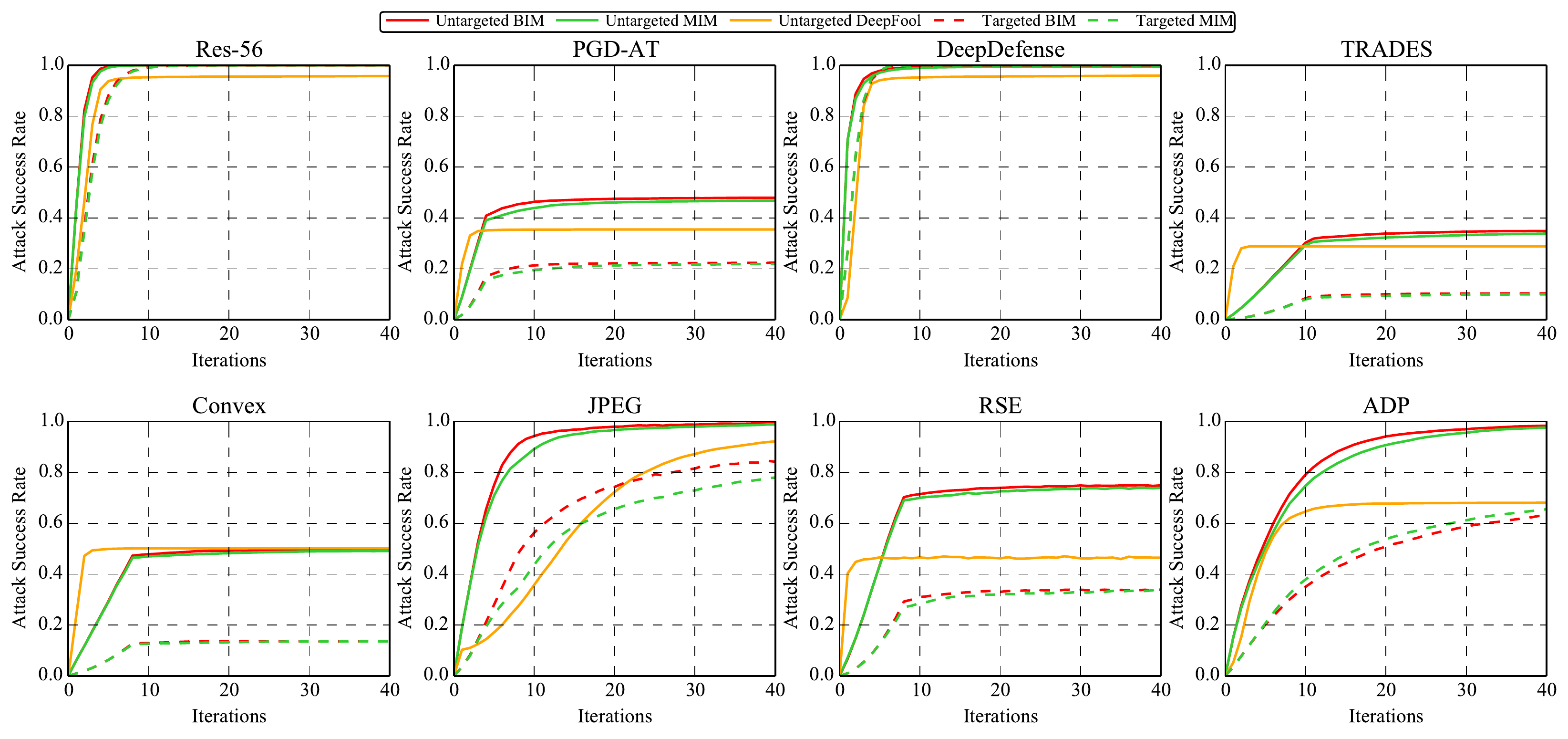}
\end{center}
\vspace{-2ex}
\caption{The \textit{attack success rate vs. attack strength} curves of white-box attacks under the $\ell_{\infty}$ norm on the $8$ models on CIFAR-10.}
\label{fig:white-linf-cifar10-asr-iter}
\end{figure*}

\begin{figure*}[t]
\begin{center}
\includegraphics[width=1.0\linewidth]{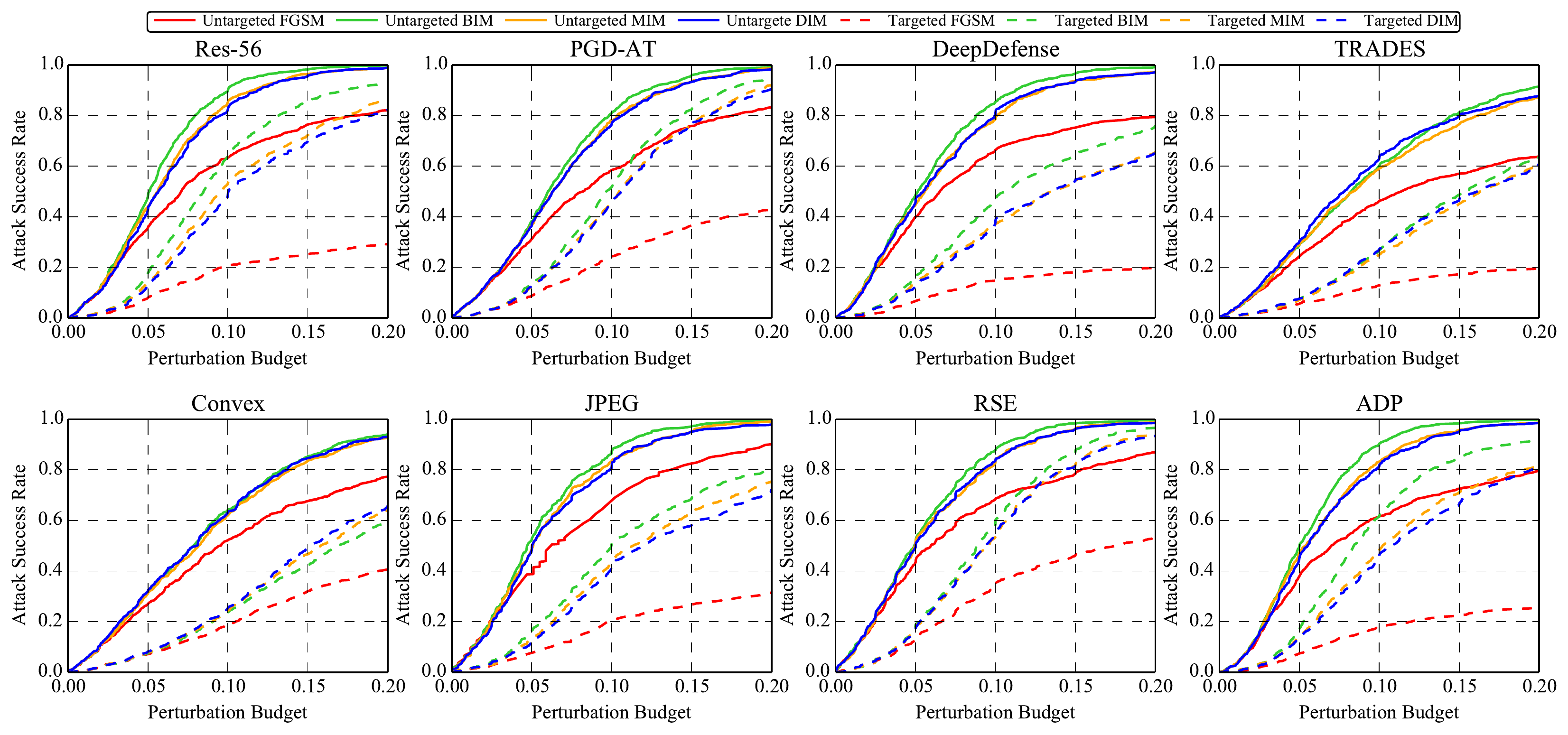}
\end{center}
\vspace{-2ex}
\caption{The \textit{attack success rate vs. perturbation budget} curves of transfer-based attacks under the $\ell_{\infty}$ norm on the $8$ models on CIFAR-10.}
\label{fig:trans-linf-cifar10-asr-pert}
\end{figure*}

\begin{figure*}[t]
\begin{center}
\includegraphics[width=1.0\linewidth]{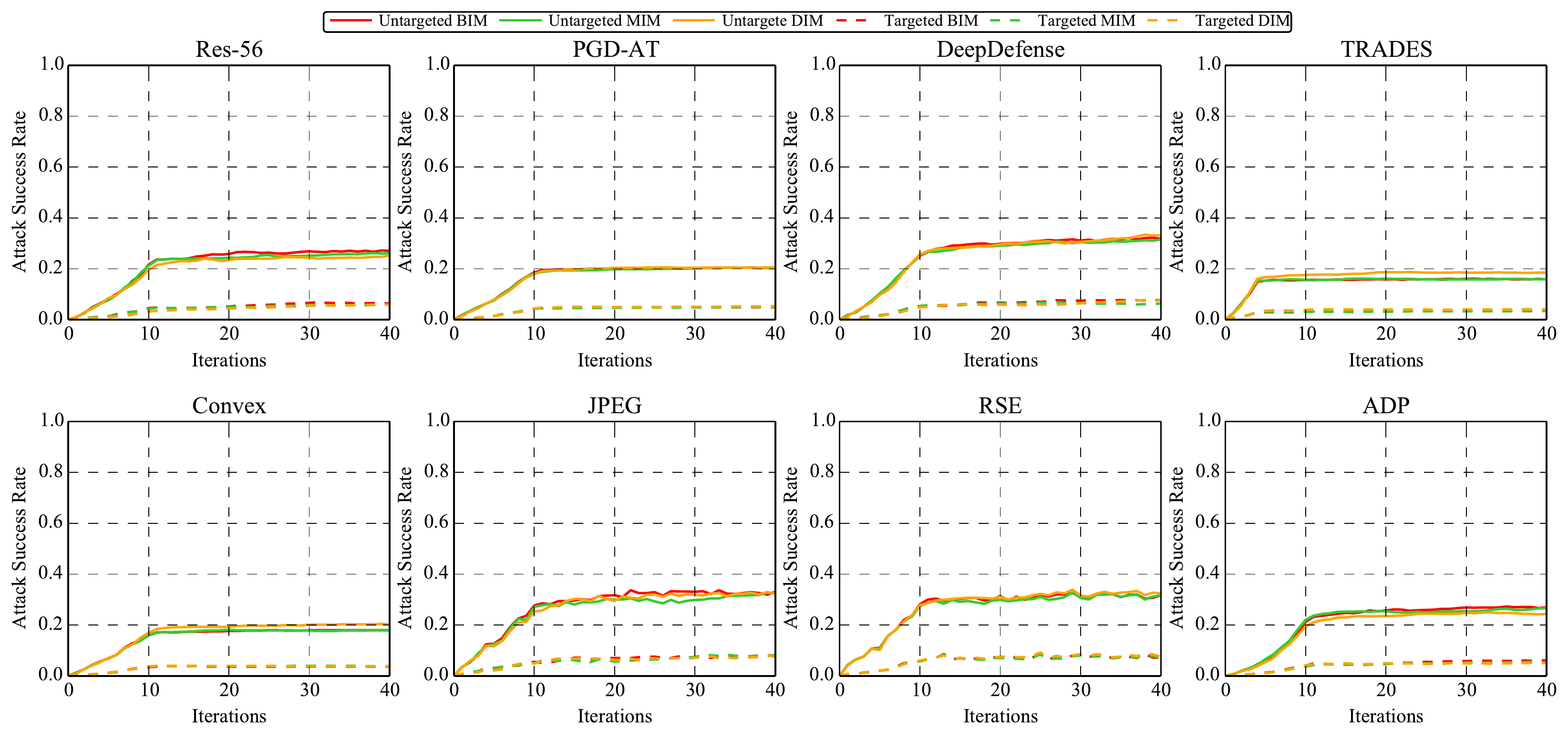}
\end{center}
\vspace{-2ex}
\caption{The \textit{attack success rate vs. attack strength} curves of transfer-based attacks under the $\ell_{\infty}$ norm on the $8$ models on CIFAR-10.}
\label{fig:trans-linf-cifar10-asr-iter}
\end{figure*}

\begin{figure*}[t]
\begin{center}
\includegraphics[width=1.0\linewidth]{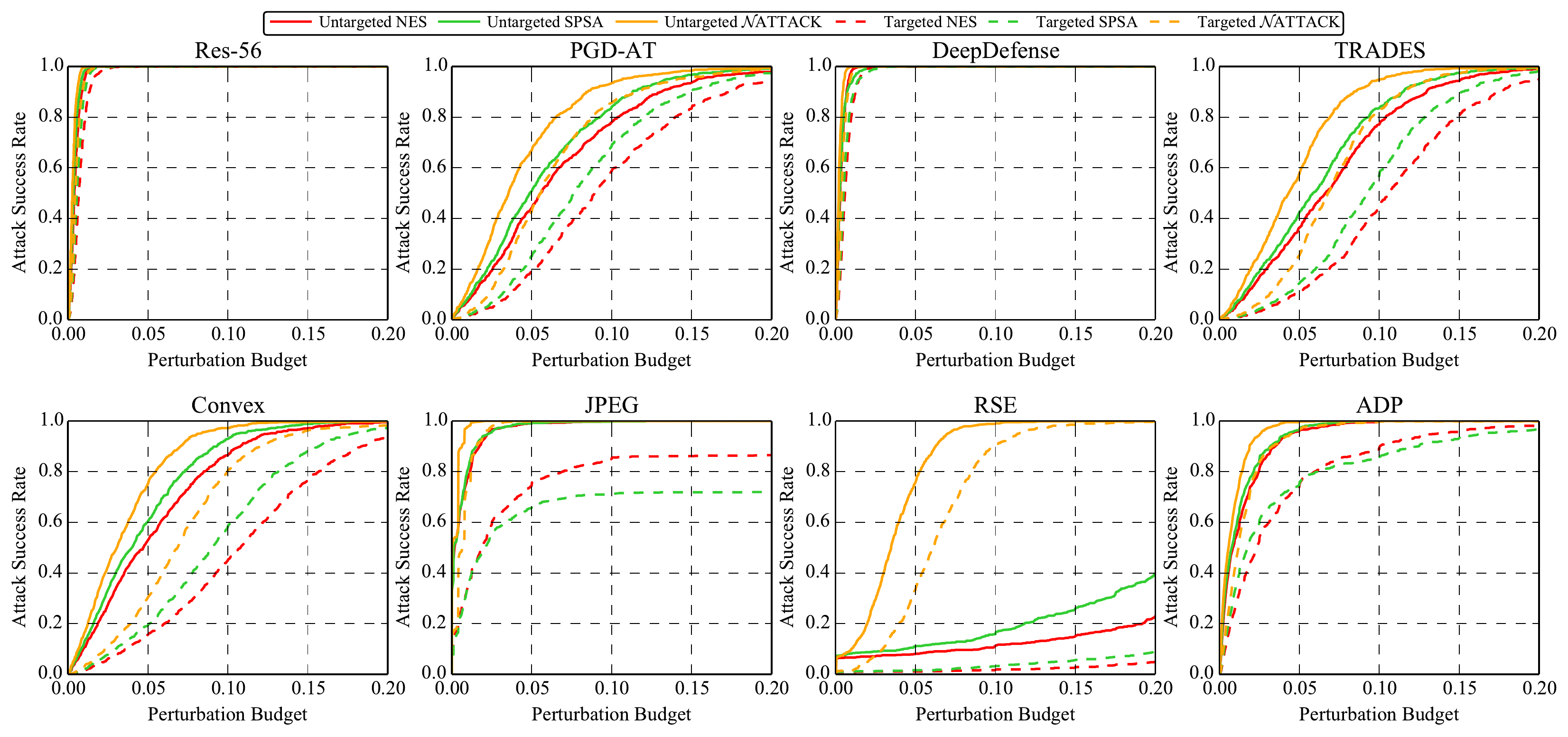}
\end{center}
\vspace{-2ex}
\caption{The \textit{attack success rate vs. perturbation budget} curves of score-based attacks under the $\ell_{\infty}$ norm on the $8$ models on CIFAR-10.}
\label{fig:score-linf-cifar10-asr-pert}
\end{figure*}

\begin{figure*}[t]
\begin{center}
\includegraphics[width=1.0\linewidth]{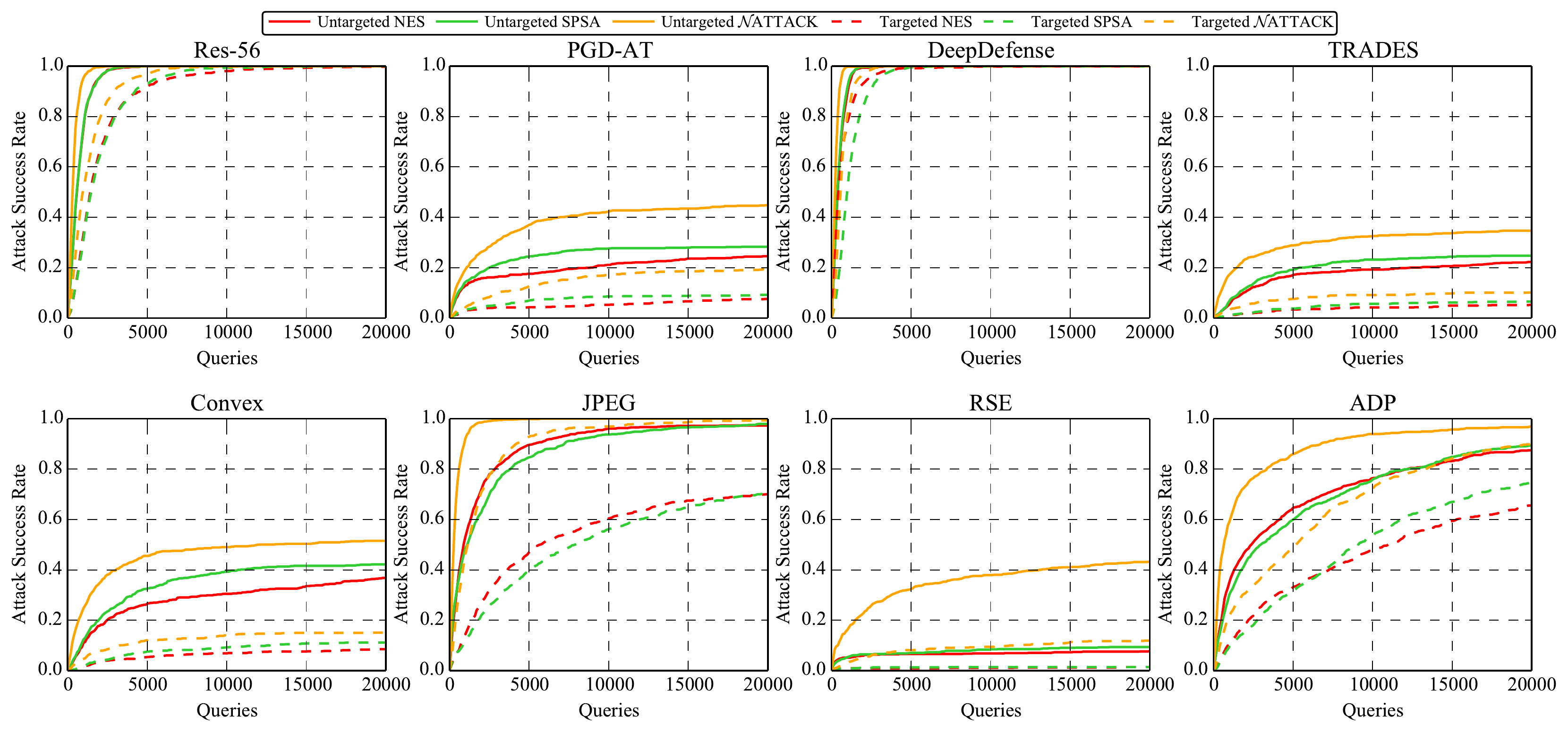}
\end{center}
\vspace{-2ex}
\caption{The \textit{attack success rate vs. attack strength} curves of score-based attacks under the $\ell_{\infty}$ norm on the $8$ models on CIFAR-10.}
\label{fig:score-linf-cifar10-asr-iter}
\end{figure*}

\textbf{Attacks under the $\ell_{\infty}$ norm:} We have shown some of the accuracy curves of the defense models against untargeted attacks under the $\ell_{\infty}$ norm in Sec.~\ref{sec:exp-cifar}. We next show the remaining curves of untargeted attacks under the $\ell_{\infty}$ norm, the curves of targeted attacks under the $\ell_{\infty}$ norm, and the attack success rate curves.
Fig.~\ref{fig:white-ut-linf-cifar10-acc-iter} shows the \emph{accuracy vs. attack strength} curves of the defenses on CIFAR-10 against untargeted white-box attacks under the $\ell_{\infty}$ norm.
Fig.~\ref{fig:white-t-linf-cifar10-acc-pert} and Fig.~\ref{fig:white-t-linf-cifar10-acc-iter} show the accuracy curves of the defenses on CIFAR-10 against targeted white-box attacks under the $\ell_{\infty}$ norm.
Fig.~\ref{fig:trans-ut-linf-cifar10-acc-iter} shows the \emph{accuracy vs. perturbation budget} curves of the defenses on CIFAR-10 against untargeted transfer-based attacks under the $\ell_{\infty}$ norm.
Fig.~\ref{fig:trans-t-linf-cifar10-acc-pert} and Fig.~\ref{fig:trans-t-linf-cifar10-acc-iter} show the accuracy curves of the defenses on CIFAR-10 against targeted transfer-based attacks under the $\ell_{\infty}$ norm.
Fig.~\ref{fig:score-t-linf-cifar10-acc-pert} and Fig.~\ref{fig:score-t-linf-cifar10-acc-iter} show the accuracy curves of the defenses on CIFAR-10 against targeted score-based attacks under the $\ell_{\infty}$ norm.
Fig.~\ref{fig:white-linf-cifar10-asr-pert} to Fig.~\ref{fig:score-linf-cifar10-asr-iter} show the \textit{attack success rate vs. perturbation budget} and \textit{attack success rate vs. attack strength} curves of white-box, transfer-based, and score-based attacks under the $\ell_{\infty}$ norm on the $8$ models on CIFAR-10.

\begin{figure*}[t]
\begin{center}
\includegraphics[width=1.0\linewidth]{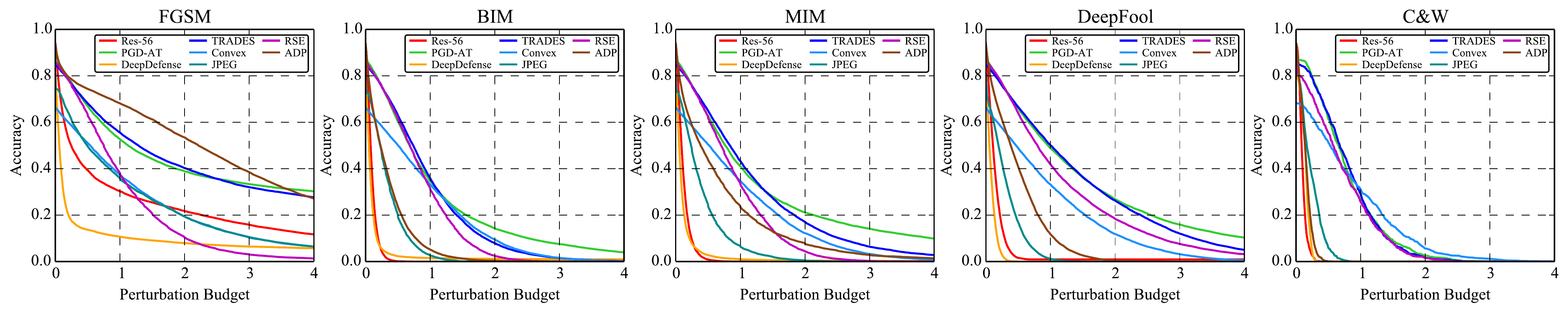}
\end{center}
\vspace{-2ex}
\caption{The \textit{accuracy vs. perturbation budget} curves of the $8$ models on CIFAR-10 against untargeted white-box attacks under the $\ell_2$ norm.}
\label{fig:white-ut-l2-cifar10-acc-pert}
\begin{center}
\includegraphics[width=1.0\linewidth]{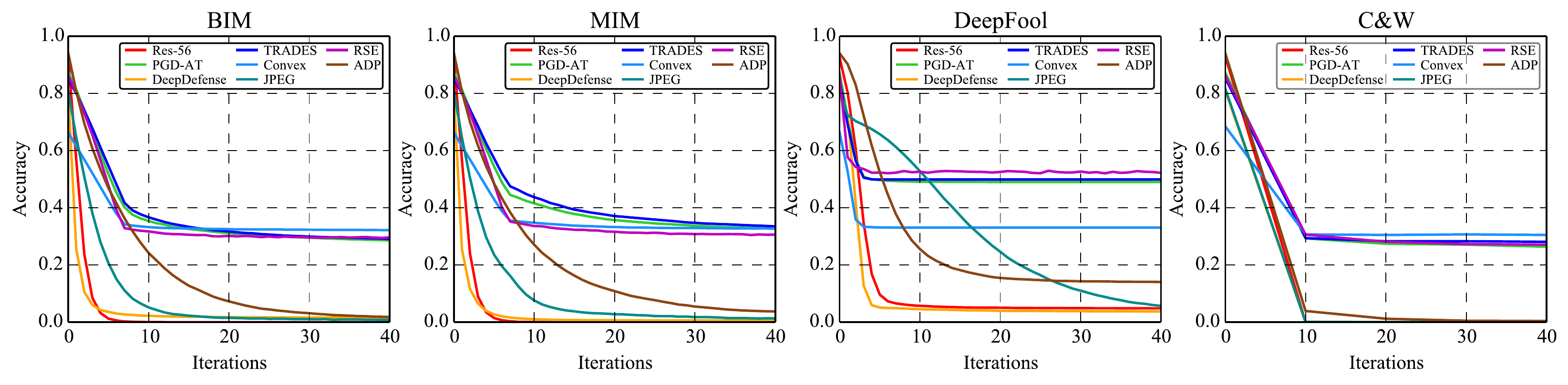}
\end{center}
\vspace{-2ex}
\caption{The \textit{accuracy vs. attack strength} curves of the $8$ models on CIFAR-10 against untargeted white-box attacks under the $\ell_2$ norm.}
\label{fig:white-ut-l2-cifar10-acc-iter}
\end{figure*}

\begin{figure*}[t]
\begin{center}
\includegraphics[width=1.0\linewidth]{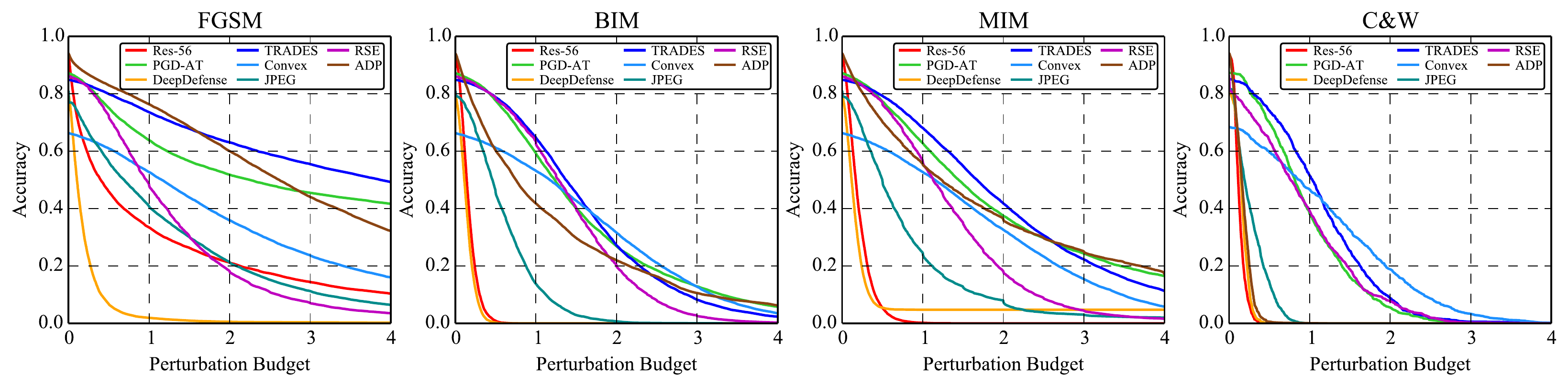}
\end{center}
\vspace{-2ex}
\caption{The \textit{accuracy vs. perturbation budget} curves of the $8$ models on CIFAR-10 against targeted white-box attacks under the $\ell_2$ norm.}
\label{fig:white-t-l2-cifar10-acc-pert}
\begin{center}
\includegraphics[width=0.85\linewidth]{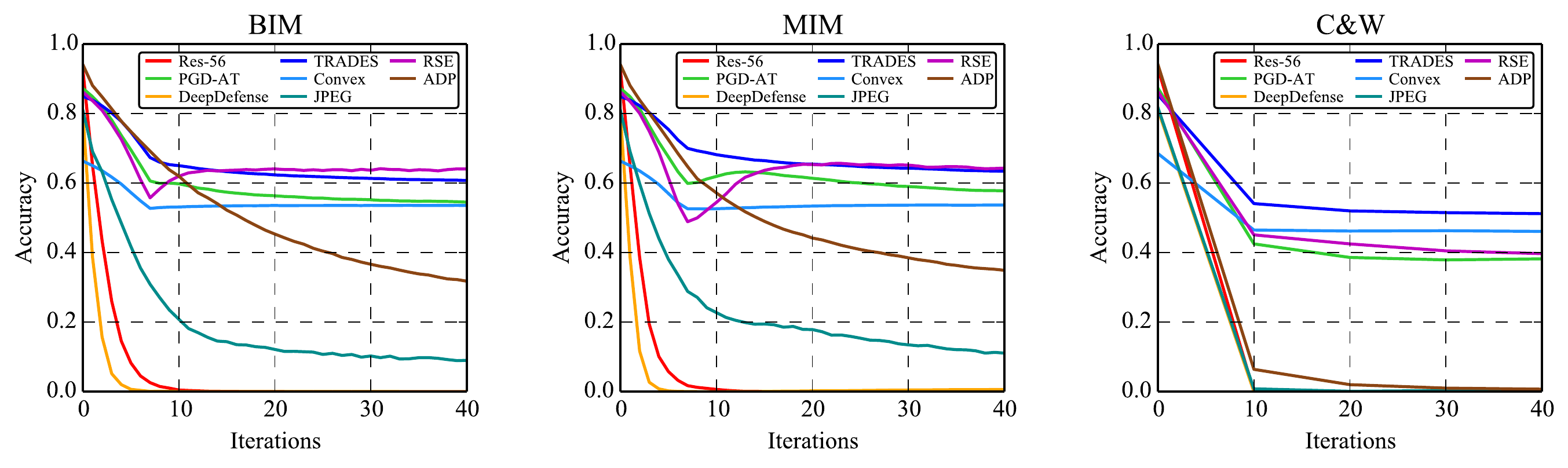}
\end{center}
\vspace{-2ex}
\caption{The \textit{accuracy vs. attack strength} curves of the $8$ models on CIFAR-10 against targeted white-box attacks under the $\ell_2$ norm.}
\label{fig:white-t-l2-cifar10-acc-iter}
\end{figure*}

\begin{figure*}[t]
\begin{center}
\includegraphics[width=1.0\linewidth]{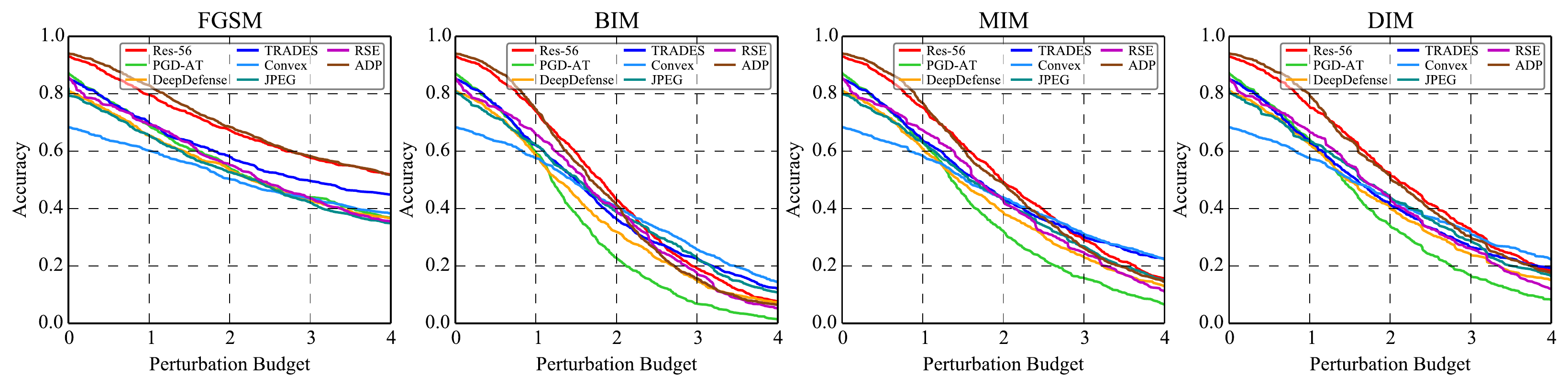}
\end{center}
\vspace{-2ex}
\caption{The \textit{accuracy vs. perturbation budget} curves of the $8$ models on CIFAR-10 against untargeted transfer-based attacks under the $\ell_2$ norm.}
\label{fig:trans-ut-l2-cifar10-acc-pert}
\begin{center}
\includegraphics[width=0.85\linewidth]{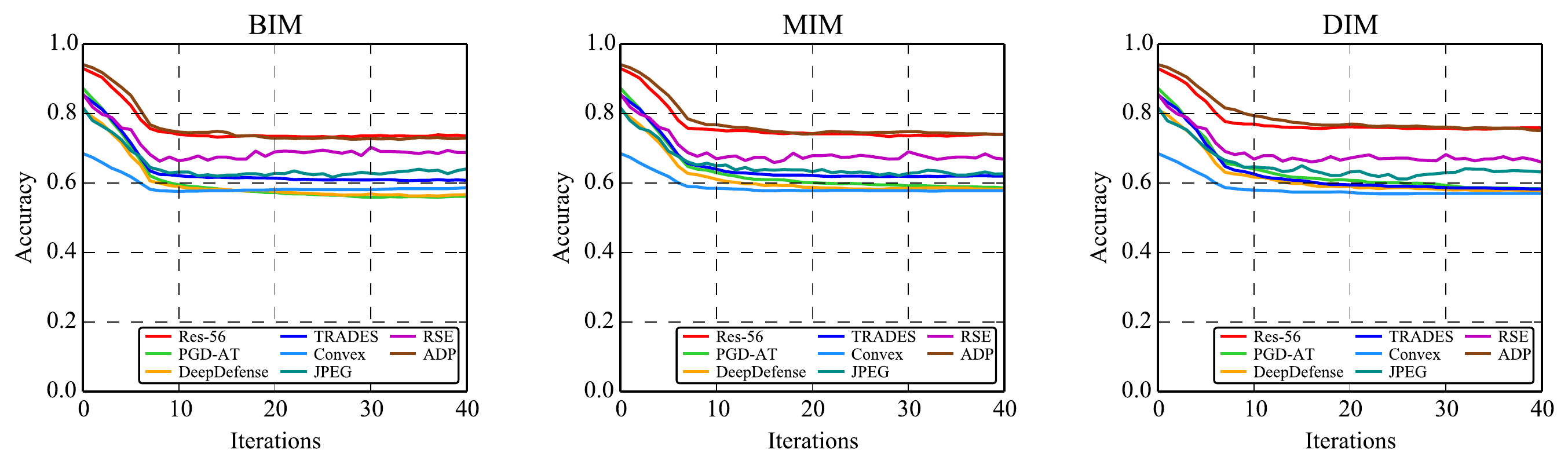}
\end{center}
\vspace{-2ex}
\caption{The \textit{accuracy vs. attack strength} curves of the $8$ models on CIFAR-10 against untargeted transfer-based attacks under the $\ell_2$ norm.}
\label{fig:trans-ut-l2-cifar10-acc-iter}
\end{figure*}

\begin{figure*}[t]
\begin{center}
\includegraphics[width=1.0\linewidth]{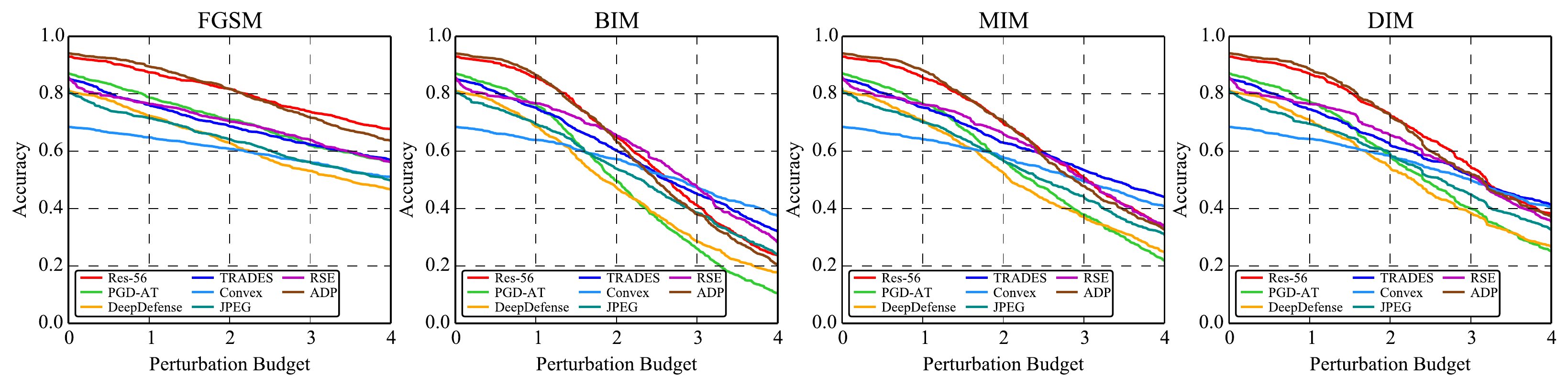}
\end{center}
\vspace{-2ex}
\caption{The \textit{accuracy vs. perturbation budget} curves of the $8$ models on CIFAR-10 against targeted transfer-based attacks under the $\ell_2$ norm.}
\label{fig:trans-t-l2-cifar10-acc-pert}
\begin{center}
\includegraphics[width=0.85\linewidth]{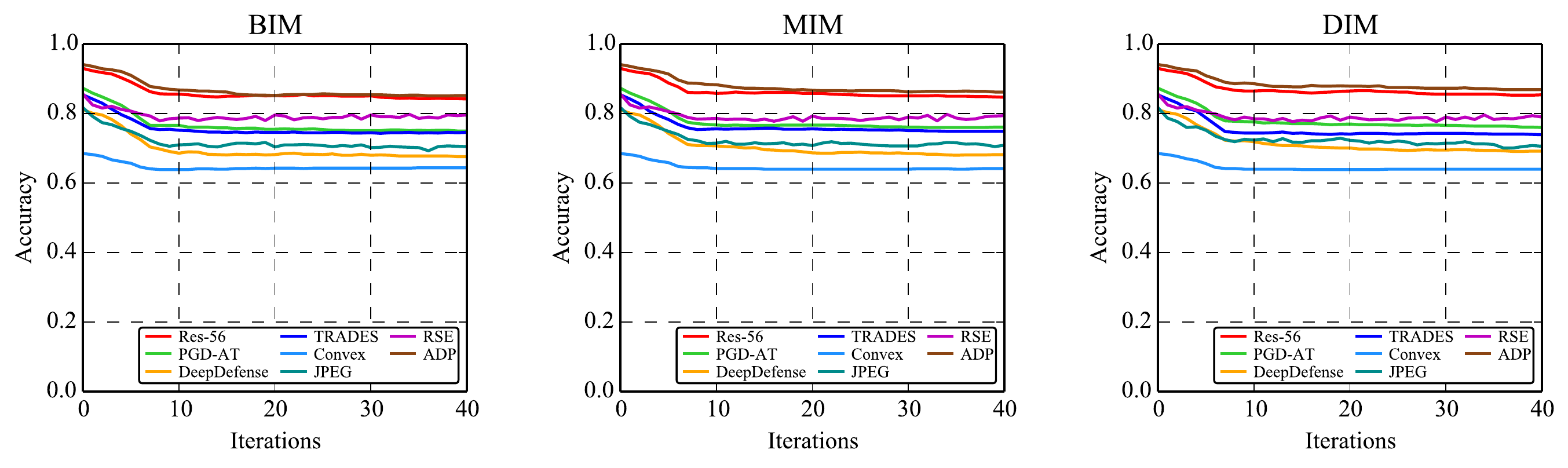}
\end{center}
\vspace{-2ex}
\caption{The \textit{accuracy vs. attack strength} curves of the $8$ models on CIFAR-10 against targeted transfer-based attacks under the $\ell_2$ norm.}
\label{fig:trans-t-l2-cifar10-acc-iter}
\end{figure*}

\begin{figure*}[t]
\begin{center}
\includegraphics[width=1.0\linewidth]{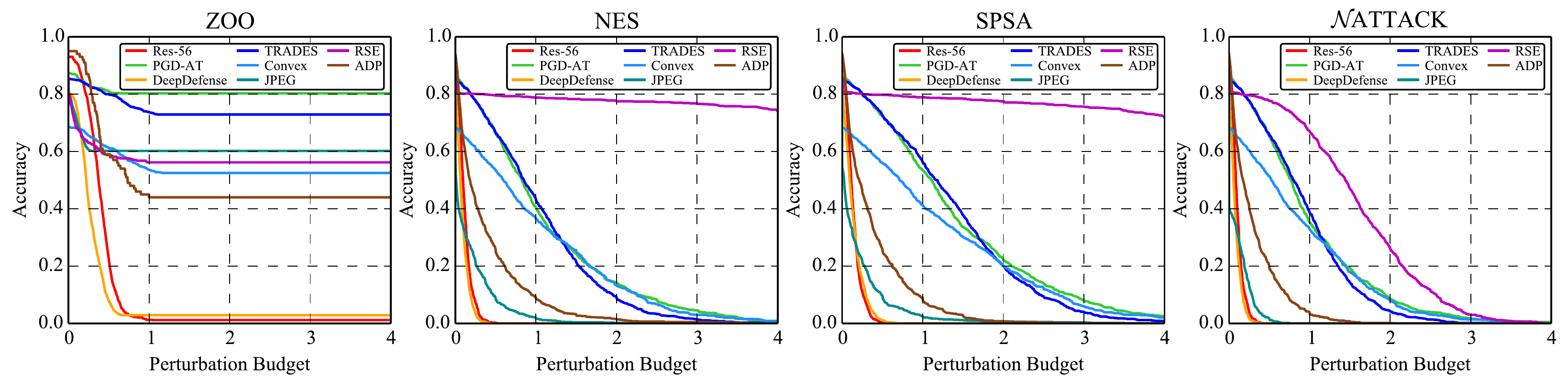}
\end{center}
\vspace{-2ex}
\caption{The \textit{accuracy vs. perturbation budget} curves of the $8$ models on CIFAR-10 against untargeted score-based attacks under the $\ell_2$ norm.}
\label{fig:score-ut-l2-cifar10-acc-pert}
\begin{center}
\includegraphics[width=1.0\linewidth]{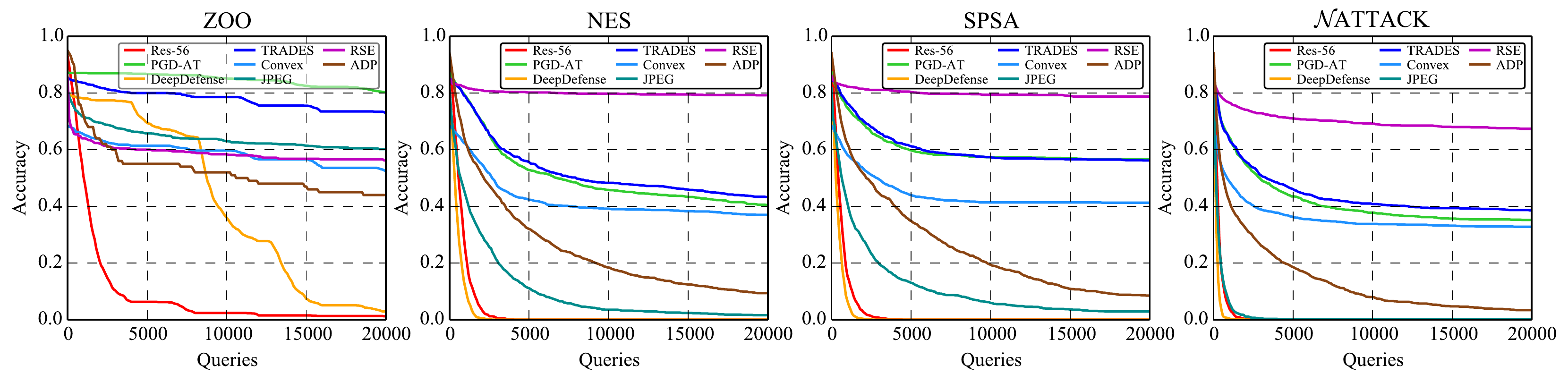}
\end{center}
\vspace{-2ex}
\caption{The \textit{accuracy vs. attack strength} curves of the $8$ models on CIFAR-10 against untargeted score-based attacks under the $\ell_2$ norm.}
\label{fig:score-ut-l2-cifar10-acc-iter}
\end{figure*}

\begin{figure*}[t]
\begin{center}
\includegraphics[width=1.0\linewidth]{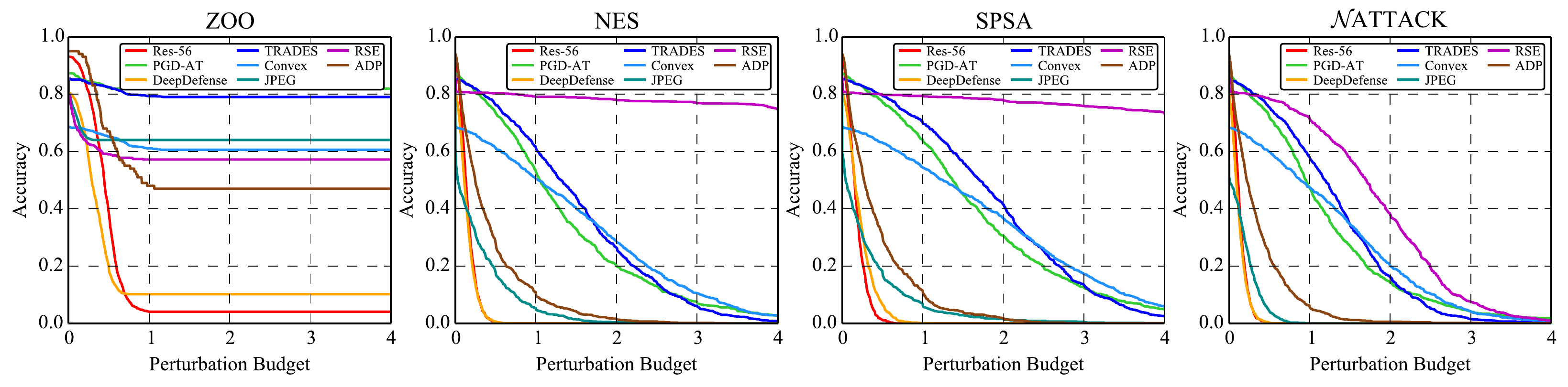}
\end{center}
\vspace{-2ex}
\caption{The \textit{accuracy vs. perturbation budget} curves of the $8$ models on CIFAR-10 against targeted score-based attacks under the $\ell_2$ norm.}
\label{fig:score-t-l2-cifar10-acc-pert}
\begin{center}
\includegraphics[width=1.0\linewidth]{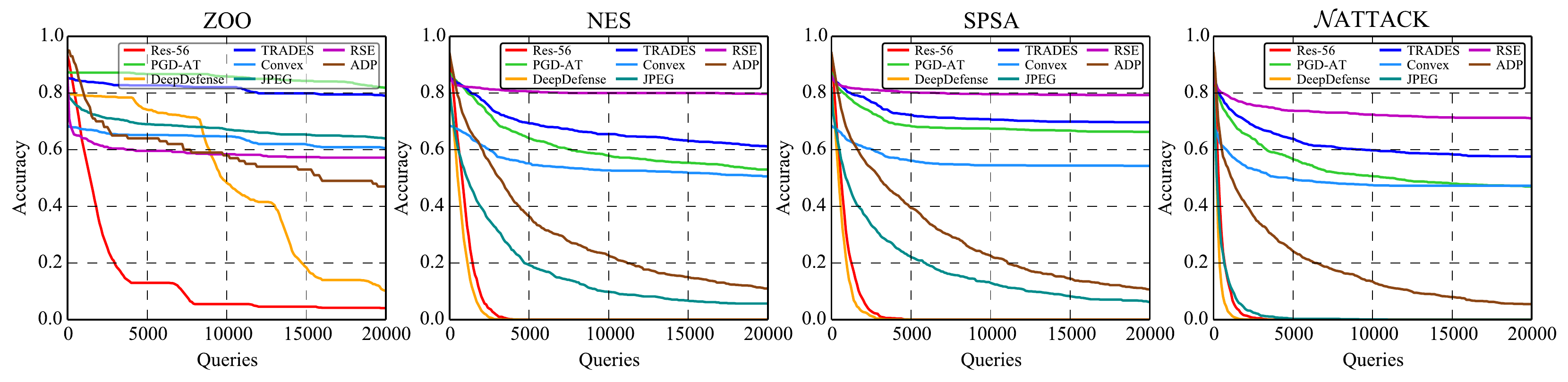}
\end{center}
\vspace{-2ex}
\caption{The \textit{accuracy vs. attack strength} curves of the $8$ models on CIFAR-10 against targeted score-based attacks under the $\ell_2$ norm.}
\label{fig:score-t-l2-cifar10-acc-iter}
\end{figure*}

\begin{figure*}[t]
\begin{minipage}{.49\linewidth}
\begin{center}
\includegraphics[width=1.0\linewidth]{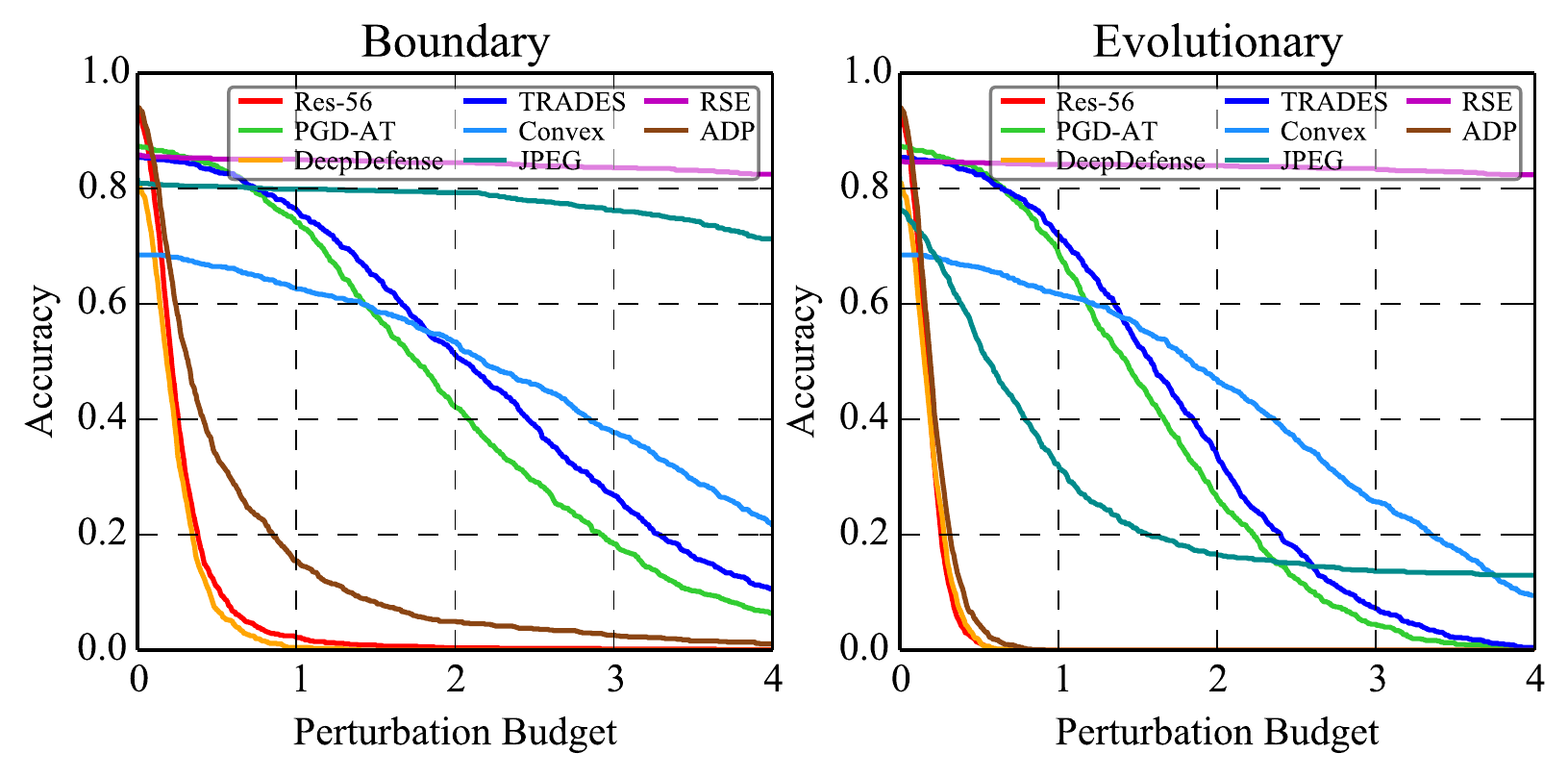}
\end{center}
\vspace{-2ex}
\caption{The \textit{accuracy vs. perturbation budget} curves of the $8$ models on CIFAR-10 against targeted decision-based attacks under the $\ell_2$ norm.}
\label{fig:decision-t-l2-cifar10-acc-pert}
\end{minipage}
\hspace{1ex}
\begin{minipage}{.49\linewidth}
\begin{center}
\includegraphics[width=1.0\linewidth]{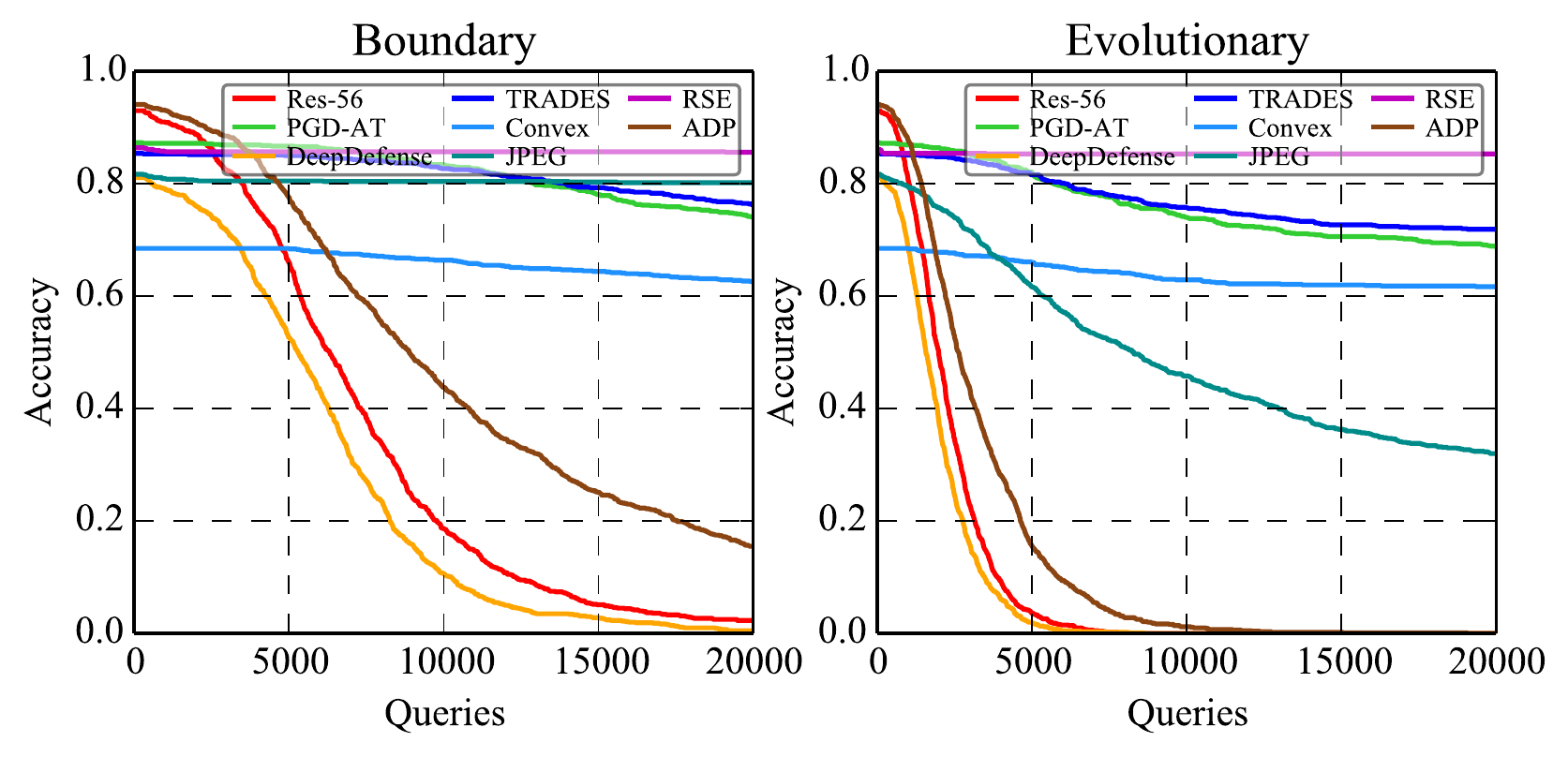}
\end{center}
\vspace{-2ex}
\caption{The \textit{accuracy vs. attack strength} curves of the $8$ models on CIFAR-10 against targeted decision-based attacks under the $\ell_2$ norm.}
\label{fig:decision-t-l2-cifar10-acc-iter}
\end{minipage}
\end{figure*}

\begin{figure*}[t]
\begin{center}
\includegraphics[width=1.0\linewidth]{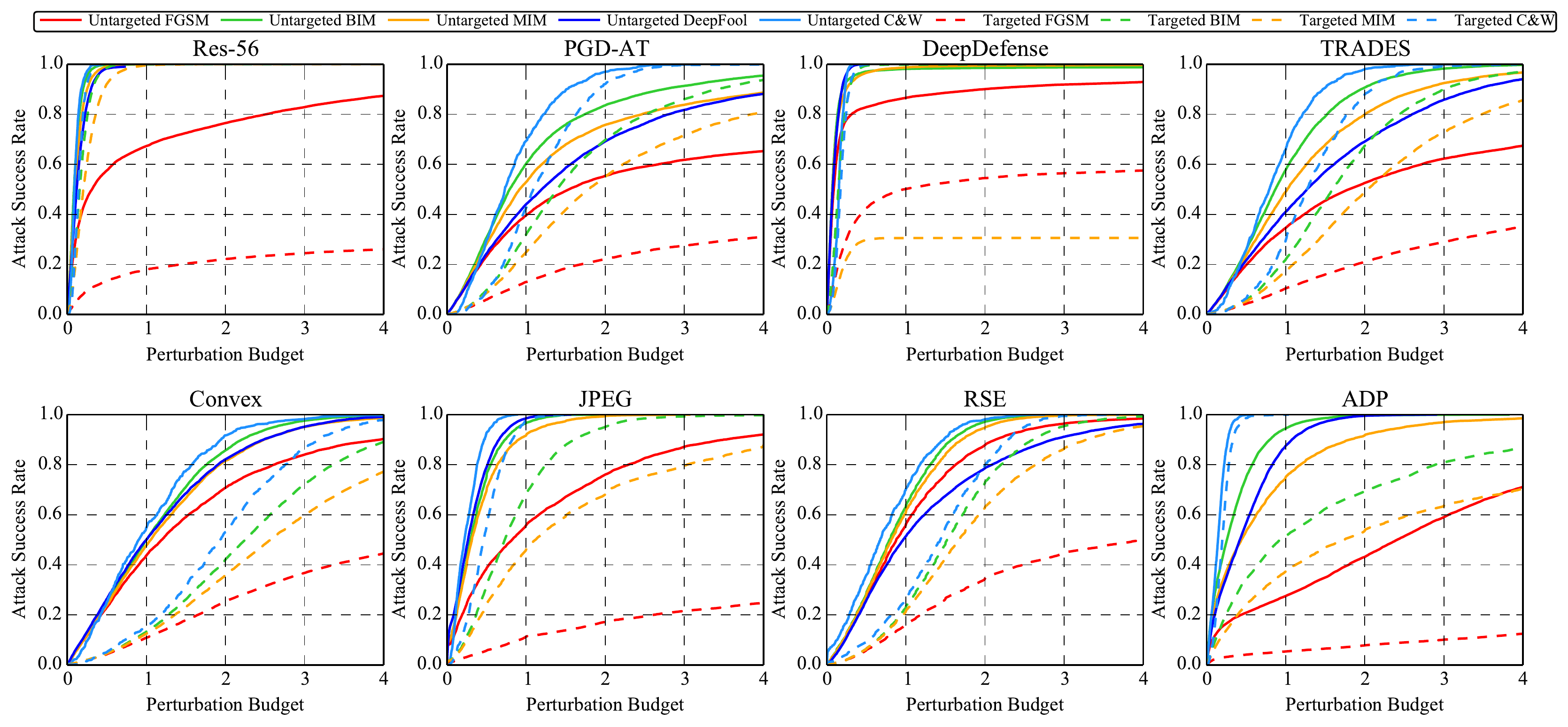}
\end{center}
\vspace{-2ex}
\caption{The \textit{attack success rate vs. perturbation budget} curves of white-box attacks under the $\ell_2$ norm on the $8$ models on CIFAR-10.}
\label{fig:white-l2-cifar10-asr-pert}
\end{figure*}

\begin{figure*}[t]
\begin{center}
\includegraphics[width=1.0\linewidth]{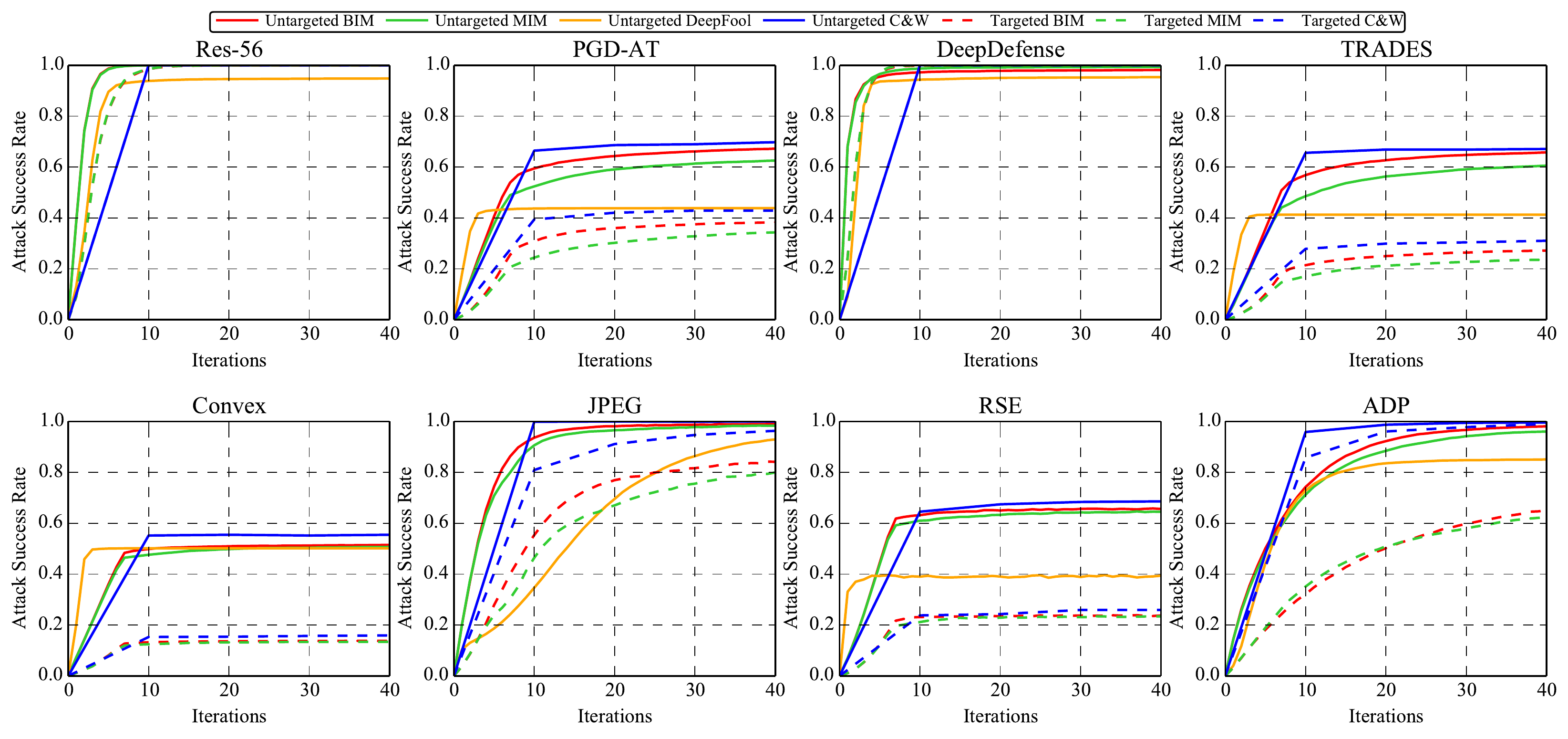}
\end{center}
\vspace{-3ex}
\caption{The \textit{attack success rate vs. attack strength} curves of white-box attacks under the $\ell_2$ norm on the $8$ models on CIFAR-10.}
\label{fig:white-l2-cifar10-asr-iter}
\end{figure*}

\begin{figure*}[t]
\begin{center}
\includegraphics[width=1.0\linewidth]{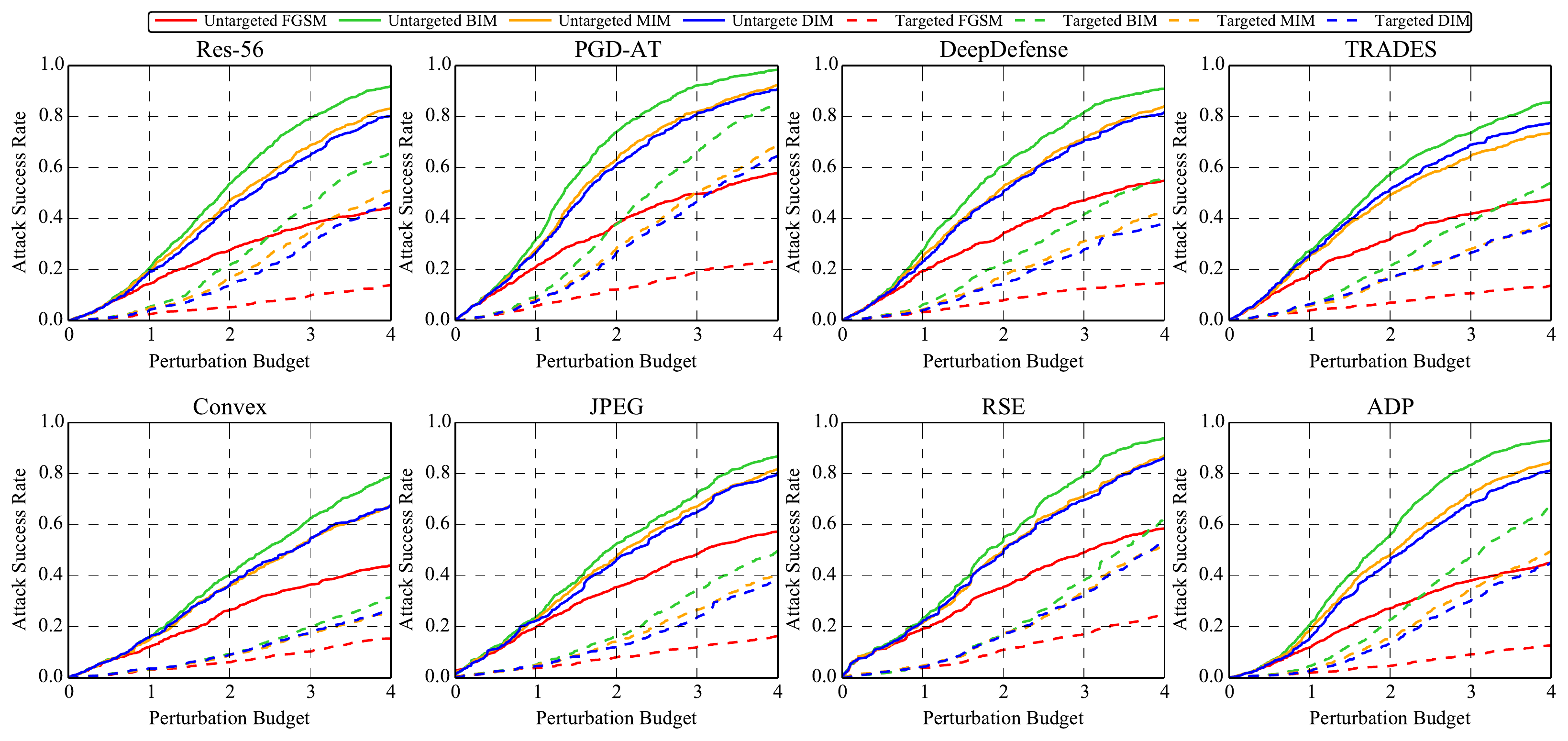}
\end{center}
\vspace{-3ex}
\caption{The \textit{attack success rate vs. perturbation budget} curves of transfer-based attacks under the $\ell_2$ norm on the $8$ models on CIFAR-10.}
\label{fig:trans-l2-cifar10-asr-pert}
\end{figure*}

\begin{figure*}[t]
\begin{center}
\includegraphics[width=1.0\linewidth]{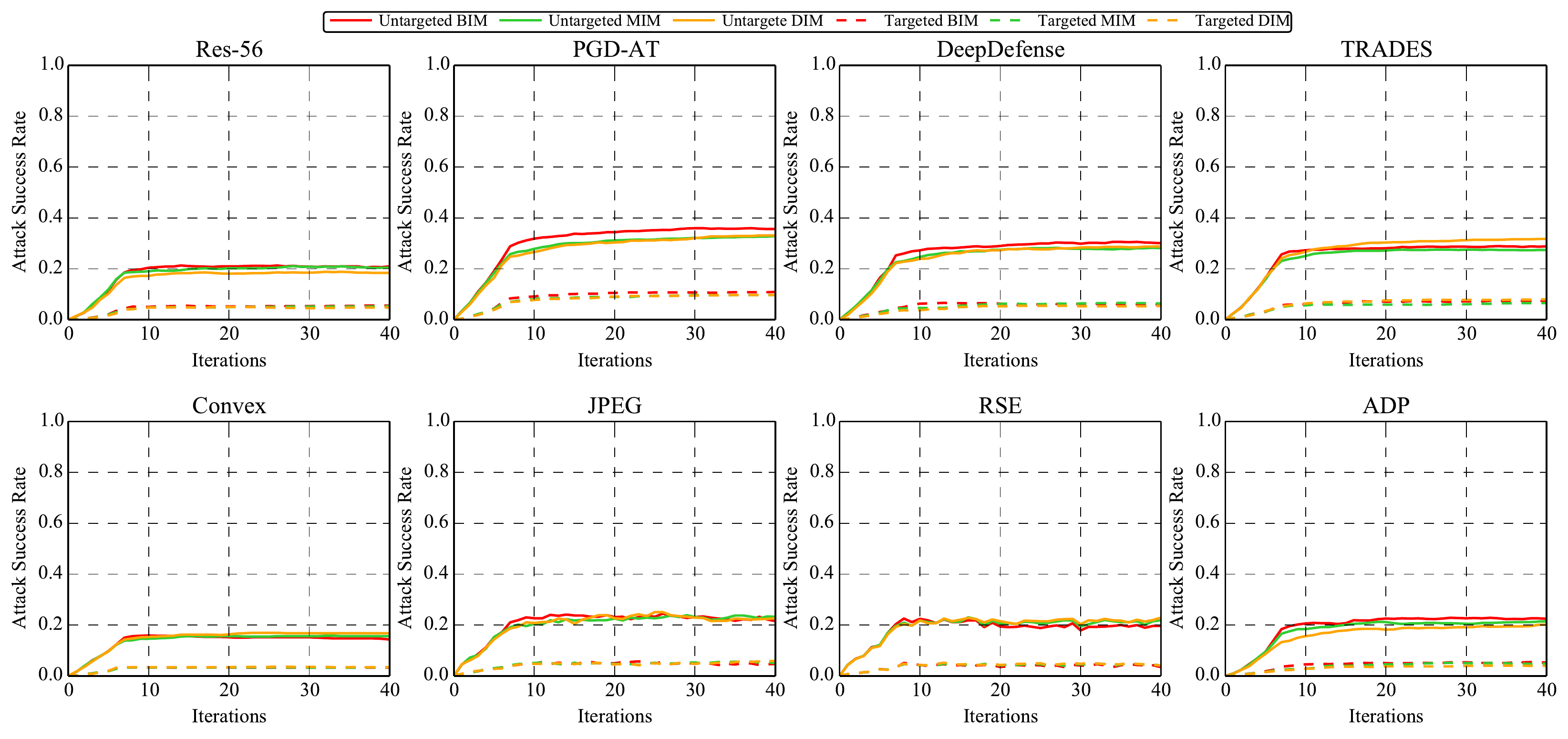}
\end{center}
\vspace{-3ex}
\caption{The \textit{attack success rate vs. attack strength} curves of transfer-based attacks under the $\ell_2$ norm on the $8$ models on CIFAR-10.}
\label{fig:trans-l2-cifar10-asr-iter}
\end{figure*}

\begin{figure*}[t]
\begin{center}
\includegraphics[width=1.0\linewidth]{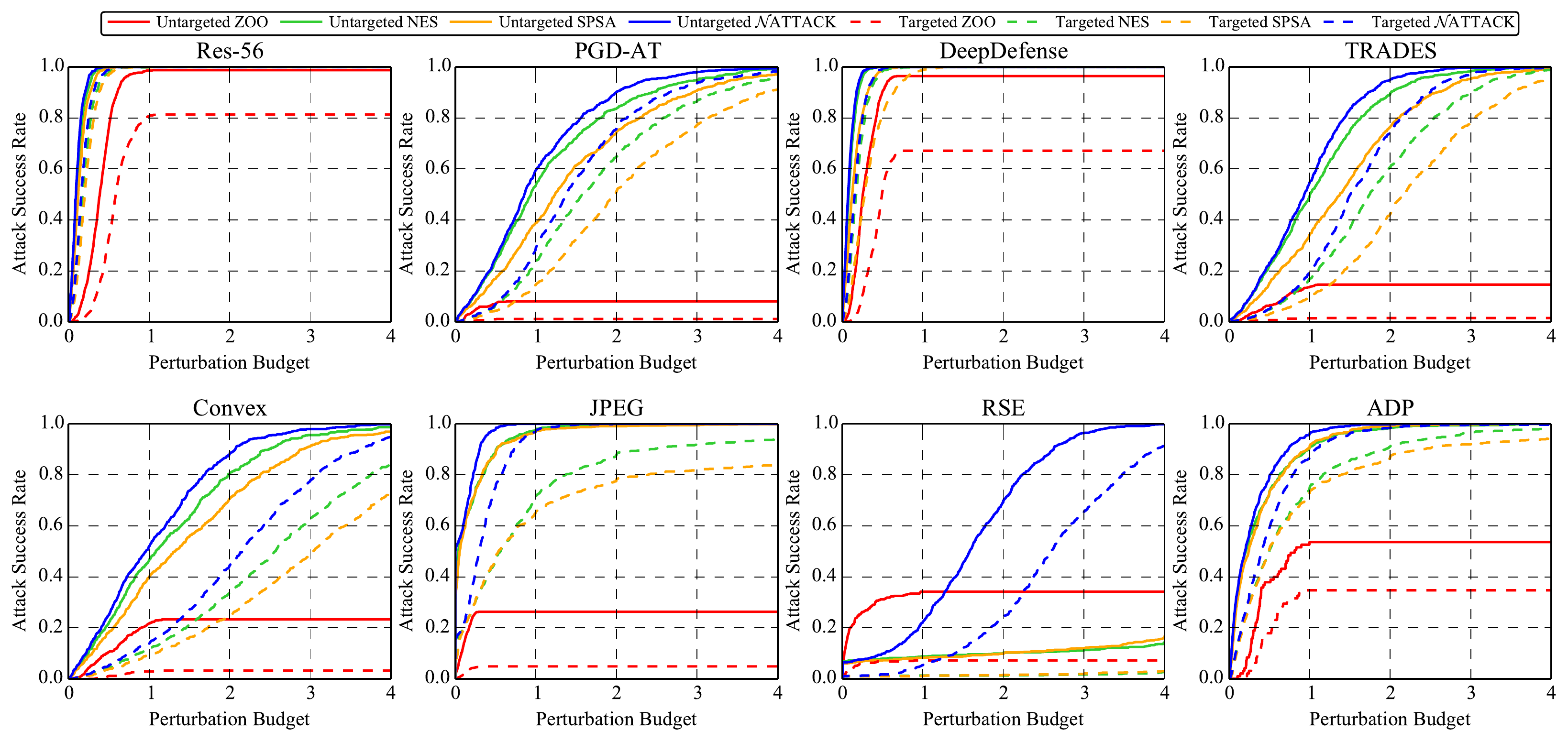}
\end{center}
\vspace{-3ex}
\caption{The \textit{attack success rate vs. perturbation budget} curves of score-based attacks under the $\ell_2$ norm on the $8$ models on CIFAR-10.}
\label{fig:score-l2-cifar10-asr-pert}
\end{figure*}

\begin{figure*}[t]
\begin{center}
\includegraphics[width=1.0\linewidth]{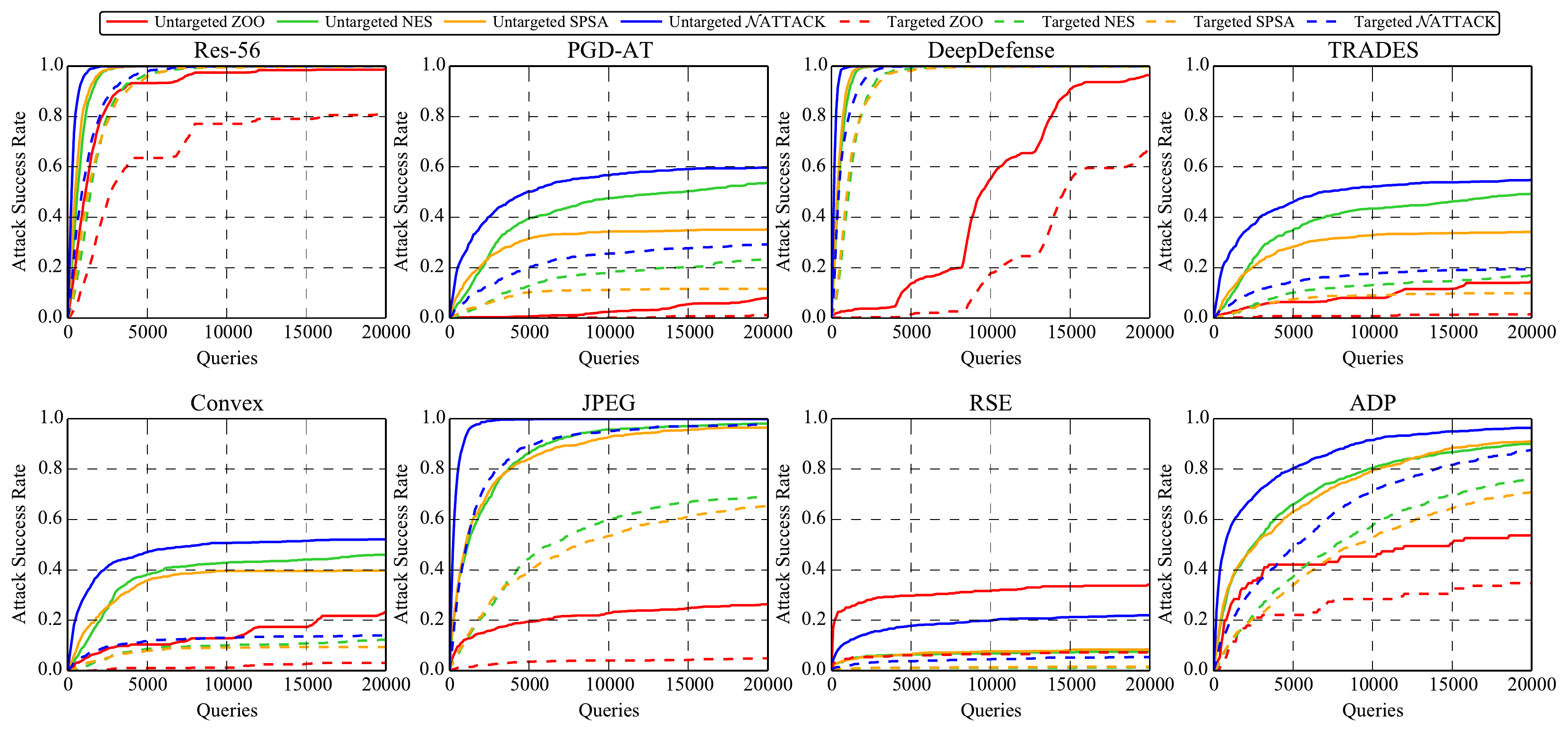}
\end{center}
\vspace{-3ex}
\caption{The \textit{attack success rate vs. attack strength} curves of score-based attacks under the $\ell_2$ norm on the $8$ models on CIFAR-10.}
\label{fig:score-l2-cifar10-asr-iter}
\end{figure*}

\begin{figure*}[t]
\begin{center}
\includegraphics[width=1.0\linewidth]{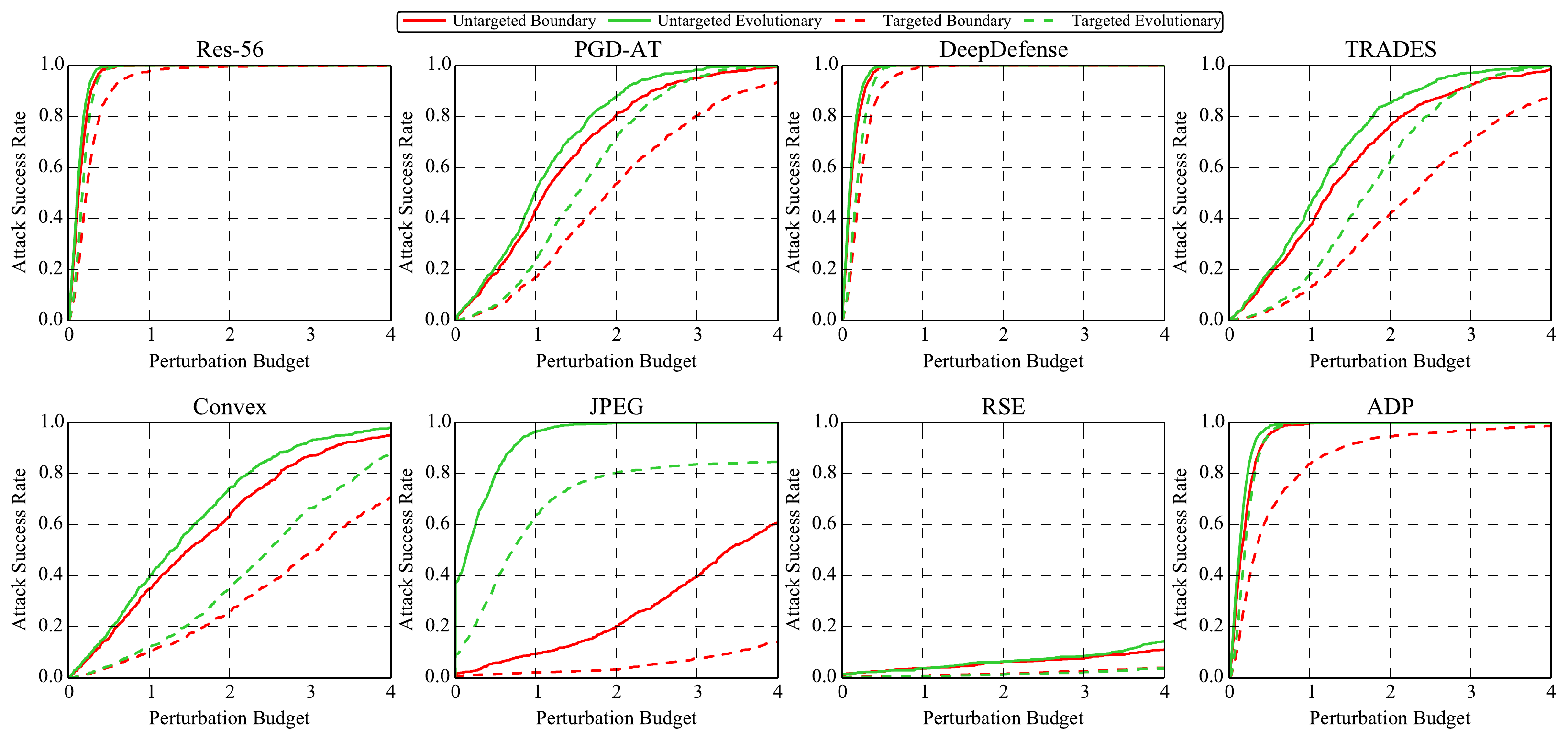}
\end{center}
\vspace{-3ex}
\caption{The \textit{attack success rate vs. perturbation budget} curves of decision-based attacks under the $\ell_2$ norm on the $8$ models on CIFAR-10.}
\label{fig:decision-l2-cifar10-asr-pert}
\end{figure*}

\begin{figure*}[t]
\begin{center}
\includegraphics[width=1.0\linewidth]{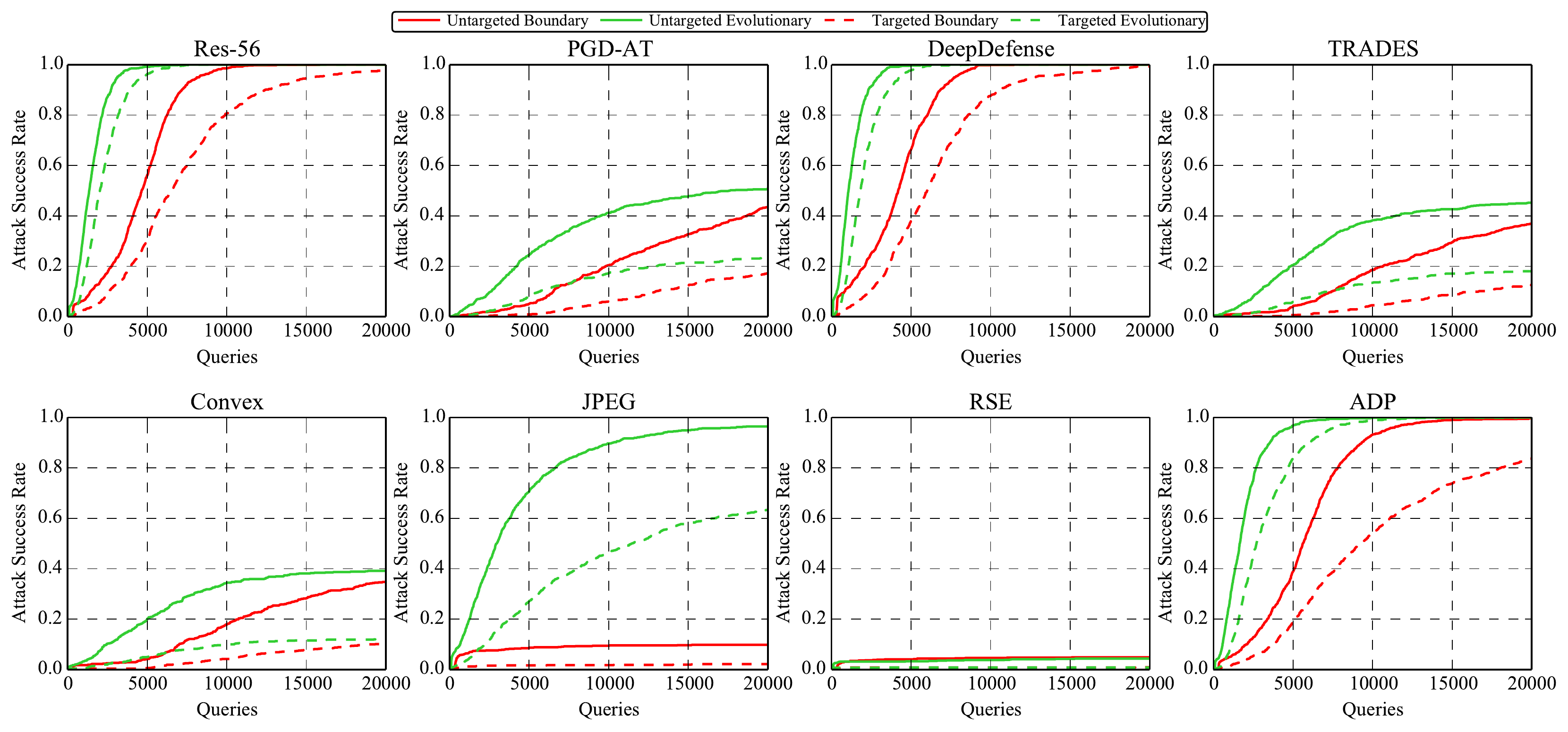}
\end{center}
\vspace{-3ex}
\caption{The \textit{attack success rate vs. attack strength} curves of decision-based attacks under the $\ell_2$ norm on the $8$ models on CIFAR-10.}
\label{fig:decision-l2-cifar10-asr-iter}
\end{figure*}

\textbf{Attacks under the $\ell_2$ norm:} We show the accuracy curves of the defenses on CIFAR-10 against untargeted and targeted white-box attacks under the $\ell_2$ norm in Fig.~\ref{fig:white-ut-l2-cifar10-acc-pert}, Fig.~\ref{fig:white-ut-l2-cifar10-acc-iter}, Fig.~\ref{fig:white-t-l2-cifar10-acc-pert}, and Fig.~\ref{fig:white-t-l2-cifar10-acc-iter}.
We show the accuracy curves of the defenses on CIFAR-10 against untargeted and targeted transfer-based attacks under the $\ell_2$ norm in Fig.~\ref{fig:trans-ut-l2-cifar10-acc-pert}, Fig.~\ref{fig:trans-ut-l2-cifar10-acc-iter}, Fig.~\ref{fig:trans-t-l2-cifar10-acc-pert}, and Fig.~\ref{fig:trans-t-l2-cifar10-acc-iter}.
We show the accuracy curves of the defenses on CIFAR-10 against untargeted and targeted score-based attacks under the $\ell_2$ norm in Fig.~\ref{fig:score-ut-l2-cifar10-acc-pert}, Fig.~\ref{fig:score-ut-l2-cifar10-acc-iter}, Fig.~\ref{fig:score-t-l2-cifar10-acc-pert}, and Fig.~\ref{fig:score-t-l2-cifar10-acc-iter}.
We show the accuracy curves of the defenses on CIFAR-10 against untargeted and targeted decision-based attacks under the $\ell_2$ norm in Fig.~\ref{fig:decision-ut-l2-cifar10-acc-pert}, Fig.~\ref{fig:decision-ut-l2-cifar10-acc-iter}, Fig.~\ref{fig:decision-t-l2-cifar10-acc-pert}, and Fig.~\ref{fig:decision-t-l2-cifar10-acc-iter}.
Fig.~\ref{fig:white-l2-cifar10-asr-pert} to Fig.~\ref{fig:decision-l2-cifar10-asr-iter} show the \textit{attack success rate vs. perturbation budget} and \textit{attack success rate vs. attack strength} curves of white-box, transfer-based, score-based, and decision-based attacks under the $\ell_2$ norm on the $8$ models on CIFAR-10.

\subsection{Full Evaluation Results on ImageNet}

\begin{figure*}[t]
\begin{center}
\includegraphics[width=0.55\linewidth]{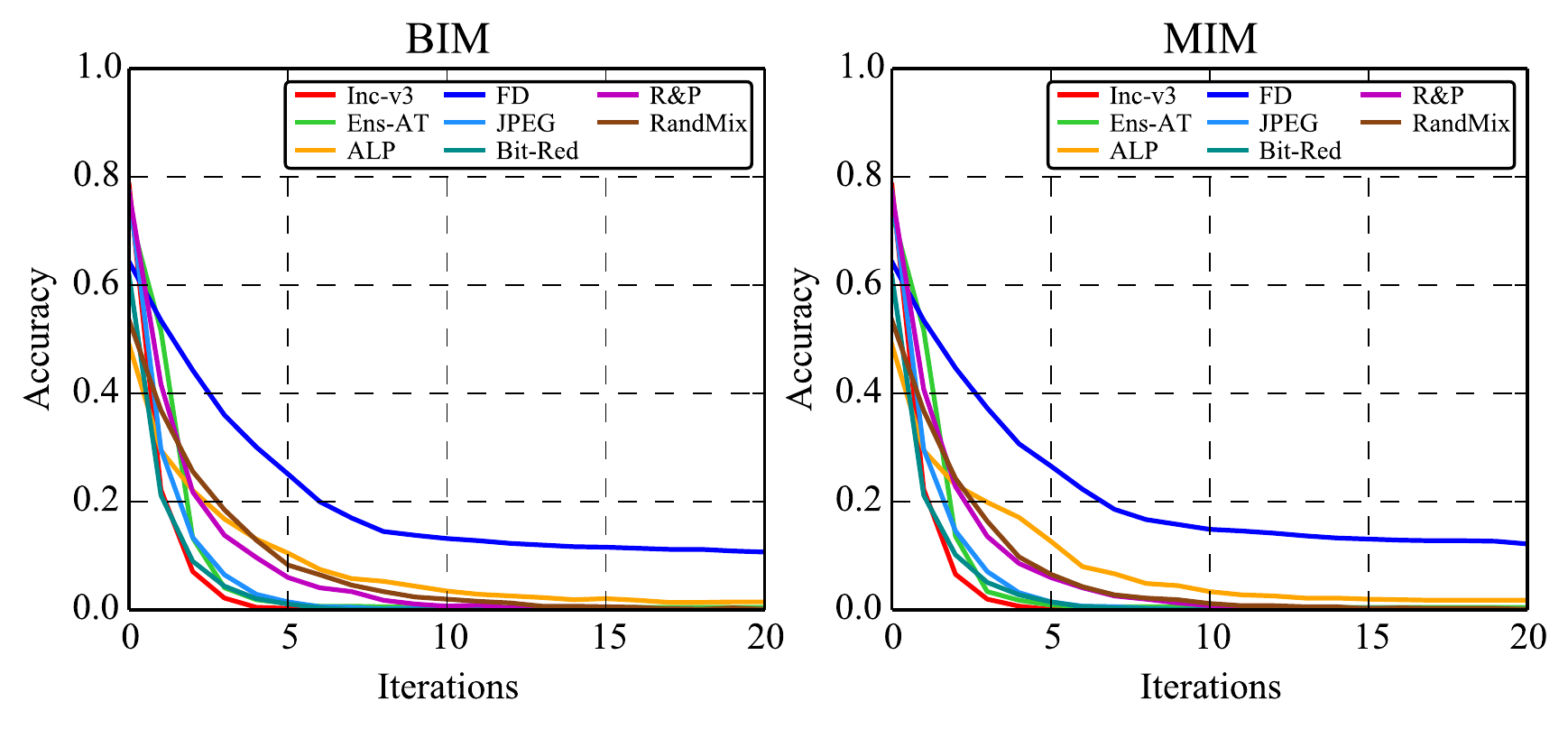}
\end{center}
\vspace{-2ex}
\caption{The \textit{accuracy vs. attack strength} curves of the $8$ models on ImageNet against untargeted white-box attacks under the $\ell_{\infty}$ norm.}
\label{fig:white-ut-linf-imagenet-acc-iter}
\end{figure*}

\begin{figure*}[t]
\begin{center}
\includegraphics[width=0.85\linewidth]{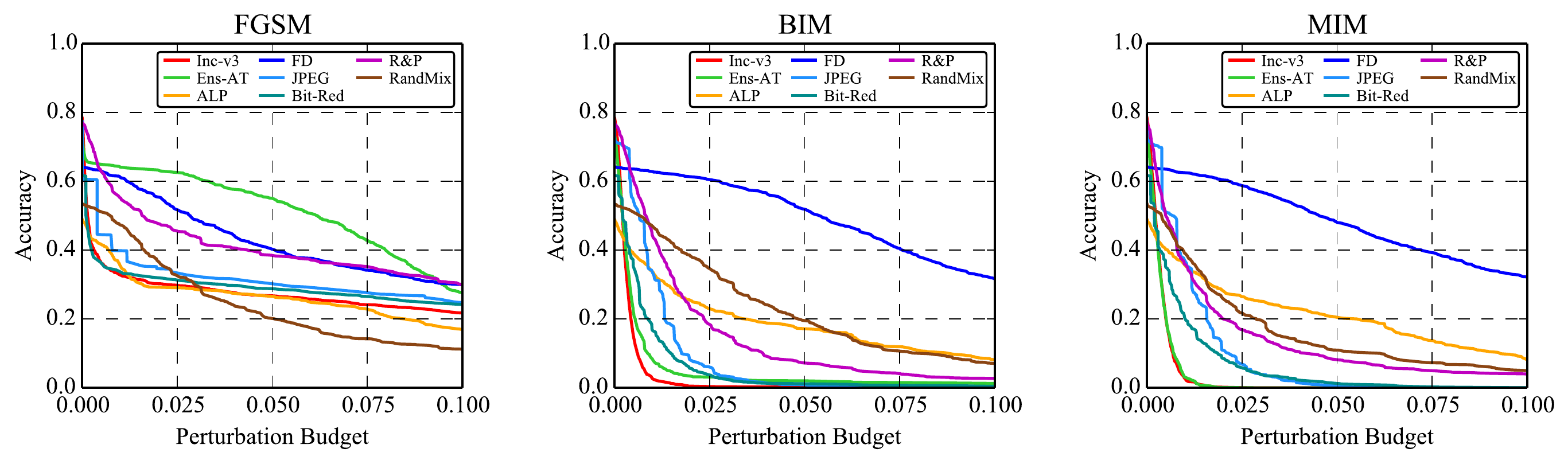}
\end{center}
\vspace{-2ex}
\caption{The \textit{accuracy vs. perturbation budget} curves of the $8$ models on ImageNet against targeted white-box attacks under the $\ell_{\infty}$ norm.}
\label{fig:white-t-linf-imagenet-acc-pert}
\begin{center}
\includegraphics[width=0.55\linewidth]{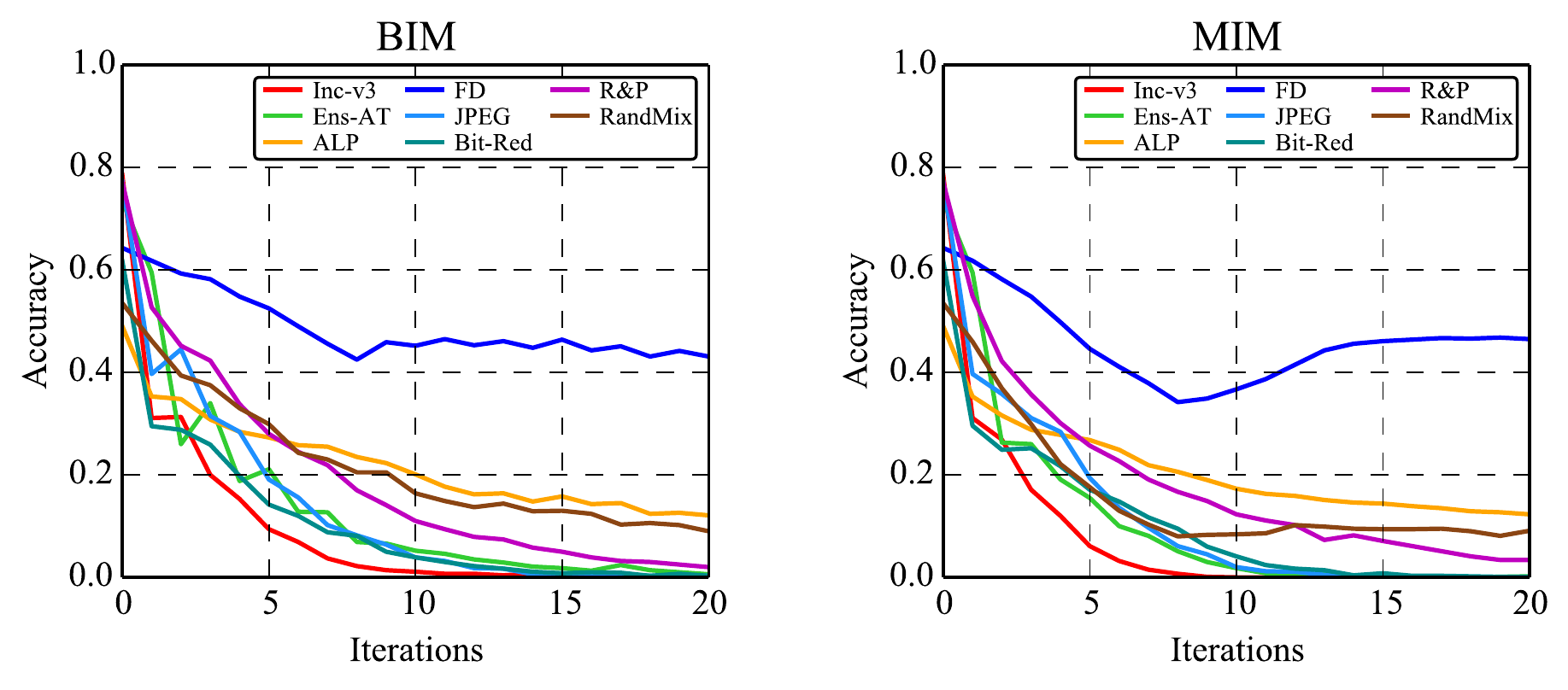}
\end{center}
\vspace{-2ex}
\caption{The \textit{accuracy vs. attack strength} curves of the $8$ models on ImageNet against targeted white-box attacks under the $\ell_{\infty}$ norm.}
\label{fig:white-t-linf-imagenet-acc-iter}
\end{figure*}

\begin{figure*}[t]
\begin{center}
\includegraphics[width=0.85\linewidth]{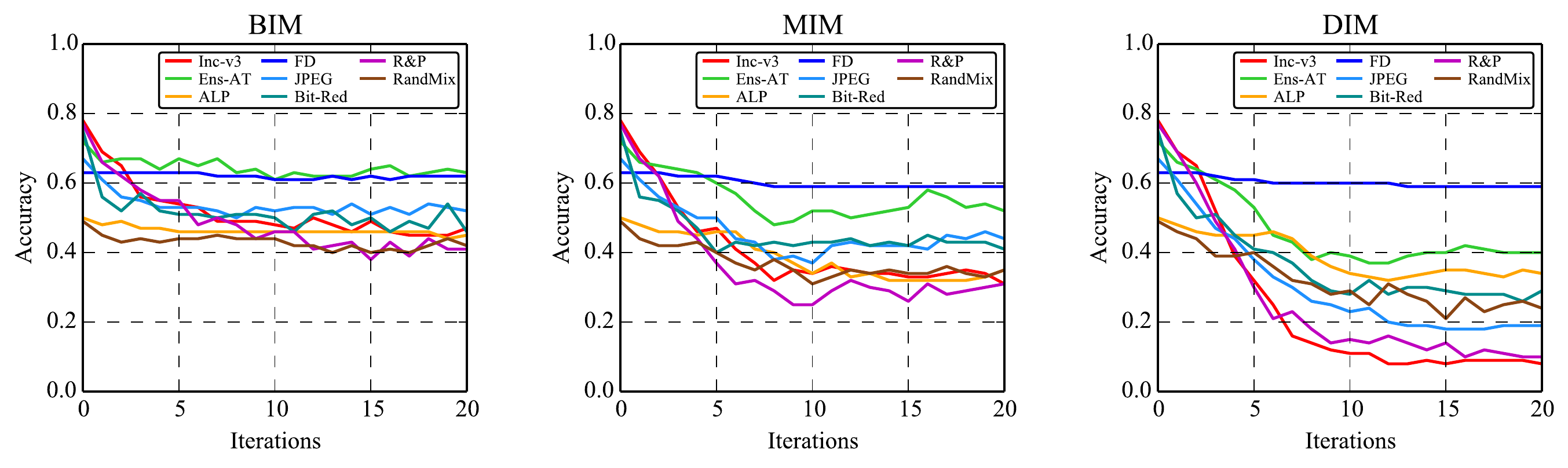}
\end{center}
\vspace{-2ex}
\caption{The \textit{accuracy vs. attack strength} curves of the $8$ models on ImageNet against untargeted transfer-based attacks under the $\ell_{\infty}$ norm.}
\label{fig:trans-ut-linf-imagenet-acc-iter}
\end{figure*}

\begin{figure*}[t]
\begin{center}
\includegraphics[width=1.0\linewidth]{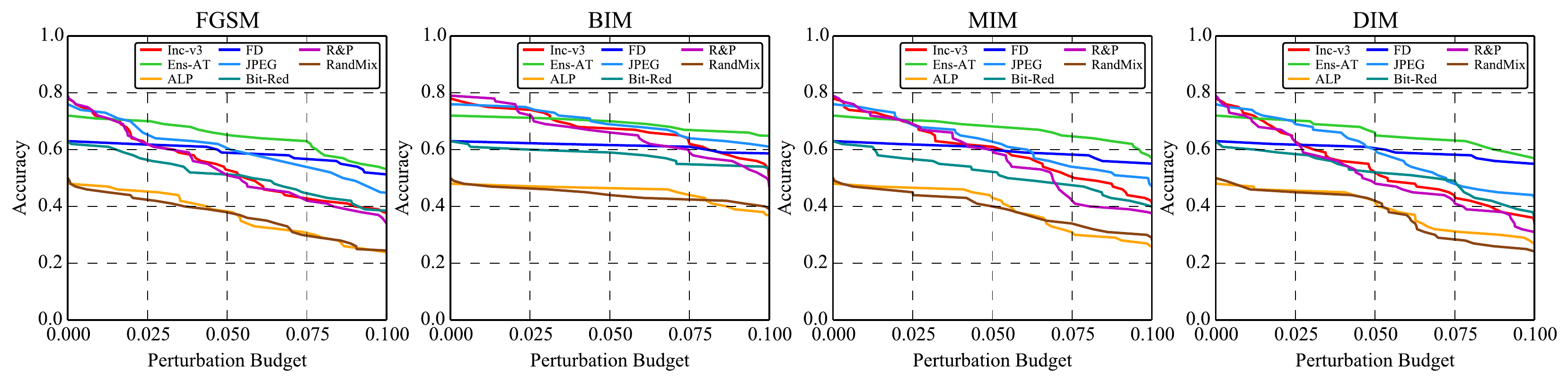}
\end{center}
\vspace{-2ex}
\caption{The \textit{accuracy vs. perturbation budget} curves of the $8$ models on ImageNet against targeted transfer-based attacks under the $\ell_{\infty}$ norm.}
\label{fig:trans-t-linf-imagenet-acc-pert}
\begin{center}
\includegraphics[width=0.85\linewidth]{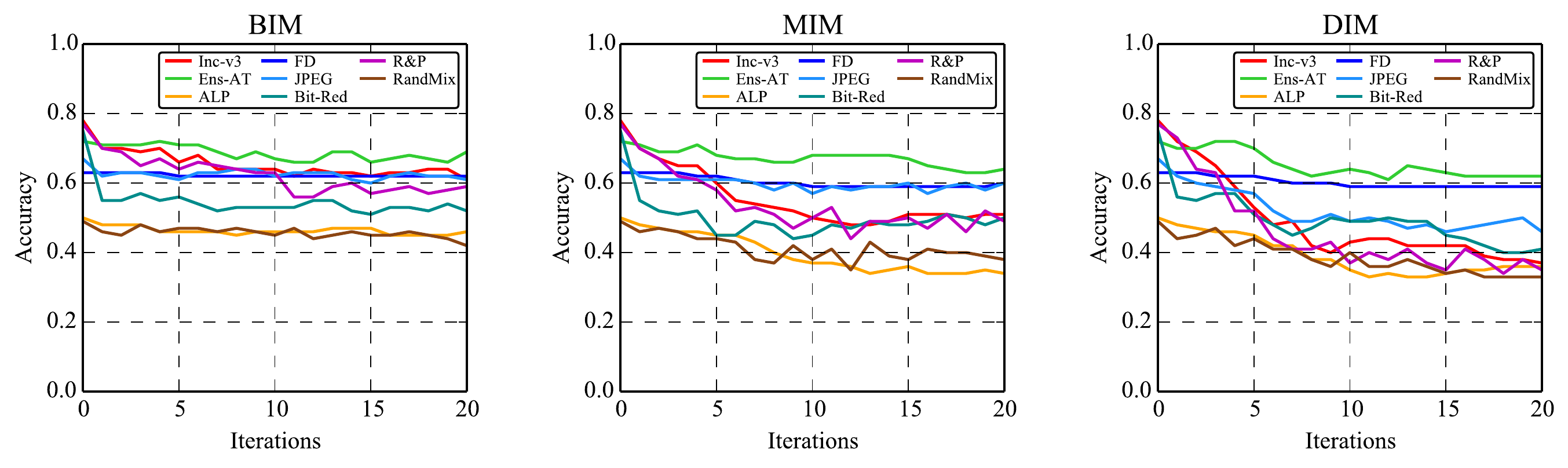}
\end{center}
\vspace{-2ex}
\caption{The \textit{accuracy vs. attack strength} curves of the $8$ models on ImageNet against targeted transfer-based attacks under the $\ell_{\infty}$ norm.}
\label{fig:trans-t-linf-imagenet-acc-iter}
\end{figure*}

\begin{figure*}[t]
\begin{center}
\includegraphics[width=0.85\linewidth]{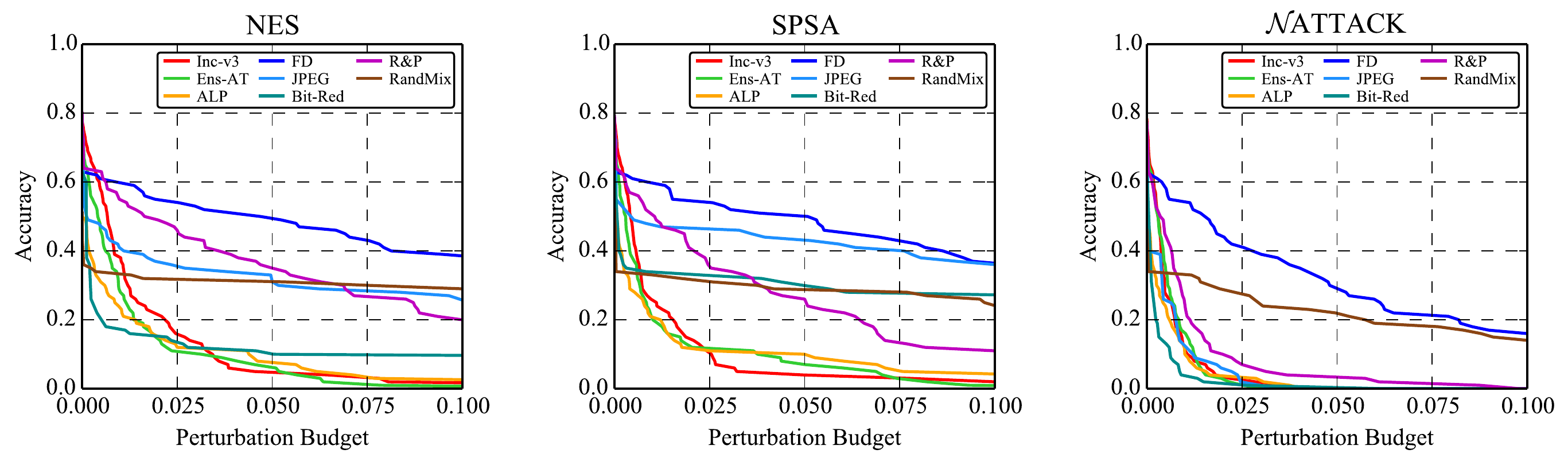}
\end{center}
\vspace{-2ex}
\caption{The \textit{accuracy vs. perturbation budget} curves of the $8$ models on ImageNet against targeted score-based attacks under the $\ell_{\infty}$ norm.}
\label{fig:score-t-linf-imagenet-acc-pert}
\begin{center}
\includegraphics[width=0.85\linewidth]{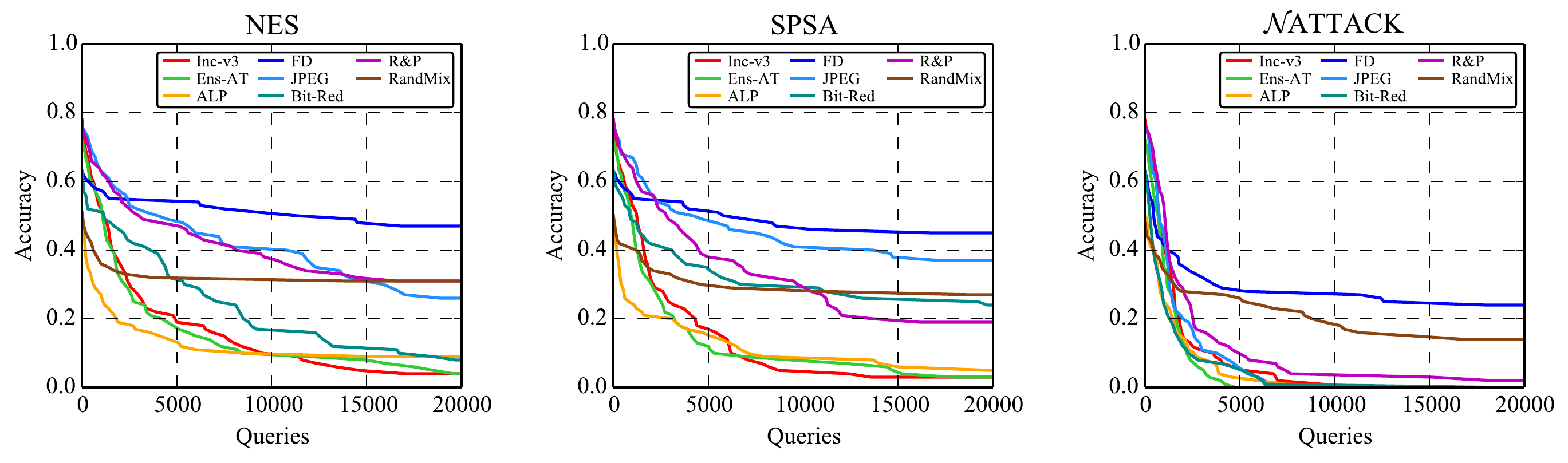}
\end{center}
\vspace{-2ex}
\caption{The \textit{accuracy vs. attack strength} curves of the $8$ models on ImageNet against targeted score-based attacks under the $\ell_{\infty}$ norm.}
\label{fig:score-t-linf-imagenet-acc-iter}
\end{figure*}

\begin{figure*}[t]
\begin{center}
\includegraphics[width=1.0\linewidth]{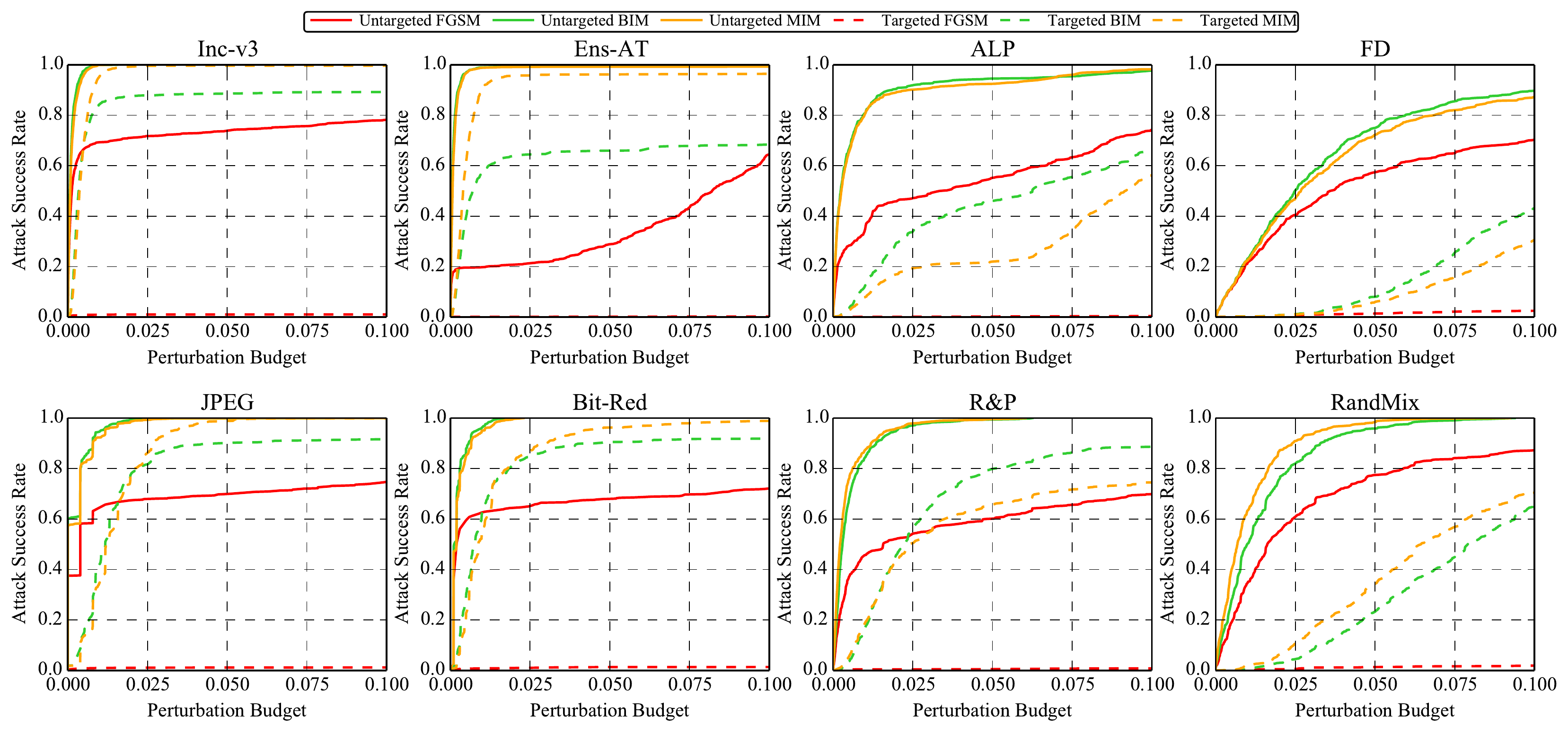}
\end{center}
\vspace{-2ex}
\caption{The \textit{attack success rate vs. perturbation budget} curves of white-box attacks under the $\ell_{\infty}$ norm on the $8$ models on ImageNet.}
\label{fig:white-linf-imagenet-asr-pert}
\end{figure*}

\begin{figure*}[t]
\begin{center}
\includegraphics[width=1.0\linewidth]{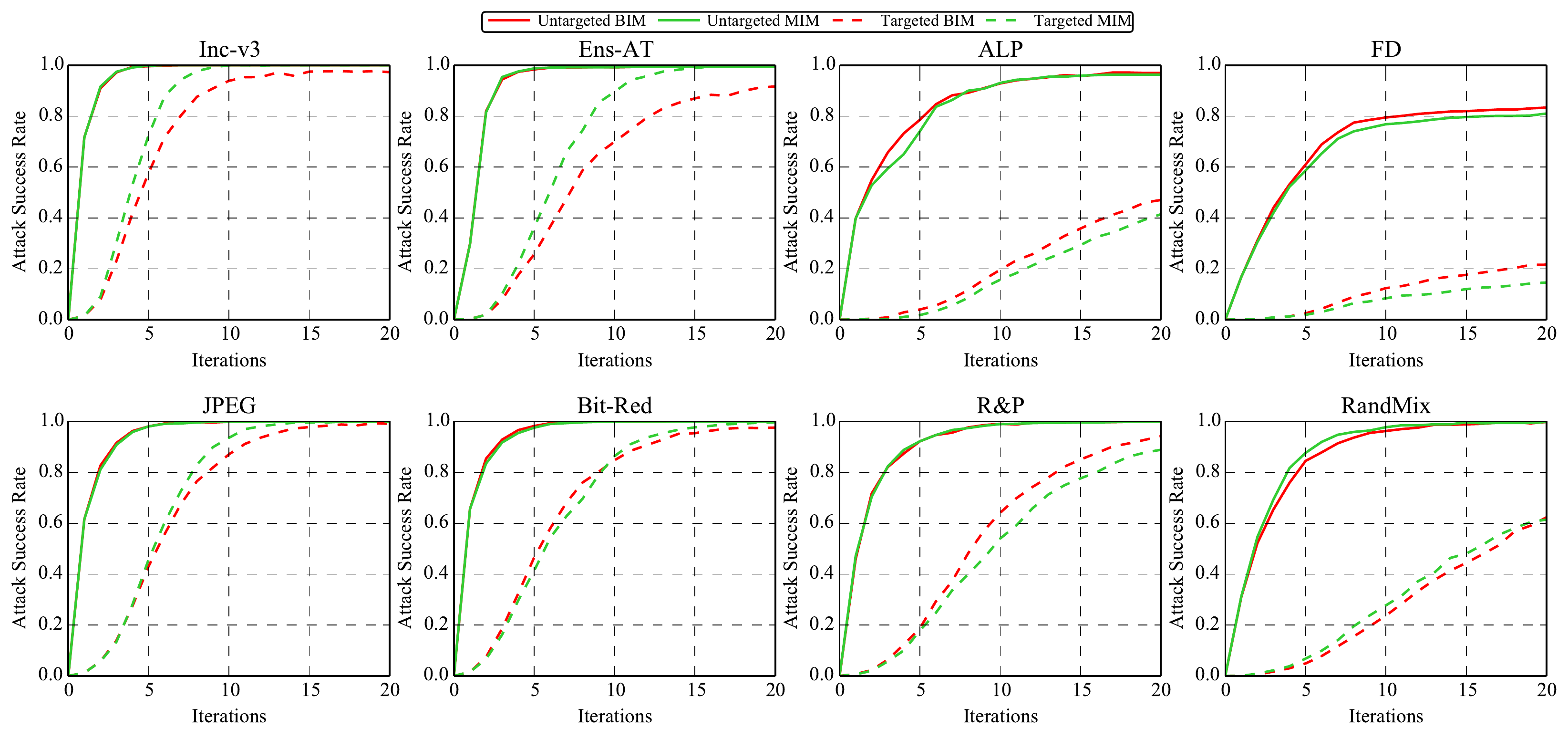}
\end{center}
\vspace{-2ex}
\caption{The \textit{attack success rate vs. attack strength} curves of white-box attacks under the $\ell_{\infty}$ norm on the $8$ models on ImageNet.}
\label{fig:white-linf-imagenet-asr-iter}
\end{figure*}

\begin{figure*}[t]
\begin{center}
\includegraphics[width=1.0\linewidth]{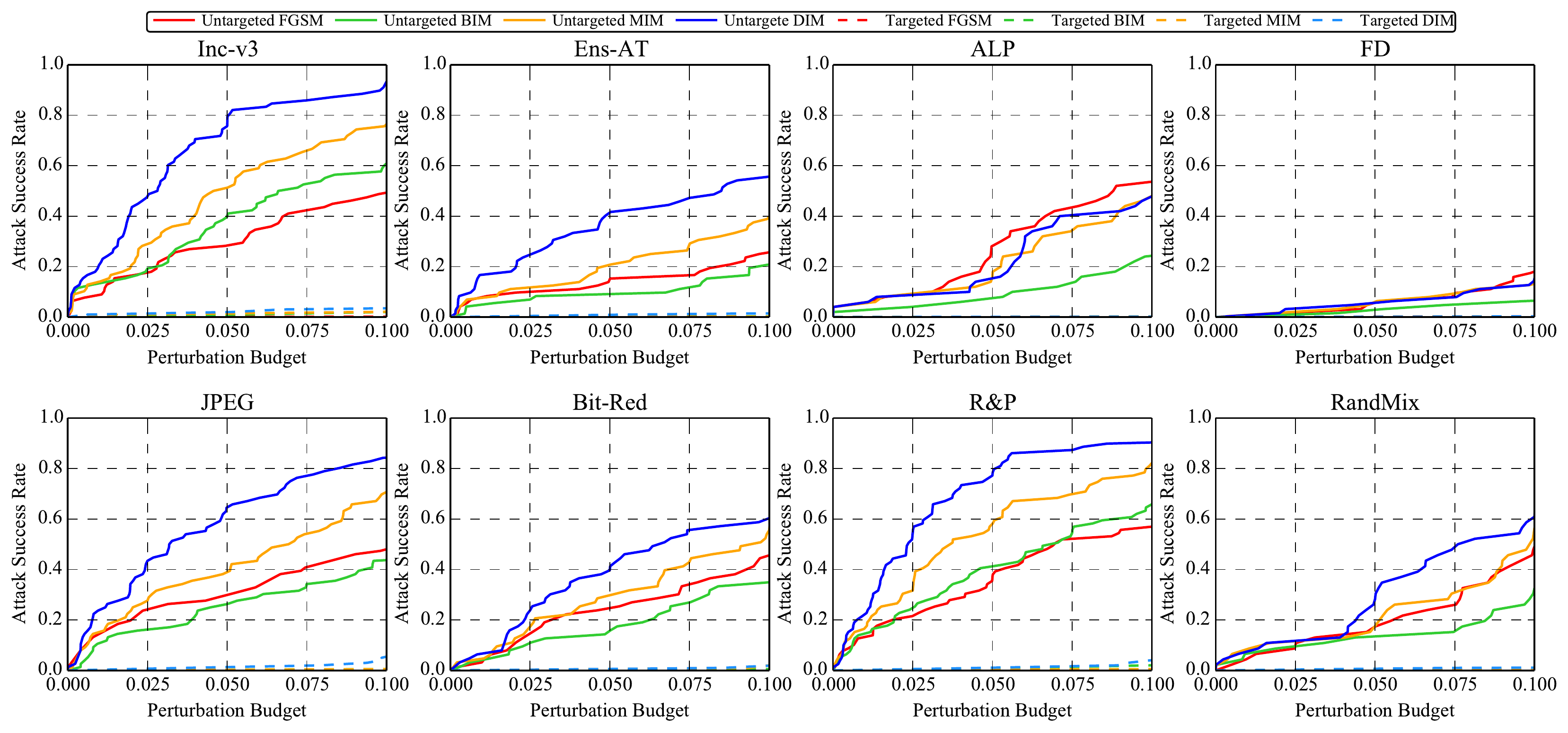}
\end{center}
\vspace{-2ex}
\caption{The \textit{attack success rate vs. perturbation budget} curves of transfer-based attacks under the $\ell_{\infty}$ norm on the $8$ models on ImageNet.}
\label{fig:trans-linf-imagenet-asr-pert}
\end{figure*}

\begin{figure*}[t]
\begin{center}
\includegraphics[width=1.0\linewidth]{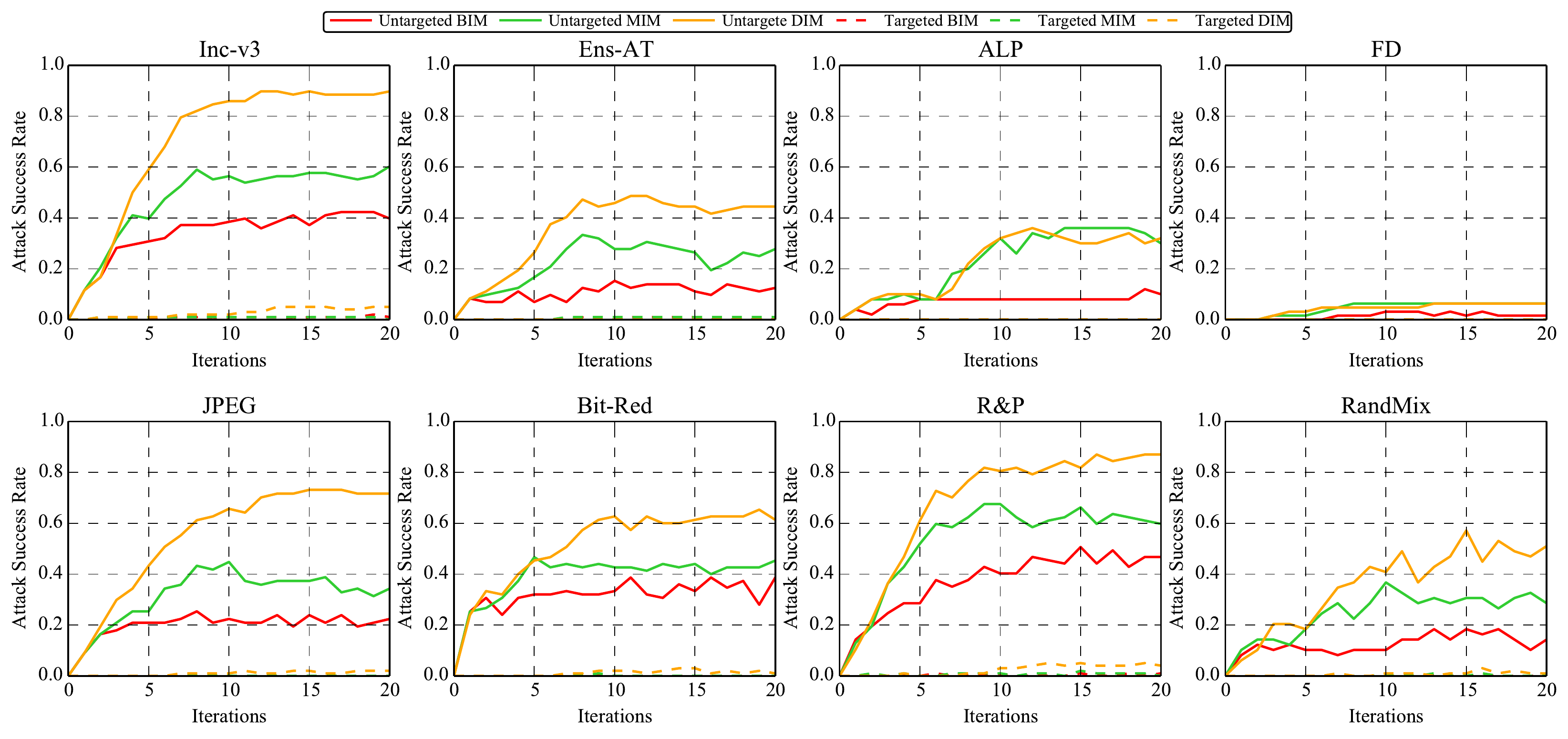}
\end{center}
\vspace{-2ex}
\caption{The \textit{attack success rate vs. attack strength} curves of transfer-based attacks under the $\ell_{\infty}$ norm on the $8$ models on ImageNet.}
\label{fig:trans-linf-imagenet-asr-iter}
\end{figure*}

\begin{figure*}[t]
\begin{center}
\includegraphics[width=1.0\linewidth]{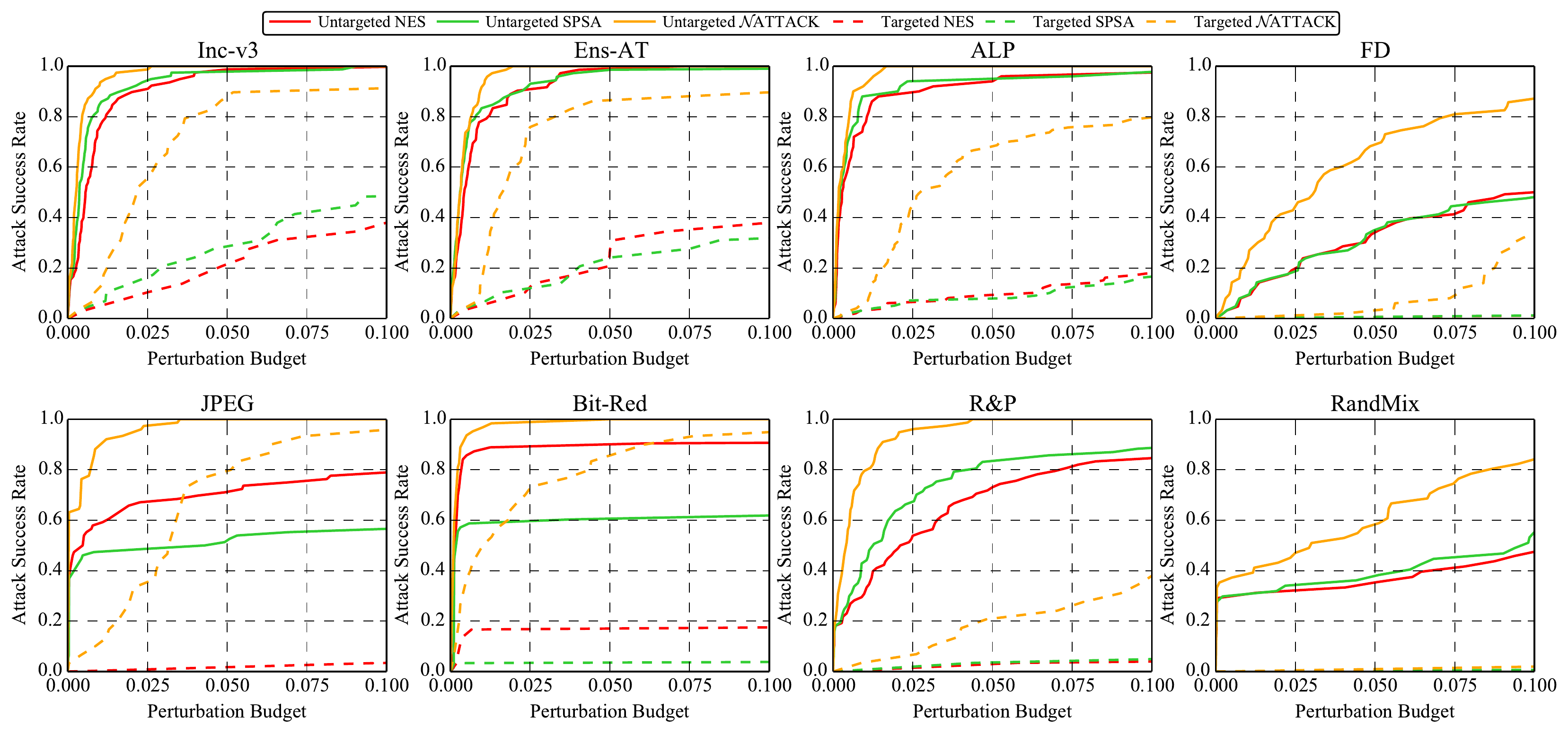}
\end{center}
\vspace{-2ex}
\caption{The \textit{attack success rate vs. perturbation budget} curves of score-based attacks under the $\ell_{\infty}$ norm on the $8$ models on ImageNet.}
\label{fig:score-linf-imagenet-asr-pert}
\end{figure*}

\begin{figure*}[t]
\begin{center}
\includegraphics[width=1.0\linewidth]{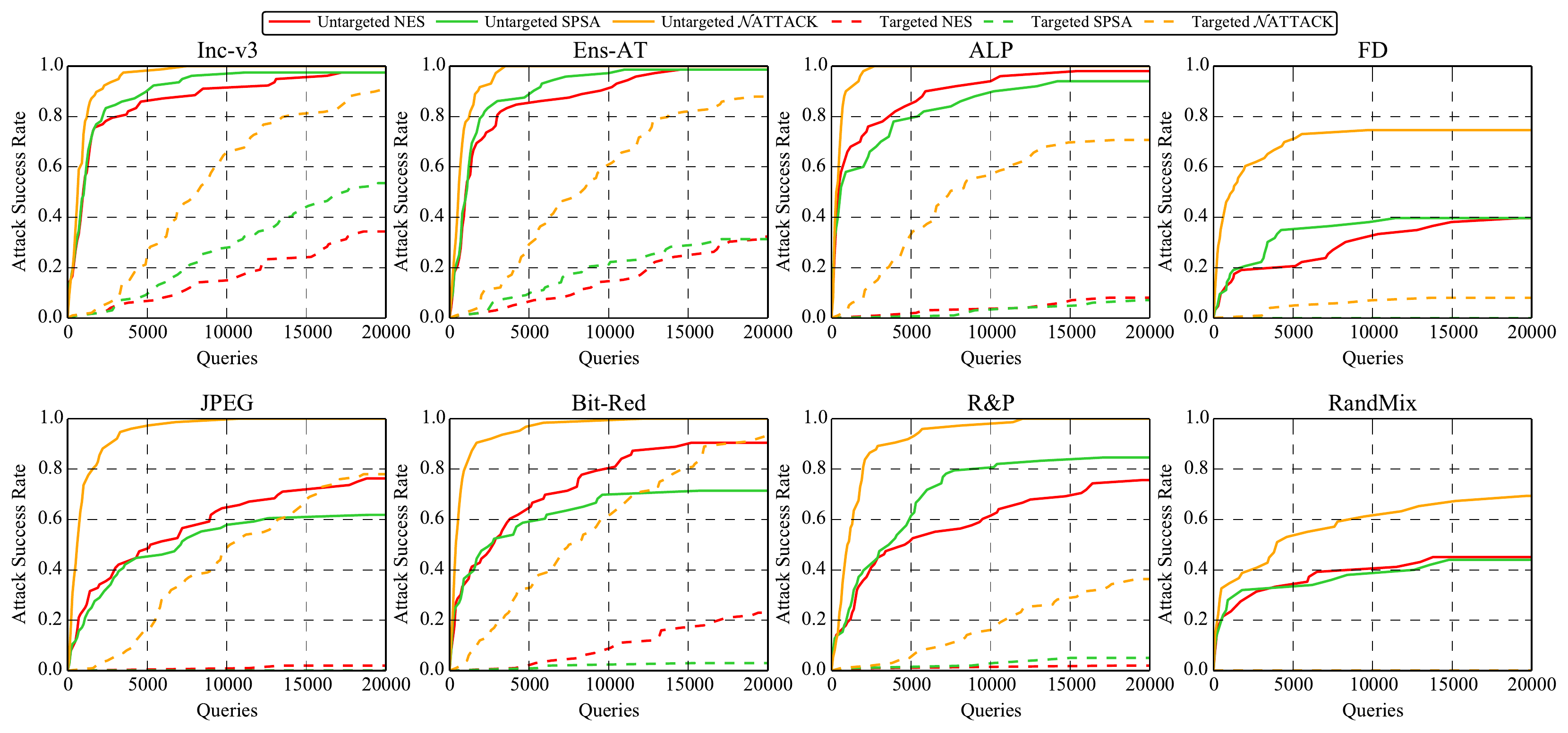}
\end{center}
\vspace{-2ex}
\caption{The \textit{attack success rate vs. attack strength} curves of score-based attacks under the $\ell_{\infty}$ norm on the $8$ models on ImageNet.}
\label{fig:score-linf-imagenet-asr-iter}
\end{figure*}

\textbf{Attacks under the $\ell_{\infty}$ norm}: Similar to CIFAR-10, we show the results of the remaining untargeted attacks, targeted attacks under the $\ell_{\infty}$ norm, and the attacks success rate curves here.
Fig.~\ref{fig:white-ut-linf-imagenet-acc-iter} shows the \emph{accuracy vs. attack strength} curves of the defenses on ImageNet against untargeted white-box attacks under the $\ell_{\infty}$ norm.
Fig.~\ref{fig:white-t-linf-imagenet-acc-pert} and Fig.~\ref{fig:white-t-linf-imagenet-acc-iter} show the accuracy curves of the defenses on ImageNet against targeted white-box attacks under the $\ell_{\infty}$ norm.
Fig.~\ref{fig:trans-ut-linf-imagenet-acc-iter} shows the \emph{accuracy vs. attack strength} curves of the defenses on ImageNet against untargeted transfer-based attacks under the $\ell_{\infty}$ norm.
Fig.~\ref{fig:trans-t-linf-imagenet-acc-pert} and Fig.~\ref{fig:trans-t-linf-imagenet-acc-iter} show the accuracy curves of the defenses on ImageNet against targeted transfer-based attacks under the $\ell_{\infty}$ norm.
Fig.~\ref{fig:score-t-linf-imagenet-acc-pert} and Fig.~\ref{fig:score-t-linf-imagenet-acc-iter} show the accuracy curves of the defenses on ImageNet against targeted score-based attacks under the $\ell_{\infty}$ norm.
Fig.~\ref{fig:white-linf-imagenet-asr-pert} to Fig.~\ref{fig:score-linf-imagenet-asr-iter} show the \textit{attack success rate vs. perturbation budget} and \textit{attack success rate vs. attack strength} curves of white-box, transfer-based, and score-based attacks under the $\ell_{\infty}$ norm on the $8$ models on ImageNet.

\begin{figure*}[t]
\begin{center}
\includegraphics[width=1.0\linewidth]{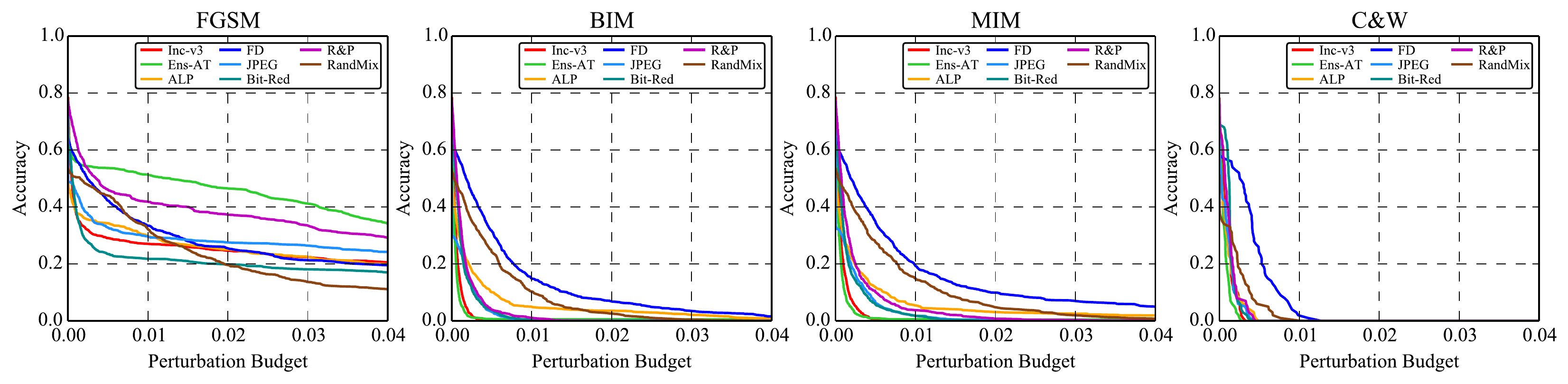}
\end{center}
\vspace{-2ex}
\caption{The \textit{accuracy vs. perturbation budget} curves of the $8$ models on ImageNet against untargeted white-box attacks under the $\ell_2$ norm.}
\label{fig:white-ut-l2-imagenet-acc-pert}
\begin{center}
\includegraphics[width=0.85\linewidth]{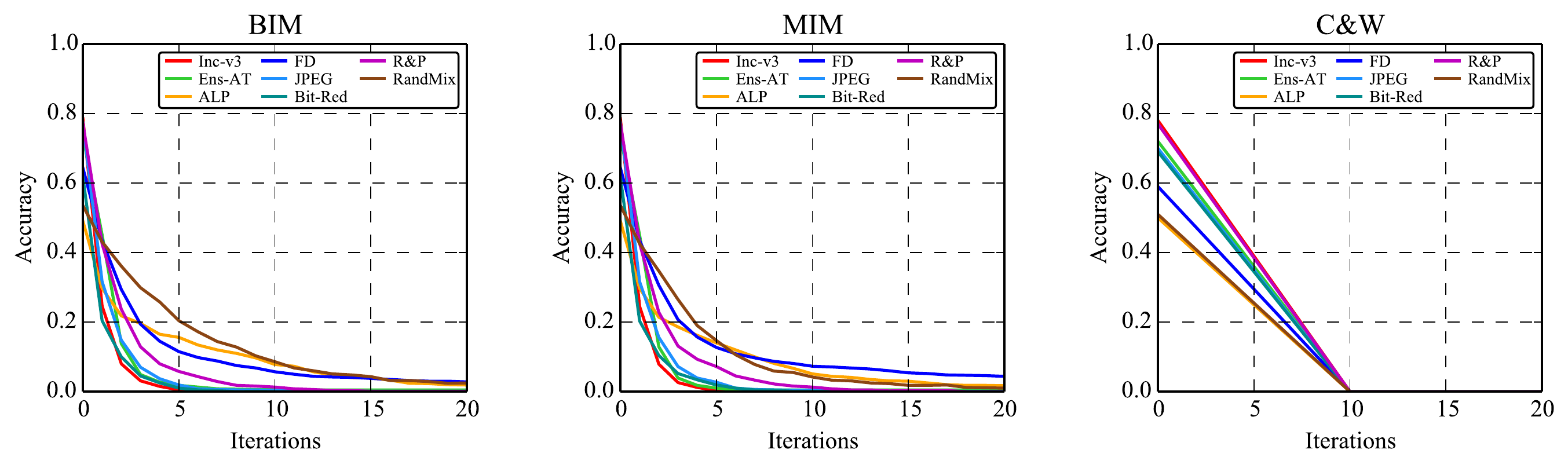}
\end{center}
\vspace{-2ex}
\caption{The \textit{accuracy vs. attack strength} curves of the $8$ models on ImageNet against untargeted white-box attacks under the $\ell_2$ norm.}
\label{fig:white-ut-l2-imagenet-acc-iter}
\end{figure*}

\begin{figure*}[t]
\begin{center}
\includegraphics[width=1.0\linewidth]{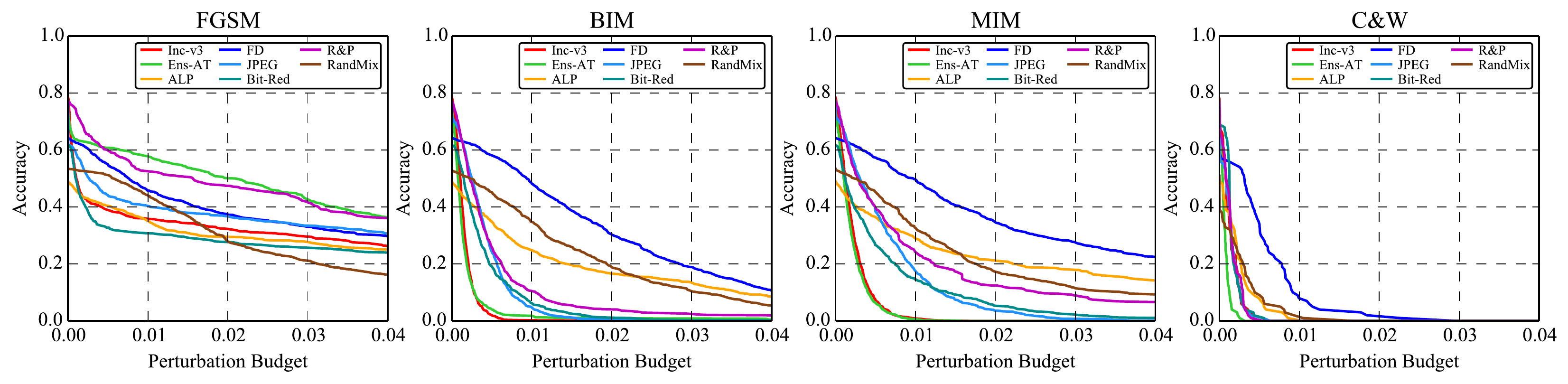}
\end{center}
\vspace{-2ex}
\caption{The \textit{accuracy vs. perturbation budget} curves of the $8$ models on ImageNet against targeted white-box attacks under the $\ell_2$ norm.}
\label{fig:white-t-l2-imagenet-acc-pert}
\begin{center}
\includegraphics[width=0.85\linewidth]{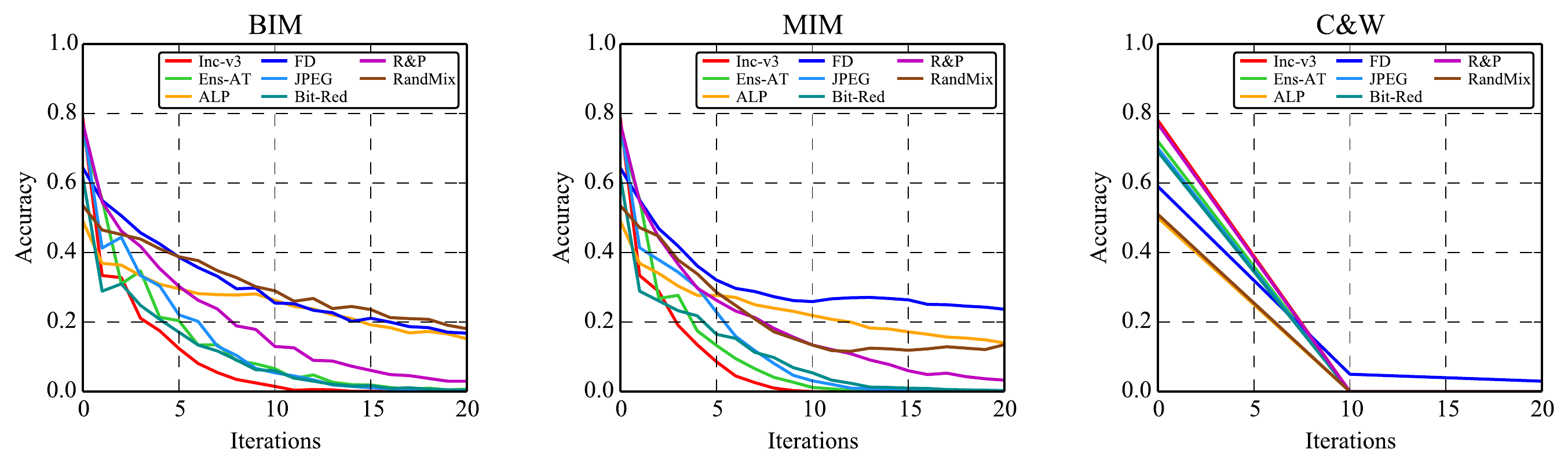}
\end{center}
\vspace{-2ex}
\caption{The \textit{accuracy vs. attack strength} curves of the $8$ models on ImageNet against targeted white-box attacks under the $\ell_2$ norm.}
\label{fig:white-t-l2-imagenet-acc-iter}
\end{figure*}

\begin{figure*}[t]
\begin{center}
\includegraphics[width=1.0\linewidth]{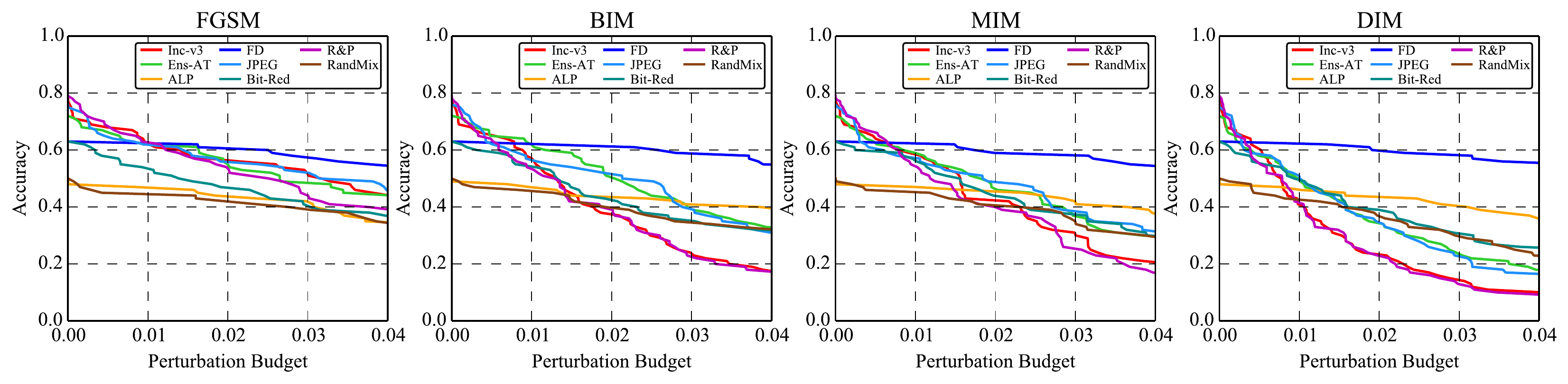}
\end{center}
\vspace{-2ex}
\caption{The \textit{accuracy vs. perturbation budget} curves of the $8$ models on ImageNet against untargeted transfer-based attacks under the $\ell_2$ norm.}
\label{fig:trans-ut-l2-imagenet-acc-pert}
\begin{center}
\includegraphics[width=0.85\linewidth]{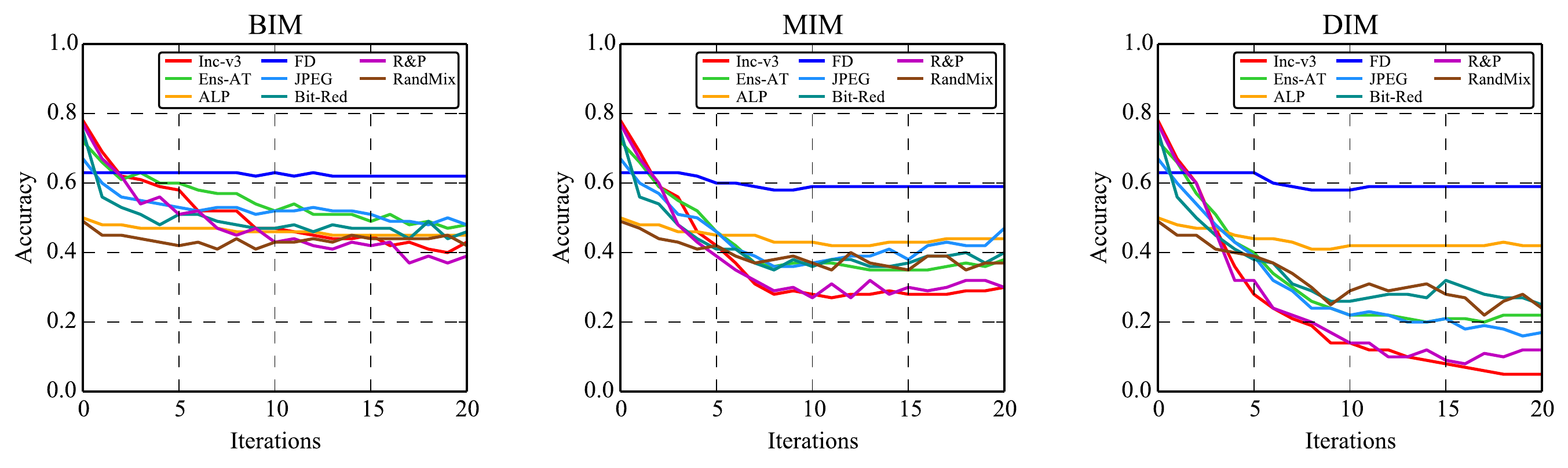}
\end{center}
\vspace{-2ex}
\caption{The \textit{accuracy vs. attack strength} curves of the $8$ models on ImageNet against untargeted transfer-based attacks under the $\ell_2$ norm.}
\label{fig:trans-ut-l2-imagenet-acc-iter}
\end{figure*}

\begin{figure*}[t]
\begin{center}
\includegraphics[width=1.0\linewidth]{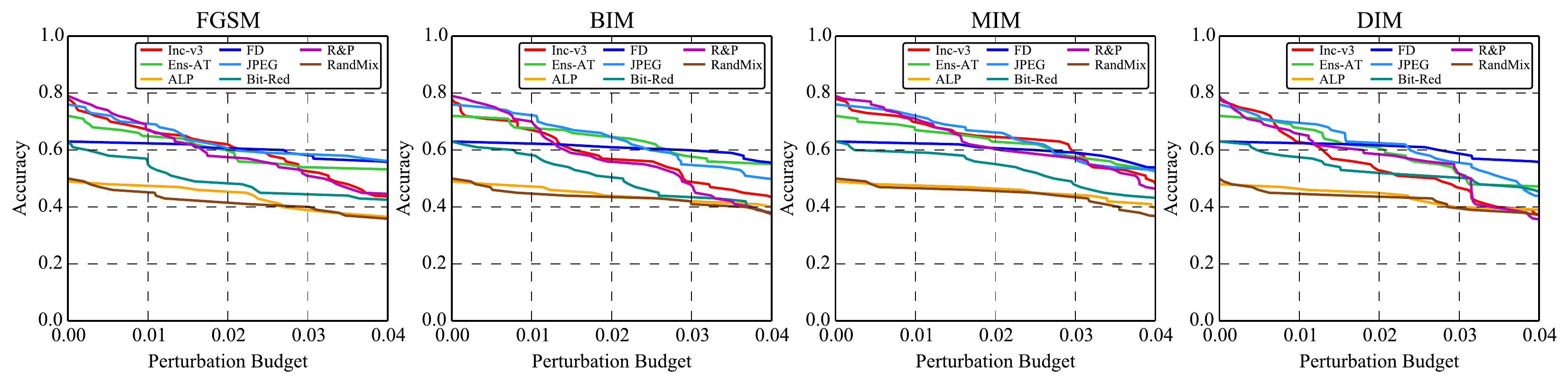}
\end{center}
\vspace{-2ex}
\caption{The \textit{accuracy vs. perturbation budget} curves of the $8$ models on ImageNet against targeted transfer-based attacks under the $\ell_2$ norm.}
\label{fig:trans-t-l2-imagenet-acc-pert}
\begin{center}
\includegraphics[width=0.85\linewidth]{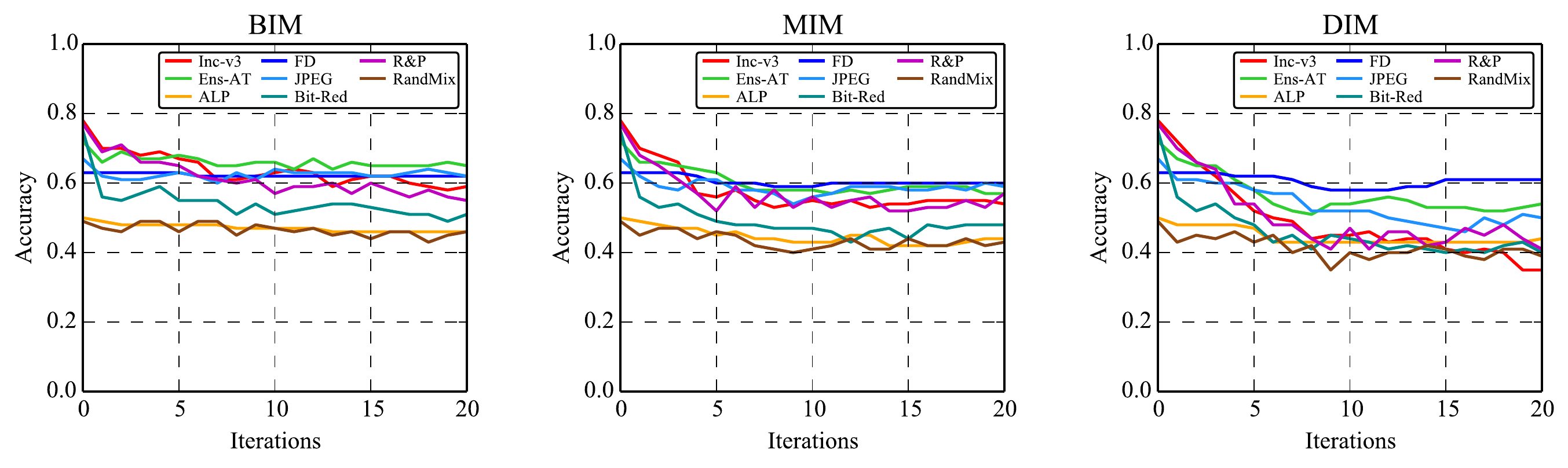}
\end{center}
\vspace{-2ex}
\caption{The \textit{accuracy vs. attack strength} curves of the $8$ models on ImageNet against targeted transfer-based attacks under the $\ell_2$ norm.}
\label{fig:trans-t-l2-imagenet-acc-iter}
\end{figure*}

\begin{figure*}[t]
\begin{center}
\includegraphics[width=0.85\linewidth]{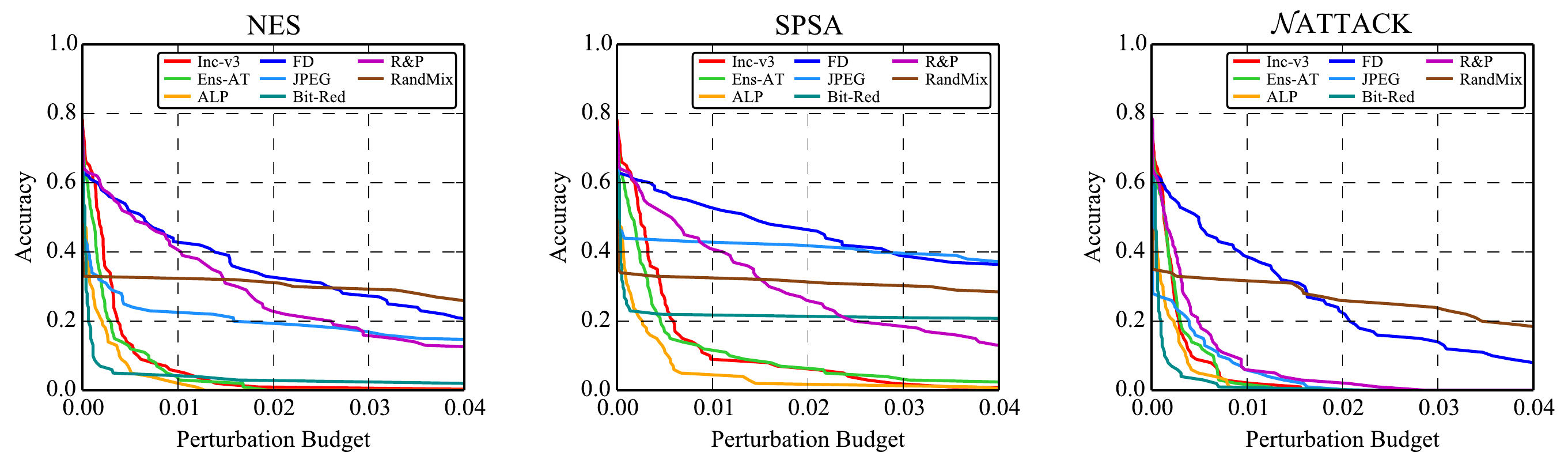}
\end{center}
\vspace{-2ex}
\caption{The \textit{accuracy vs. perturbation budget} curves of the $8$ models on ImageNet against untargeted score-based attacks under the $\ell_2$ norm.}
\label{fig:score-ut-l2-imagenet-acc-pert}
\begin{center}
\includegraphics[width=0.85\linewidth]{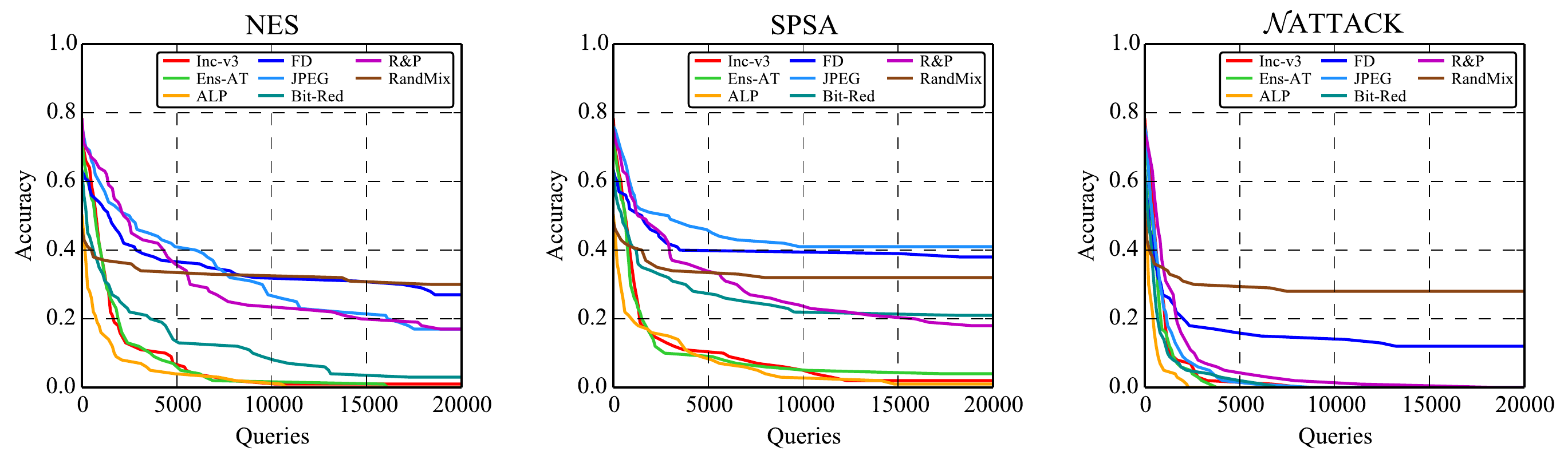}
\end{center}
\vspace{-2ex}
\caption{The \textit{accuracy vs. attack strength} curves of the $8$ models on ImageNet against untargeted score-based attacks under the $\ell_2$ norm.}
\label{fig:score-ut-l2-imagenet-acc-iter}
\end{figure*}

\begin{figure*}[t]
\begin{center}
\includegraphics[width=0.85\linewidth]{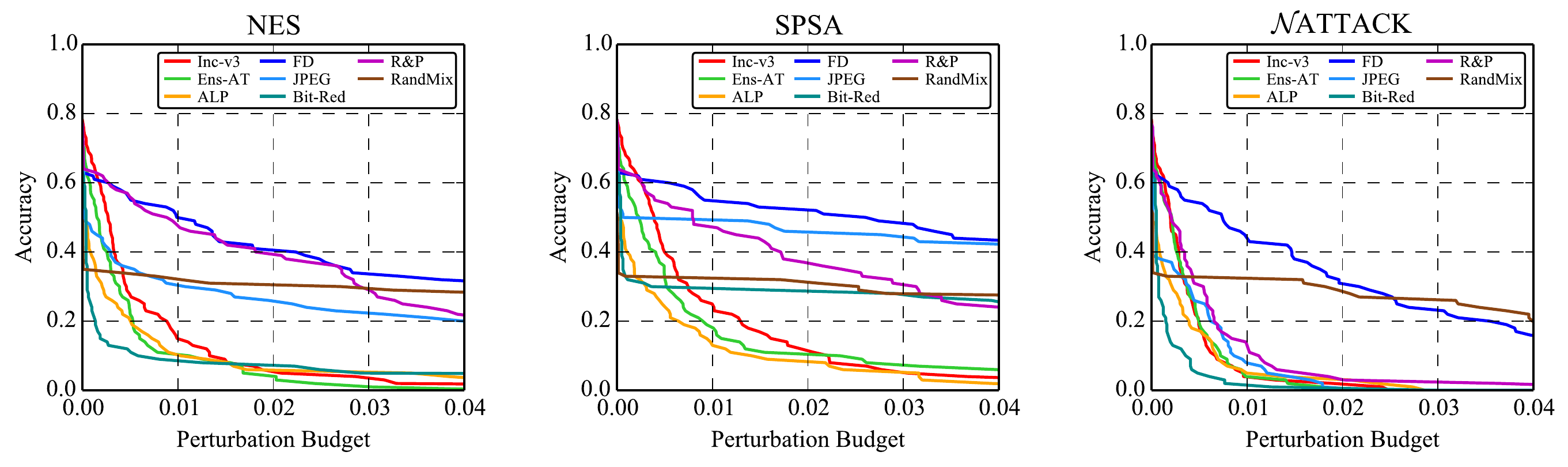}
\end{center}
\vspace{-2ex}
\caption{The \textit{accuracy vs. perturbation budget} curves of the $8$ models on ImageNet against targeted score-based attacks under the $\ell_2$ norm.}
\label{fig:score-t-l2-imagenet-acc-pert}
\begin{center}
\includegraphics[width=0.85\linewidth]{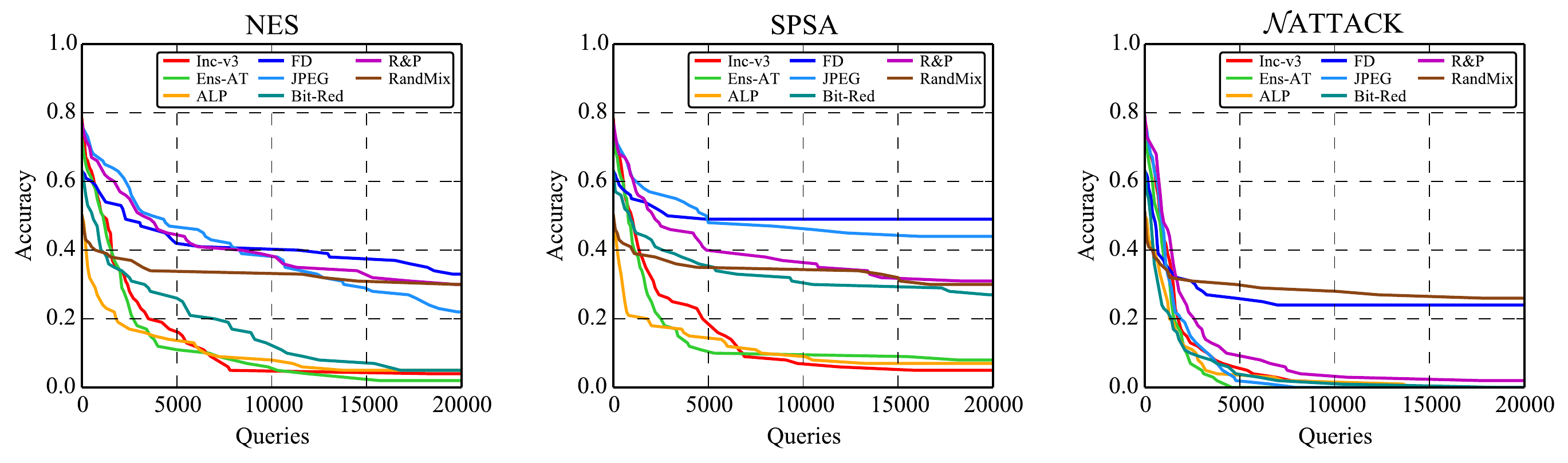}
\end{center}
\vspace{-2ex}
\caption{The \textit{accuracy vs. attack strength} curves of the $8$ models on ImageNet against targeted score-based attacks under the $\ell_2$ norm.}
\label{fig:score-t-l2-imagenet-acc-iter}
\end{figure*}

\begin{figure*}[t]
\begin{minipage}{.49\linewidth}
\begin{center}
\includegraphics[width=1.0\linewidth]{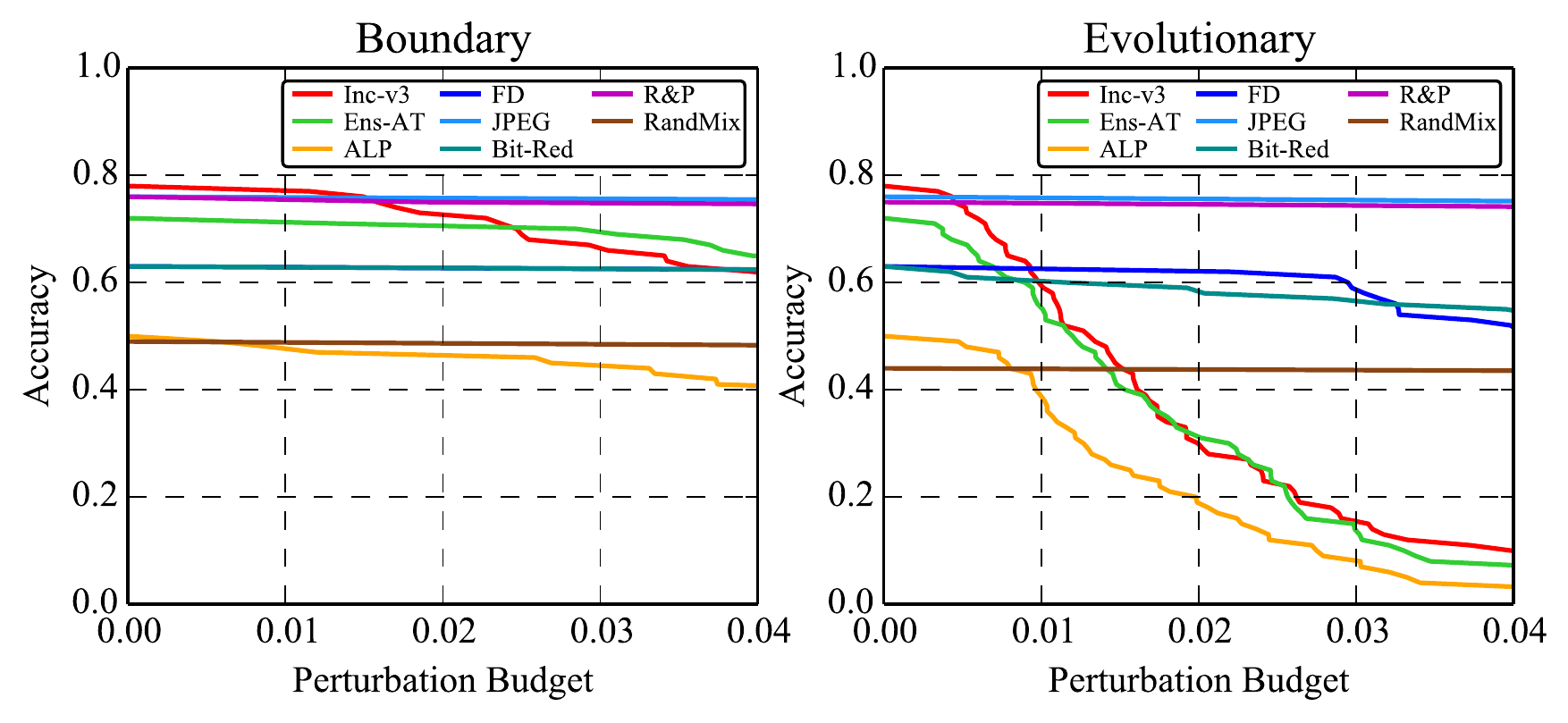}
\end{center}
\vspace{-2ex}
\caption{The \textit{accuracy vs. perturbation budget} curves of the $8$ models on ImageNet against targeted decision-based attacks under the $\ell_2$ norm.}
\label{fig:decision-t-l2-imagenet-acc-pert}
\end{minipage}
\hspace{1ex}
\begin{minipage}{.49\linewidth}
\begin{center}
\includegraphics[width=1.0\linewidth]{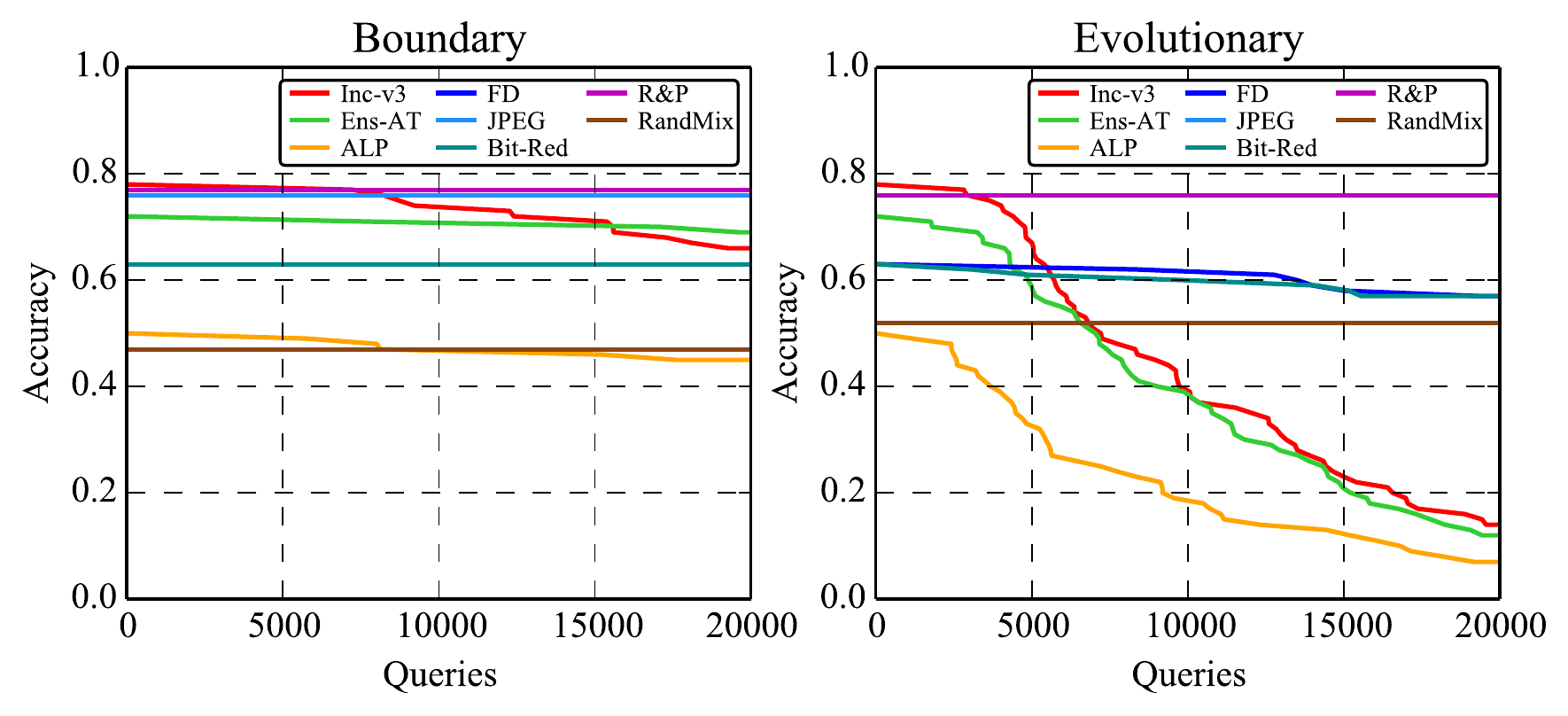}
\end{center}
\vspace{-2ex}
\caption{The \textit{accuracy vs. attack strength} curves of the $8$ models on ImageNet against targeted decision-based attacks under the $\ell_2$ norm.}
\label{fig:decision-t-l2-imagenet-acc-iter}
\end{minipage}
\end{figure*}

\begin{figure*}[t]
\begin{center}
\includegraphics[width=1.0\linewidth]{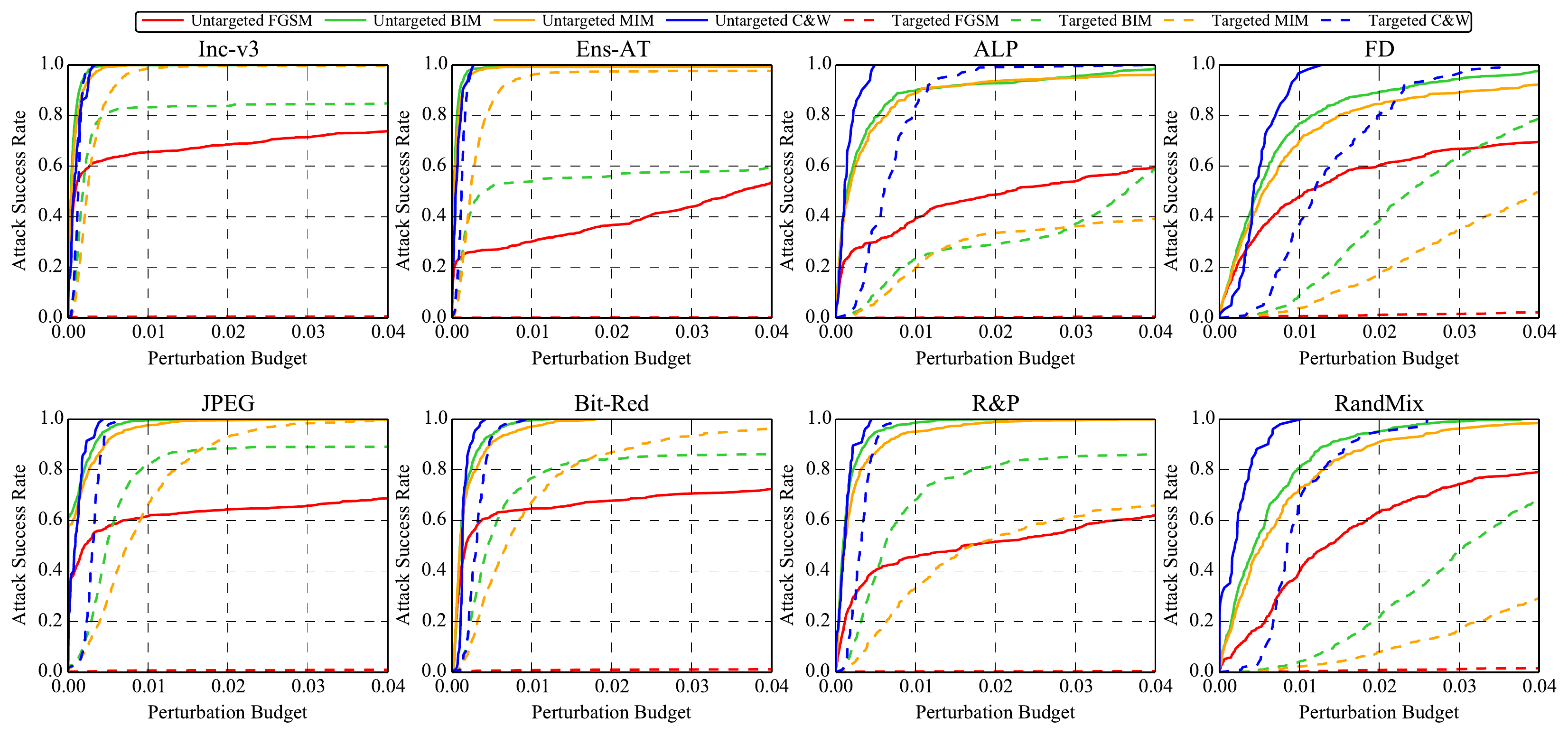}
\end{center}
\vspace{-2ex}
\caption{The \textit{attack success rate vs. perturbation budget} curves of white-box attacks under the $\ell_2$ norm on the $8$ models on ImageNet.}
\label{fig:white-l2-imagenet-asr-pert}
\end{figure*}

\begin{figure*}[t]
\begin{center}
\includegraphics[width=1.0\linewidth]{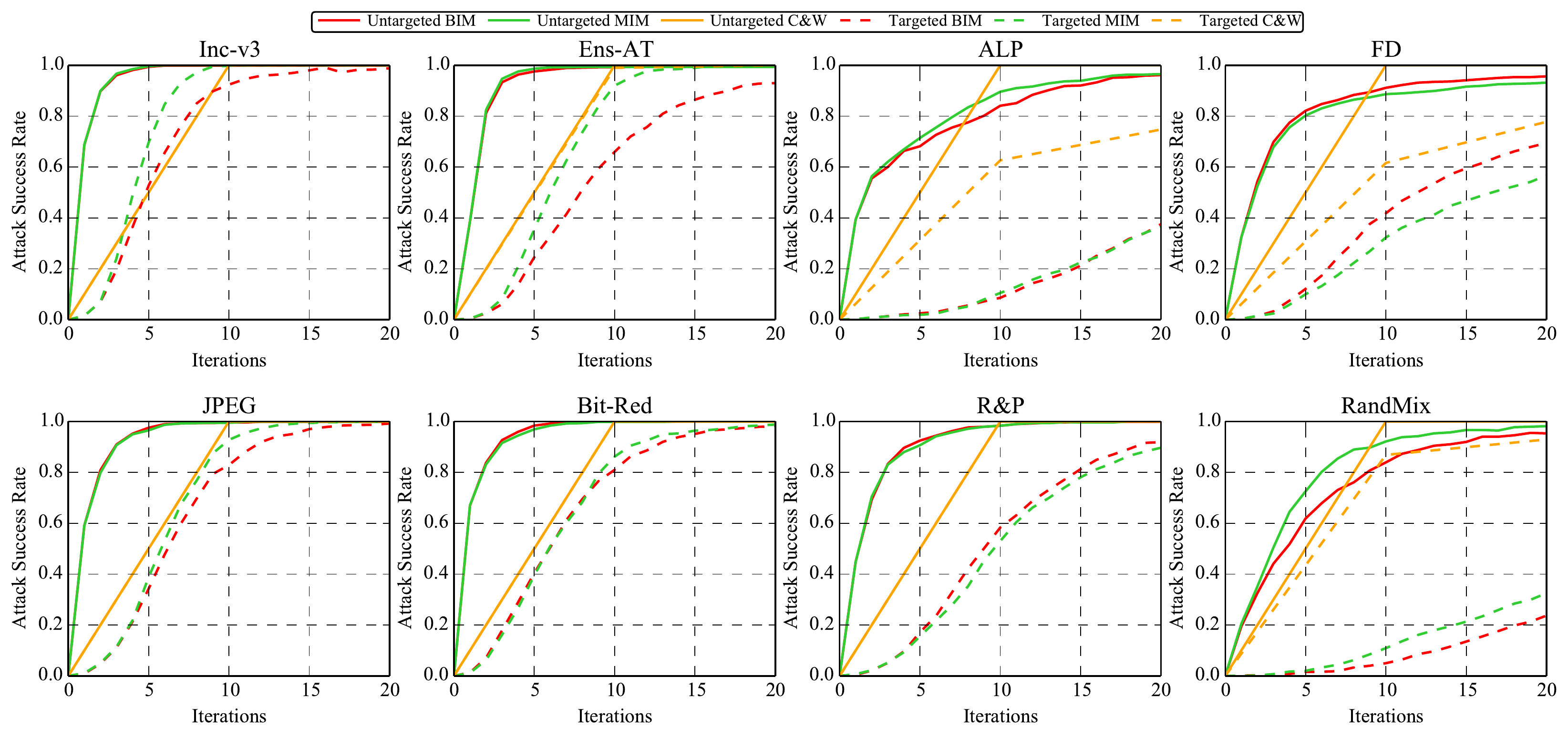}
\end{center}
\vspace{-2ex}
\caption{The \textit{attack success rate vs. attack strength} curves of white-box attacks under the $\ell_2$ norm on the $8$ models on ImageNet.}
\label{fig:white-l2-imagenet-asr-iter}
\end{figure*}

\begin{figure*}[t]
\begin{center}
\includegraphics[width=1.0\linewidth]{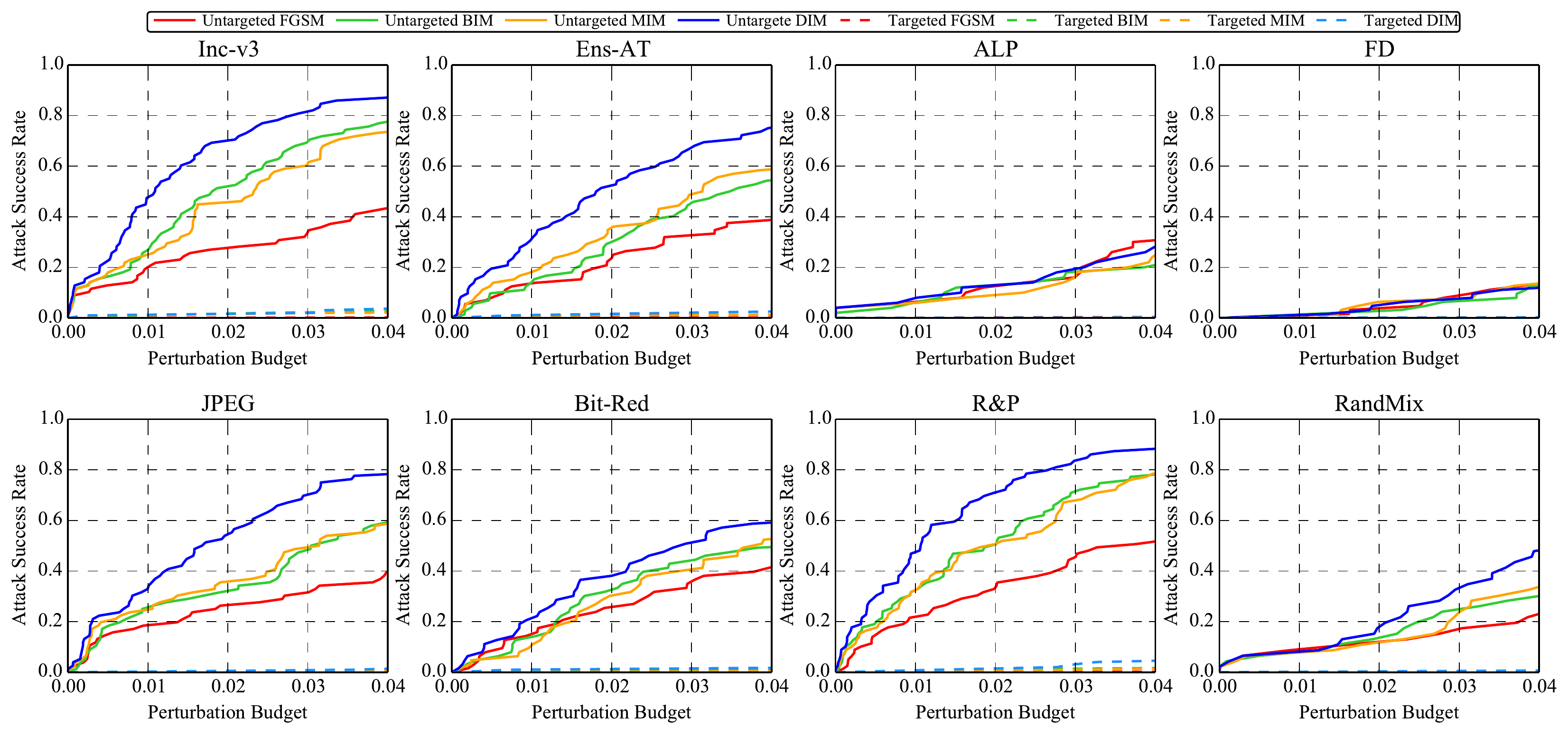}
\end{center}
\vspace{-2ex}
\caption{The \textit{attack success rate vs. perturbation budget} curves of transfer-based attacks under the $\ell_2$ norm on the $8$ models on ImageNet.}
\label{fig:trans-l2-imagenet-asr-pert}
\end{figure*}

\begin{figure*}[t]
\begin{center}
\includegraphics[width=1.0\linewidth]{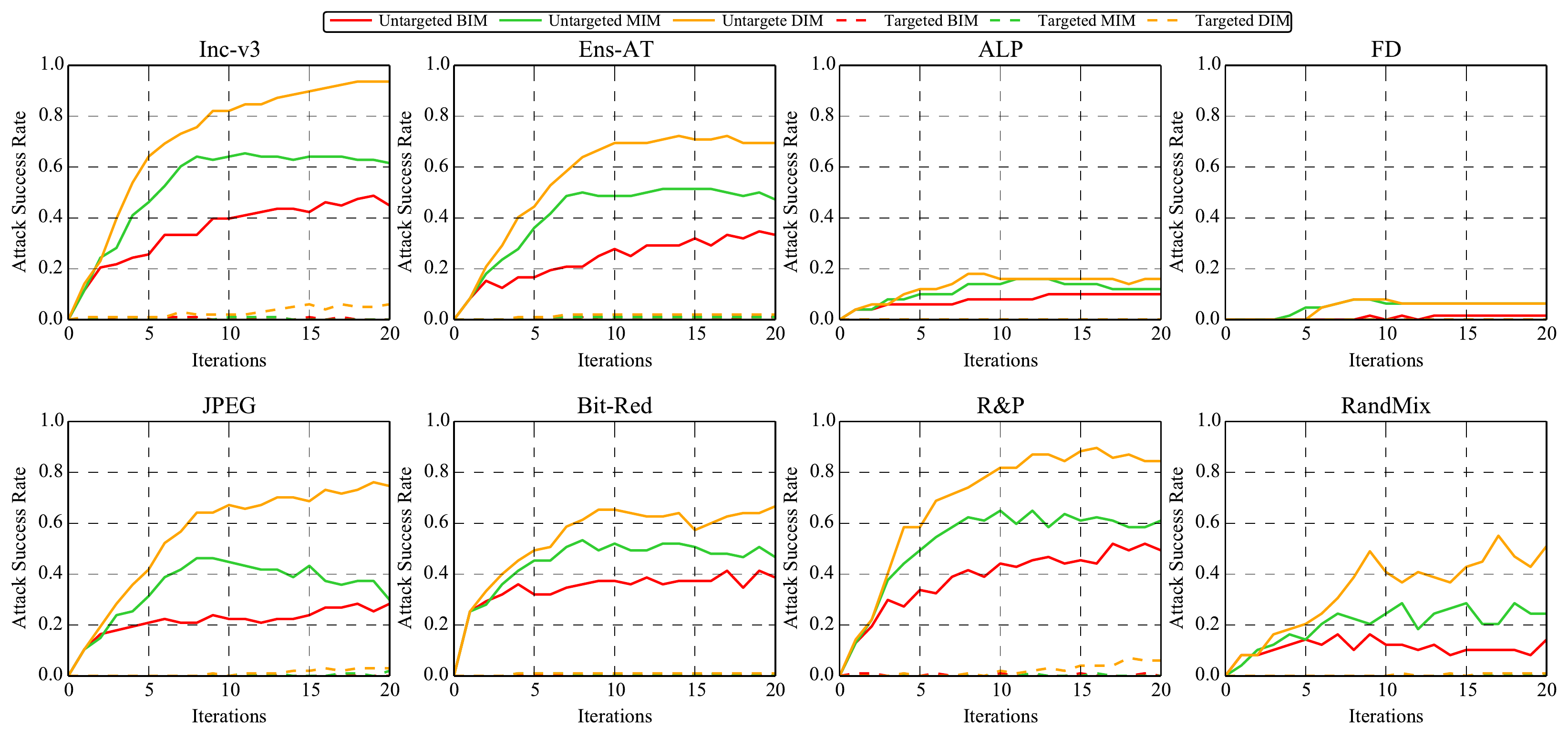}
\end{center}
\vspace{-2ex}
\caption{The \textit{attack success rate vs. attack strength} curves of transfer-based attacks under the $\ell_2$ norm on the $8$ models on ImageNet.}
\label{fig:trans-l2-imagenet-asr-iter}
\end{figure*}

\begin{figure*}[t]
\begin{center}
\includegraphics[width=1.0\linewidth]{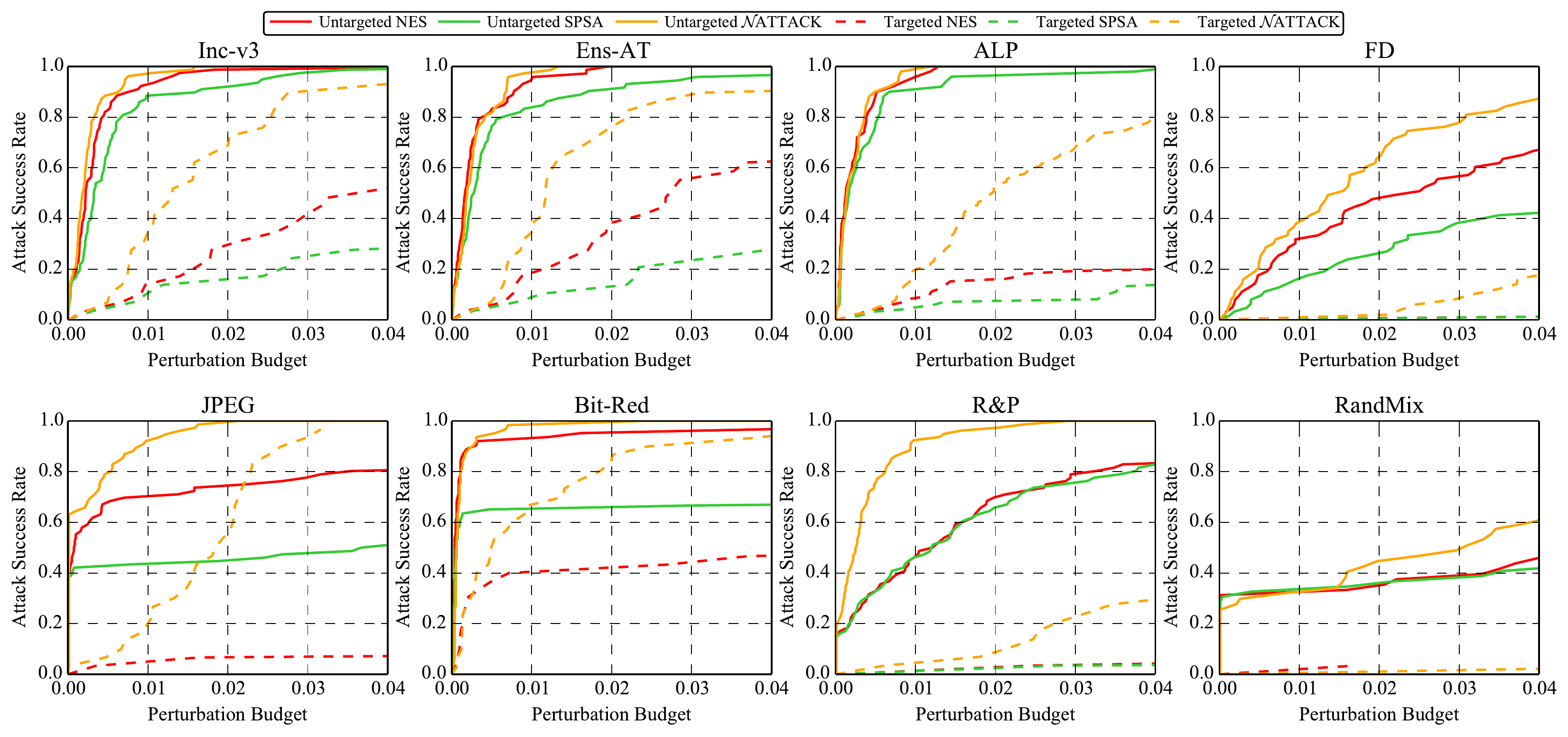}
\end{center}
\vspace{-2ex}
\caption{The \textit{attack success rate vs. perturbation budget} curves of score-based attacks under the $\ell_2$ norm on the $8$ models on ImageNet.}
\label{fig:score-l2-imagenet-asr-pert}
\end{figure*}

\begin{figure*}[t]
\begin{center}
\includegraphics[width=1.0\linewidth]{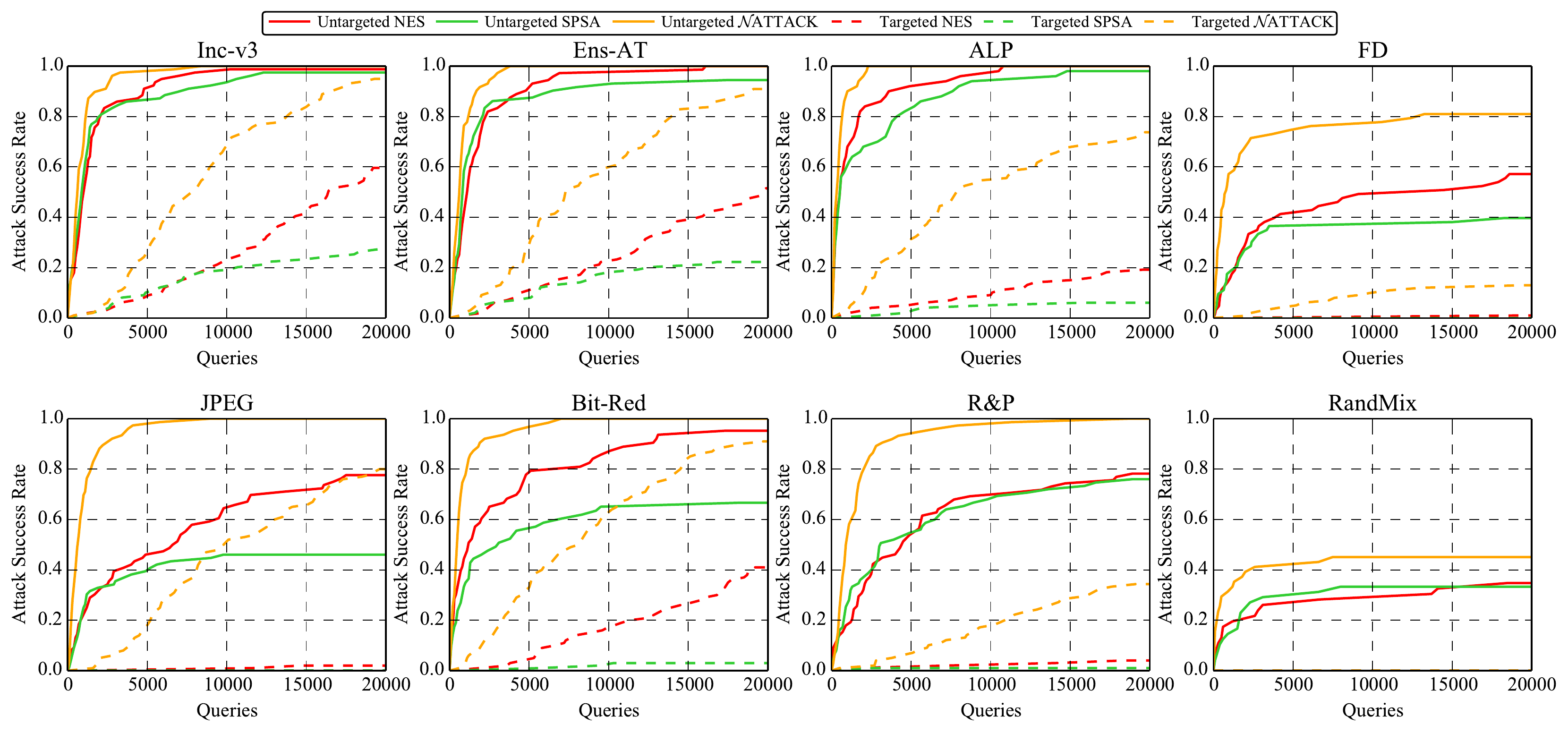}
\end{center}
\vspace{-2ex}
\caption{The \textit{attack success rate vs. attack strength} curves of score-based attacks under the $\ell_2$ norm on the $8$ models on ImageNet.}
\label{fig:score-l2-imagenet-asr-iter}
\end{figure*}

\begin{figure*}[t]
\begin{center}
\includegraphics[width=1.0\linewidth]{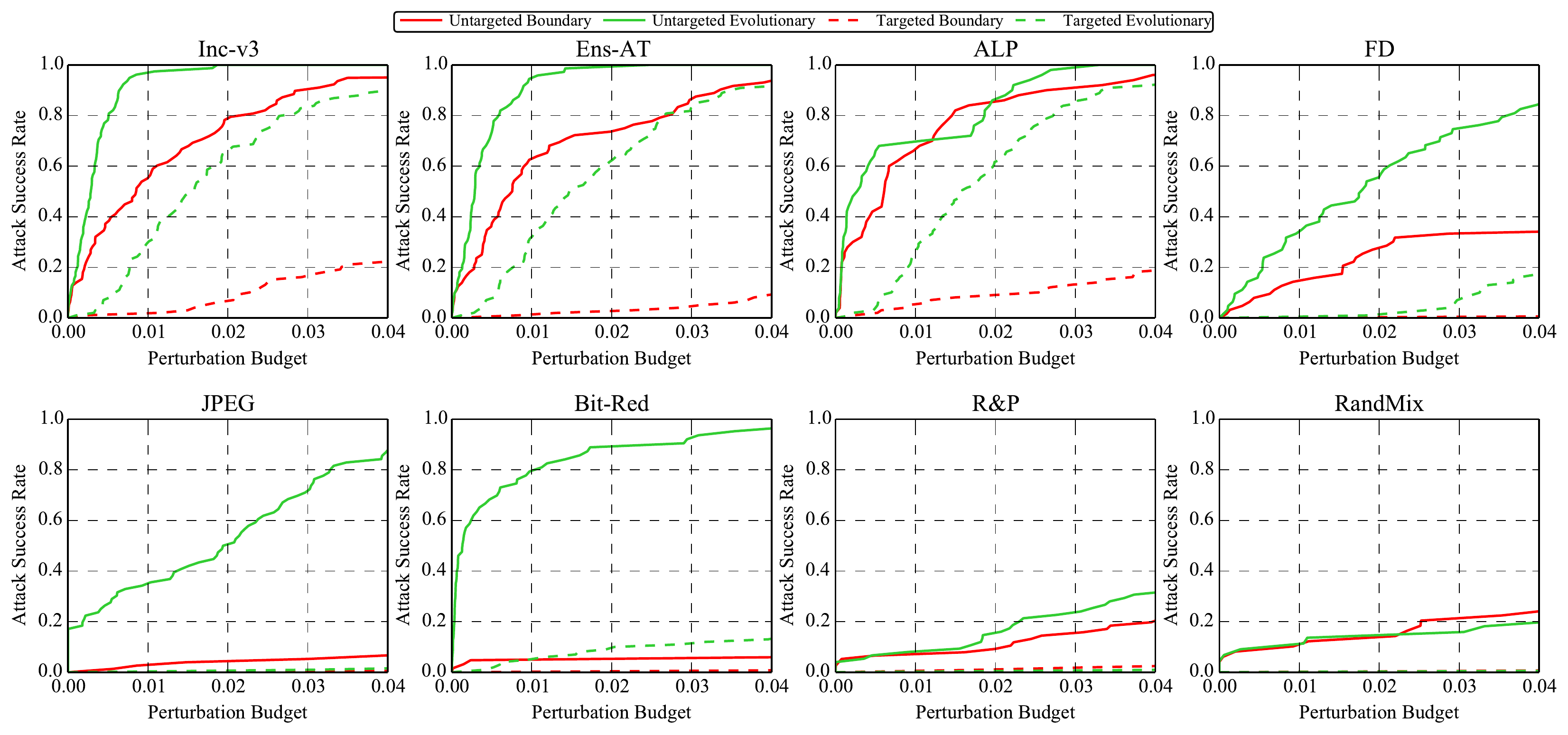}
\end{center}
\vspace{-2ex}
\caption{The \textit{attack success rate vs. perturbation budget} curves of decision-based attacks under the $\ell_2$ norm on the $8$ models on ImageNet.}
\label{fig:decision-l2-imagenet-asr-pert}
\end{figure*}

\begin{figure*}[t]
\begin{center}
\includegraphics[width=1.0\linewidth]{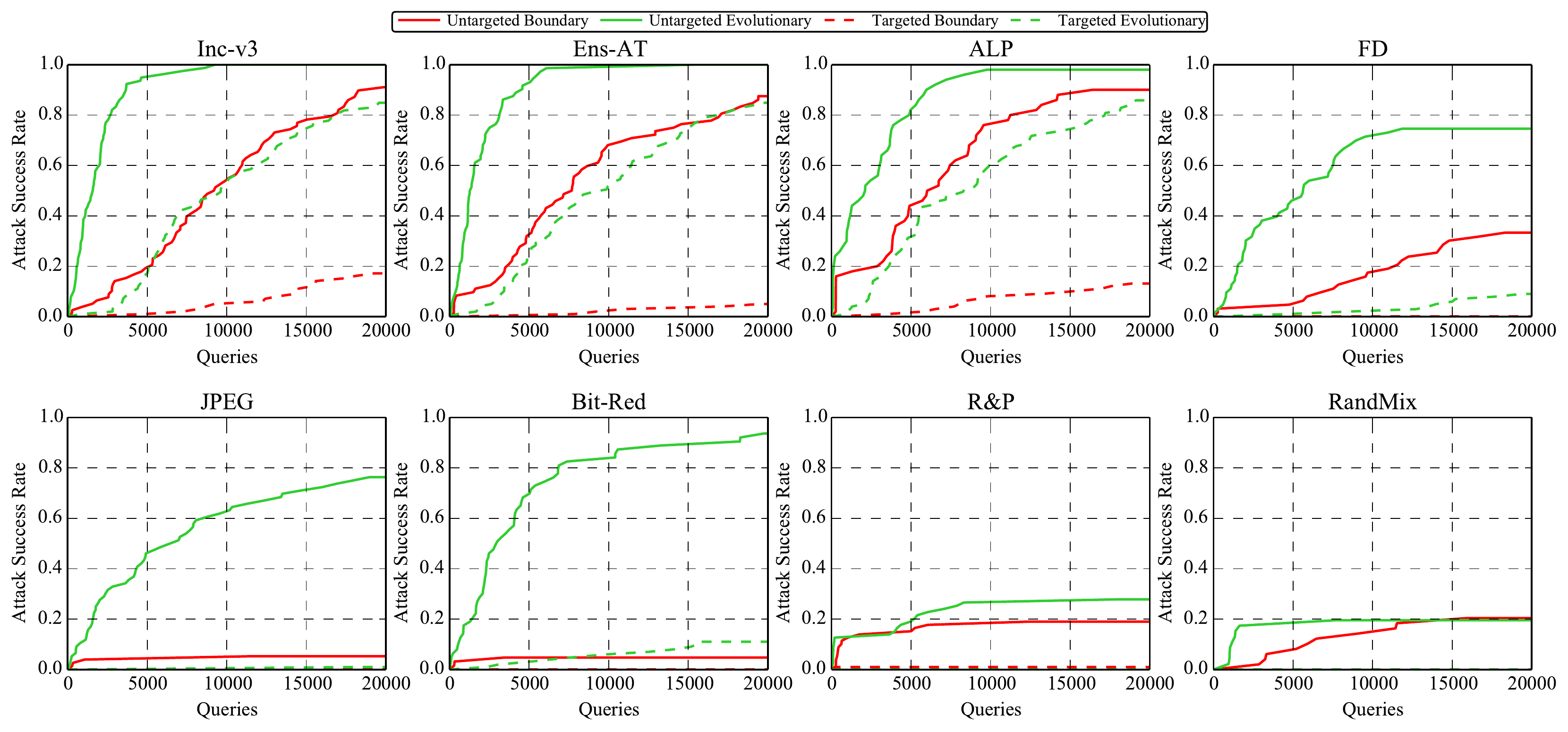}
\end{center}
\vspace{-2ex}
\caption{The \textit{attack success rate vs. attack strength} curves of decision-based attacks under the $\ell_2$ norm on the $8$ models on ImageNet.}
\label{fig:decision-l2-imagenet-asr-iter}
\end{figure*}

\textbf{Attacks under the $\ell_2$ norm:}
We show the accuracy curves of the defenses on ImageNet against untargeted and targeted white-box attacks under the $\ell_2$ norm in Fig.~\ref{fig:white-ut-l2-imagenet-acc-pert}, Fig.~\ref{fig:white-ut-l2-imagenet-acc-iter}, Fig.~\ref{fig:white-t-l2-imagenet-acc-pert}, and Fig.~\ref{fig:white-t-l2-imagenet-acc-iter}.
We show the accuracy curves of the defenses on ImageNet against untargeted and targeted transfer-based attacks under the $\ell_2$ norm in Fig.~\ref{fig:trans-ut-l2-imagenet-acc-pert}, Fig.~\ref{fig:trans-ut-l2-imagenet-acc-iter}, Fig.~\ref{fig:trans-t-l2-imagenet-acc-pert}, and Fig.~\ref{fig:trans-t-l2-imagenet-acc-iter}.
We show the accuracy curves of the defenses on ImageNet against untargeted and targeted score-based attacks under the $\ell_2$ norm in Fig.~\ref{fig:score-ut-l2-imagenet-acc-pert}, Fig.~\ref{fig:score-ut-l2-imagenet-acc-iter}, Fig.~\ref{fig:score-t-l2-imagenet-acc-pert}, and Fig.~\ref{fig:score-t-l2-imagenet-acc-iter}.
We show the accuracy curves of the defenses on ImageNet against untargeted and targeted decision-based attacks under the $\ell_2$ norm in Fig.~\ref{fig:decision-ut-l2-imagenet-acc-pert}, Fig.~\ref{fig:decision-ut-l2-imagenet-acc-iter}, Fig.~\ref{fig:decision-t-l2-imagenet-acc-pert}, and Fig.~\ref{fig:decision-t-l2-imagenet-acc-iter}.
Fig.~\ref{fig:white-l2-imagenet-asr-pert} to Fig.~\ref{fig:decision-l2-imagenet-asr-iter} show the \textit{attack success rate vs. perturbation budget} and \textit{attack success rate vs. attack strength} curves of white-box, transfer-based, score-based, and decision-based attacks under the $\ell_2$ norm on the $8$ models on ImageNet.

\clearpage

\clearpage

\clearpage

\end{document}